\documentclass[nohyperref]{article}

\usepackage{microtype}
\usepackage{graphicx}
\usepackage{subfigure}
\usepackage{booktabs} 
\usepackage{threeparttable}
\usepackage{multirow}

\usepackage{hyperref}

\usepackage[accepted]{icml2022}

\usepackage{amsmath}
\usepackage{amssymb}
\usepackage{mathtools}
\usepackage{amsthm}
\usepackage{bm}

\usepackage[capitalize,noabbrev]{cleveref}

\theoremstyle{plain}

\theoremstyle{definition}

\theoremstyle{remark}

\usepackage[textsize=tiny]{todonotes}

\icmltitlerunning{Preformer: Predictive Transformers with Multi-Scale Segment-wise Correlations}

\begin{document}

\twocolumn[
\icmltitle{Preformer: Predictive Transformer with Multi-Scale \\ Segment-wise Correlations
     for Long-Term Time Series Forecasting}



\icmlsetsymbol{equal}{*}

\begin{icmlauthorlist}
\icmlauthor{Dazhao Du}{iscas,ucas}
\icmlauthor{Bing Su}{rmu}
\icmlauthor{Zhewei Wei}{rmu}
\end{icmlauthorlist}

\icmlaffiliation{iscas}{Institute of Software Chinese Academy of Sciences, Beijing, China}
\icmlaffiliation{ucas}{University of Chinese Academy of Sciences, Beijing, China}
\icmlaffiliation{rmu}{Renmin University of China, Beijing, China}

\icmlcorrespondingauthor{Bing Su}{subingats@gmail.com}

\icmlkeywords{Machine Learning, ICML}

\vskip 0.3in
]



\printAffiliationsAndNotice{}  

\begin{abstract}
Transformer-based methods have shown great potential in long-term time series forecasting. However, most of these methods adopt the standard point-wise self-attention mechanism, which not only becomes intractable for long-term forecasting since its complexity increases quadratically with the length of time series, but also cannot explicitly capture the predictive dependencies from contexts since the corresponding key and value are transformed from the same point. This paper proposes a predictive Transformer-based model called {\em Preformer}. Preformer introduces a novel efficient {\em Multi-Scale Segment-Correlation} mechanism that divides time series into segments and utilizes segment-wise correlation-based attention for encoding time series. A multi-scale structure is developed to aggregate dependencies at different temporal scales and facilitate the selection of segment length. Preformer further designs a predictive paradigm for decoding, where the key and value come from two successive segments rather than the same segment. In this way, if a key segment has a high correlation score with the query segment, its successive segment contributes more to the prediction of the query segment. Extensive experiments demonstrate that our Preformer outperforms other Transformer-based methods.
\end{abstract}

\section{Introduction}
\label{intro}




Long-term time series forecasting has a wide range of real-world applications such as financial investment, traffic management, and electricity management. Existing deep learning-based methods can be divided into three categories, i.e., RNN-based models~\cite{dual-attention,LSTNet,song2018attend,deepar}, TCN-based models \cite{borovykh2017conditional,sen2019think} and Transformer-based models \cite{Transformer,influenza}. RNN-based methods suffer from the problem of gradient vanishing, gradient exploding, and lack of parallelism. TCN-based methods need deeper layers to achieve larger local receptive fields. For both categories of methods, signals must pass through a long path between two far-away temporal locations, hence the number of operations required to associate two elements increases with their temporal distance. Differently, transformer-based methods directly model the relationships between any element pairs and can better capture long-term dependencies, which is crucial for long-term forecasting.

\begin{figure*}[ht]
\vskip 0.2in
\begin{center}
    \subfigure[Conventional paradigm]{
    \includegraphics[width=0.5\columnwidth]{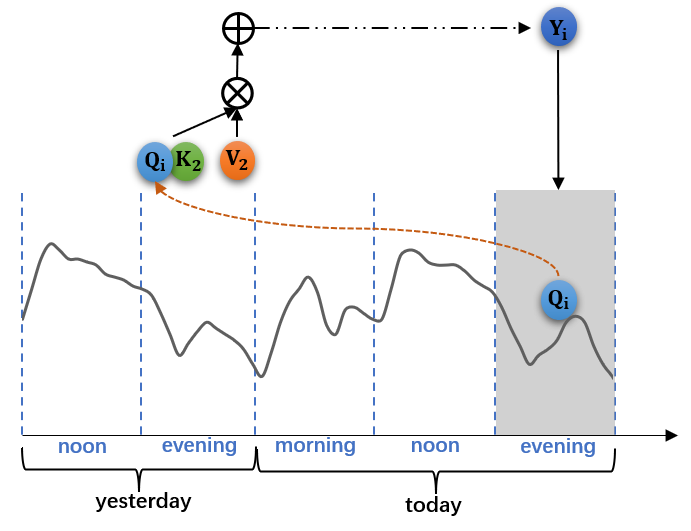}
    \includegraphics[width=0.385\columnwidth]{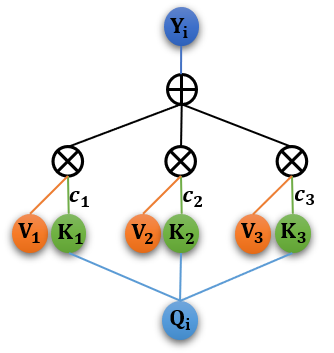}
    }
    \hspace{0.5cm}
    \subfigure[Predictive paradigm]{
    \includegraphics[width=0.5\columnwidth]{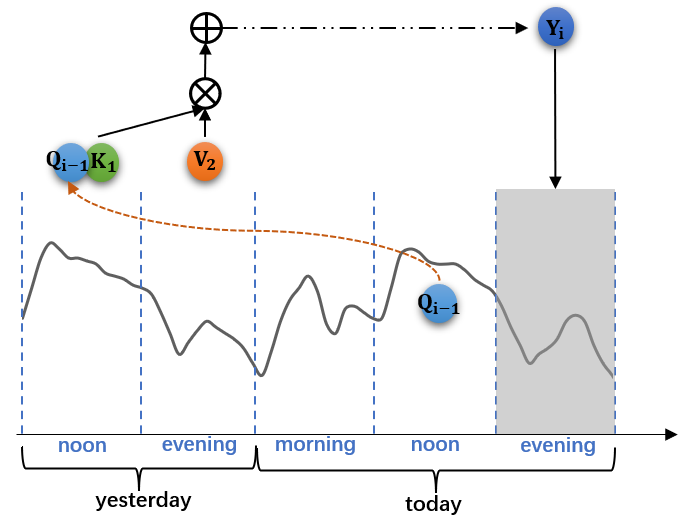}
    \includegraphics[width=0.38\columnwidth]{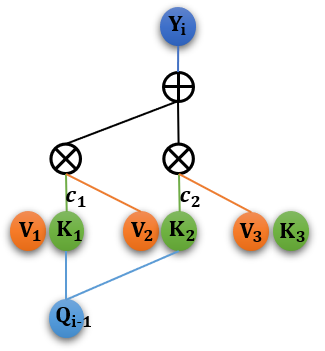}
    }
  \caption{(a) Segment-Correlation with the conventional paradigm. (b) Segment-Correlation with the predictive paradigm. The blue, green, and orange nodes represent the queries, keys and values generated by different segments (different time periods), respectively. The gray part in the time series is the segment to be predicted. Multiplication and addition represent weighted aggregation operations. $c_j$ is the correlation score between $Q_i$ and $K_j$. In the predictive paradigm, to obtain the output $Y_{i}$ at the current segment $i$, the similarity between $Q_{i-1}$ at the previous segment $i-1$ and the key $K_j$ is used as the weight of the value $V_{j+1}$ in the aggregation operation.}
\label{fig:intro}
\end{center}
\vskip -0.2in
\end{figure*}

On the other hand, the standard self-attention mechanism of Transformer calculates all similarities between any element pairs, where the computational and space complexities increase quadratically with the length of the time series. Recent works \cite{ConvTransformer,reformer,informer} explore different sparse attention mechanisms to suppress the contribution of irrelevant time steps and ease the computational pressure. These models still perform dot-product attention to time steps individually and utilize the point-wise connections to capture temporal dependencies. However, a single point may have limited influence on predicting the future. Autoformer \cite{autoformer} conducts the series-wise dependencies discovery by performing Auto-Correlation of the time series to the top-k time delayed series. The aggregation operation acts on the whole delayed series and complicated Fourier transforms are required for Auto-Correlation. 


These methods perform the correlation either at the point level or at the overall series level, which not only require high computational redundancy to intensively tackle point pairs or perform time-frequency domain transformations, but also do not directly reflect the true dependencies within the time series. For instance, in traffic flow forecasting, the flows of a single previous time point and the shifted whole time series may have limited contribution on the flow of the future period. There exist stronger correlations at the segment level, e.g., the flow at evening today should be more related to the flow at evening yesterday than the flow of other time periods any day and the flow of a period that has passed a long time ago.



To this end, we propose a novel sparse attention mechanism called {\em Multi-Scale Segment-Correlation (MSSC)}. Segment-wise correlation not only reduces the amount of calculation since the number of segments is much smaller than the number of points, but also better explore the locality of neighboring points. The length of segment is a critical hyperparameter. Long segments ignore fine-grained information while short segments have high computational redundancy. To tackle this issue, MSSC performs correlation calculations and fusion on multiple segment lengths, i.e., multi-scale resolutions, while maintaining low complexity. 


Standard attention paradigm can be applied to MSSC for time series encoding, but it is not well suited for forecasting since the unknown prediction segment generates the query by itself to predict itself. For prediction tasks, it is more reasonable to utilize the query of the previous segment to generate the prediction of the unknown segment. In this case, if the given query for prediction is similar to keys of some segments, their next segments rather than these segments themselves should contribute more to the prediction. As shown in \cref{fig:intro}, when we predict the flow at evening today given the flow at noon, we use the given flow at noon as the query rather than perform correlation for the unknown evening flow. If the flow at noon today is highly correlated with the flow at noon yesterday, then the flow of this evening should largely depend on the flow at evening yesterday rather than the flow at noon yesterday.

Motivated by this, we further propose a {\em Predictive Multi-Scale Segment-Correlation (PreMSSC)}, where current segment output $Y_i$ can be obtained via using the previous segment $Q_{i-1}$ to query all segments $\{K_1, K_2, \ldots\}$ and weighting the values of their next segments $\{V_2, V_3, \ldots\}$ by the calculated correlations. We derive our predictive model, namely {\em Predictive Transformer (Preformer)}, via replacing the standard self-attention and cross-attention in the original Transformer model with MSSC and PreMSSC, respectively.



The main contributions of this paper are as follows:
\begin{itemize}
\item We propose a novel MSSC mechanism to replace the canonical point-wise attention mechanism, which can increase the efficiency, extract more relevant information from the time series, and avoid the selection of segment length. 

\item We design a PreMSSC paradigm for forecasting by introducing a one-segment delay between the keys and their corresponding values. We develop a long-term forecasting model namely Preformer based on MSSC and PreMSSC. 

\item Extensive experiments show that our Preformer model achieves better forecasting performance than other Transformer-based prediction models.
\end{itemize}

\section{Related Work}

\textbf{Time Series Forecasting.} Early works on the TSF problem are based on classical mathematical models such as vector autoregression (VAR)~\cite{VAR} and auto regressive intergrated moving average (ARIMA) \cite{ARIMA}. Support vector regression (SVR) \cite{SVR} introduces a traditional machine learning method to regress the future. Gaussian Process \cite{GP} predicts the distribution of future values without assuming any certain form of the prediction function. However, all these classical models can not handle complicated data distributions or high-dimensional data.

With the development of deep learning, neural networks have shown stronger modeling ability than classical models. Recurrent Neural Network (RNN) \cite{rnn} and Temporal Convolution Network (TCN) \cite{Wavenet,TCN} are two common types of deep models for modeling sequence data. LSTNet \cite{LSTNet} combines convolutional layers and recurrent layers to capture both long-term and short-term dependencies. There are also some works \cite{dual-attention,song2018attend} that introduce additional attention mechanism to RNN to achieve better performance in forecasting. However, RNN-based models suffer from the gradient vanishing and gradient exploding problem. Popular variants of RNN such as LSTM \cite{LSTM} and GRU \cite{GRU} can not solve this problem fundamentally. The lack of parallelizability is another main limitation of RNN-based models. Benefiting from the good parallelism of convolution operations, TCN-based models \cite{borovykh2017conditional,sen2019think} have also achieved good results in time series tasks. Both RNN-based and TCN-based models do not explicitly model the dependencies between two far-away temporal locations, and the information exchange between them must go through a long path.


\textbf{Transfomer-based models.}  Transformer \cite{Transformer} was originally proposed as a sequence-to-sequence model in natural language processing to deal with machine translation. Due to its powerful modeling capabilities, it has even been widely applied in processing non-sequential data such as images \cite{ViT,objectdetection}. Self-attention plays an important role for explicitly discovering the dependencies between any element pairs, but both the time and space complexities increase quadratically with the length of the sequence, which limits the application of Transformer in long-term forecasting.

Therefore, various spare self-attention mechanisms have been proposed in recent years. Logfomer \cite{ConvTransformer} proposes LogSparse self-attention which selects elements in exponentially increasing intervals to break the memory bottleneck. Informer \cite{informer} defines a sparsity measurement for queries and selects dominant queries based on this measurement to obtain ProbSparse self-attention. These works use point-wise dot product to compute attention score, and differ in the way of selecting point pairs. 

AutoFomer \cite{autoformer} develops an Auto-Correlation mechanism to replace self-attention, which utilizes series-wise correlation instead of point-wise dot product. In this work, we introduce a new Segment-Correlation mechanism to explore the context information within neighboring points and capture the segment-wise correlation in the sequence. Our method differs from the Auto-Correlation mechanism in the way of correlation computation and aggregation. Instead of the complicated Fast Fourier Transforms calculation in Auto-Correlation, we directly segment the time series based on implicit period and compute the correlation between segments. Benefiting from the simplicity and intuitiveness of Segment-Correlation, we have customized a multi-scale structure and predictive paradigm for long-term forecasting.


\section{Method}
\label{method}

\subsection{Problem Definition}
Multi-horizon time series forecasting aims to predict values at multiple future time steps. Typically, given the previous time series $X_{1:t_0}=\{x_1, x_2, \dots, x_{t_0}\}$, where $x_t\in \mathbb{R}^{d_x}$ and $d_x$ is the dimension of the variable, we aim to predict the future values $Y_{t_0+1:t_0+\tau}=\{y_{t_0+1},y_{t_0+2},\dots,y_{t_0+\tau}\}$, where $y_t\in \mathbb{R}^{d_y}$ is the prediction at every time step $t$ and $d_y$ is the dimension of the output variable. The prediction model $f$ can be formulated as:
\begin{equation}
\label{equ:problem}
    \hat Y_{t_0+1:t_0+\tau}=f(X_{1:t_0};\Omega),
\end{equation}

where $\hat Y_{t_0+1:t_0+\tau}$ is the predicted time series and $\Omega$ is the learnable parameters of the model. For long-term forecasting, the future time duration $\tau$ to be predicted, is longer. The problem can be categorized into two types based on whether the dimension of the output variable $d_y$ is larger than one: univariate forecasting and multivariate forecasting.

\subsection{The Preformer Model}

\begin{figure}[ht]
\vskip 0.2in
\begin{center}
\centerline{\includegraphics[width=0.75\columnwidth]{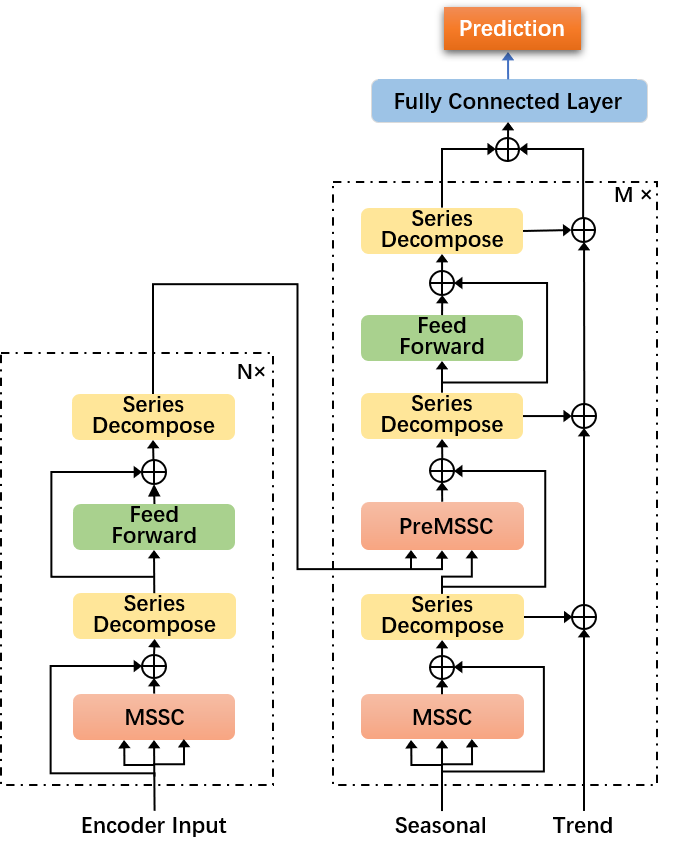}}
\caption{The overall framework of the Preformer model.}
\label{fig:model}
\end{center}
\vskip -0.2in
\end{figure}

As shown in Figure \ref{fig:model}, the overall architecture of Preformer is similar to the vanilla Transformer. We replace the dot-product self-attention in Transformer with MSSC and replace the cross-attention with PreMSSC. Time series decomposition methods which deconstruct time series into several components are widely used in time series analysis. Decoupling the time series into several components has been shown to be helpful for forecasting tasks \cite{Prophet,nbeats}. Here we use the series decomposition module proposed in Autoformer \cite{autoformer} to decompose the time series into trend and seasonal components. A fully connected layer is added to convert the decoder output into prediction values, which will be used to compute the MSE loss $\mathcal{L}_{MSE}$ with the ground truth. 

\textbf{Model inputs.}\quad 
The way the original time series is input to the Preformer (encoder and decoder) is consistent with AutoFormer \cite{autoformer}. Following other methods \cite{informer,autoformer}, we utilize additional time-dependent features (e.g., a set of dummy variables like hour-of-the-day, day-of-the-week, etc) called covariates as parts of inputs. Specifically, we concatenate the values of the past original time series $X_{1:t_0}$ and covariates $C_{1:t_0}$ to derive the inputs of the Preformer encoder $\mathcal{X}_{1:t_0}=[X_{1:t_0};C_{1:t_0}]$. Because these covariates can be predetermined (e.g. the hour-of-the-day at any time), we also use $C_{t_0+1:t_0+\tau}$ as parts of the decoder inputs. Before being fed to the MSSC module, inputs to the encoder and decoder are transformed into the feature dimension through an embedding layer which has been widely used in Transformer-based models.

\textbf{Encoder.}\quad
The encoder consists of $N$ identical layers, where each layer consists of a MSSC module and a feed forward network each followed by a series decomposition module with residual connections. The input of the encoder is $\mathcal{X}_{1:t_0}$, which includes the past time series values and covariates. The series decomposition module passes the inputs $\mathcal{X}$ through an average pooling layer to get the trend component $\mathcal{X}^{trend}$, and subtracts the trend component from the inputs to get the seasonal component $\mathcal{X}^{season}=\mathcal{X}-\mathcal{X}^{trend}$. All decomposition modules in the encoder eliminate the trend component, which makes the encoder focus on seasonal pattern modeling.

\textbf{Decoder.}\quad
The decoder consists of $M$ identical layers. Different from the encoder, there is an additional PreMSSC module where key and value matrices are transformed from the outputs of the encoder in each decoder layer. The inputs of the decoder include two components: seasonal and trend components. Each component is composed of two parts: information from the latter half part of the original time series and placeholders filled by scalars. Specifically, the latter half part of the original time series $X_{half}$ (i.e., $X_{t_0/2:t_0}$) are decomposed into $X^{season}_{half}$ and $X^{trend}_{half}$, which will be concatenated with placeholders to get seasonal inputs $[X^{season}_{half}, X_0]$ and trend inputs $[X^{trend}_{half}, X_{mean}]$, where $X_0,X_{mean}\in \mathbb{R}^{\tau\times d_x}$ denote the placeholders filled with zero and the mean of $X_{half}$ respectively. The decomposition modules in the decoder extract the trend part from hidden variables progressively, which is finally added to the seasonal part to derive the output.

An additional fully connected layer takes the output of the decoder as input, and it generates final prediction values $\hat Y_{t_0+1:t_0+\tau}\in \mathbb{R}^{\tau\times d_y}$.

\subsection{Segment-Correlation}
\begin{figure*}[ht]
\vskip 0.2in
\begin{center}
\includegraphics[scale=0.36]{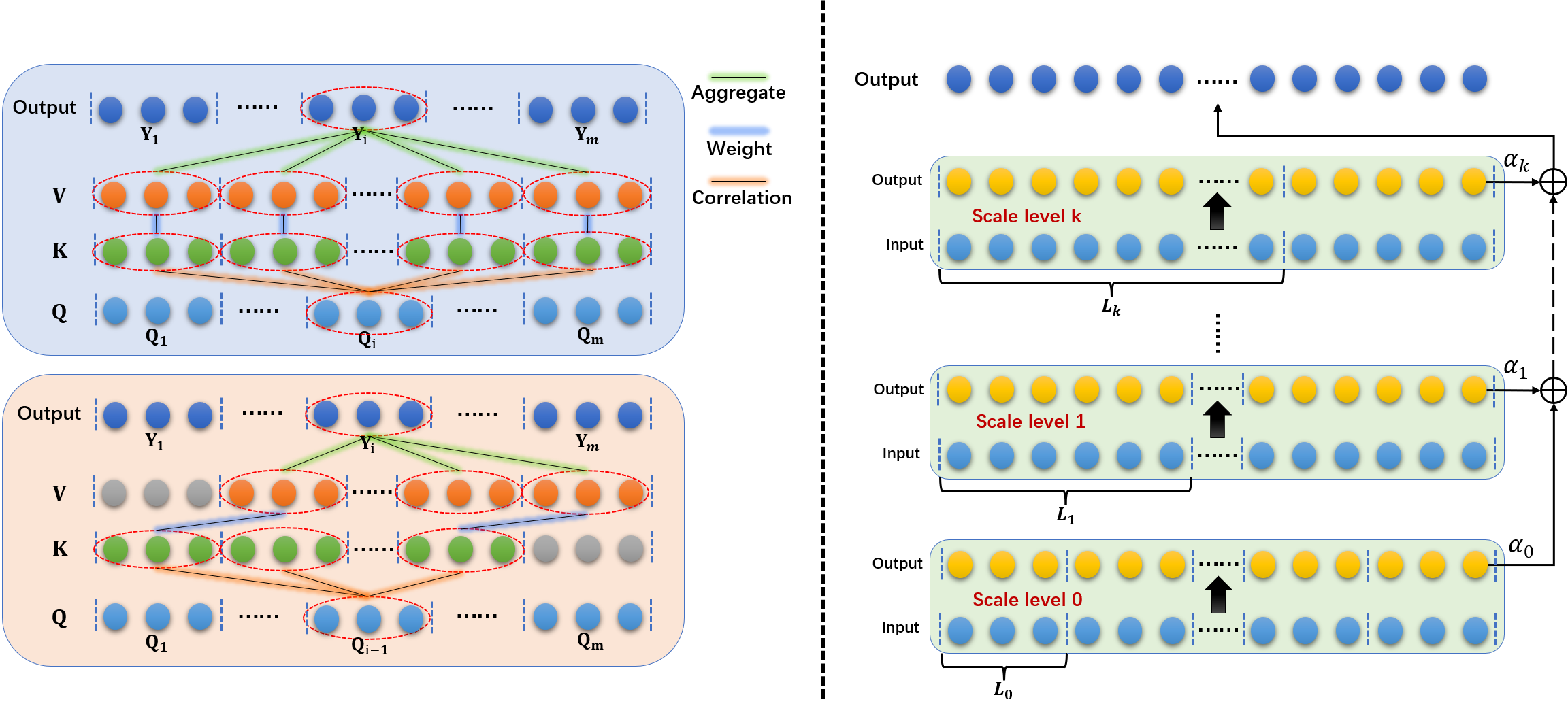}
\caption{Segment-Correlation (upper left), predictive paradigm (bottom Left) and the multi-scale architecture (right). In Segment-Correlation, the query, key and value are segmented evenly. Then the correlation measurements between all segments are calculated, and the derived weights are used to aggregate the value segments to get the output. Only the query of the $i$-th segment $Q_i$ and the output $Y_i$ are illustrated here for simplicity. In the multi-scale structure, each scale level is a Segment-Correlation module with or without predictive paradigm, representing MSSC and PreMSSC respectively.}
\label{fig:PreMSSC}
\end{center}
\vskip -0.2in
\end{figure*}

Segment-Correlation is the key module in Preformer, which performs segment-wise attention instead of point-wise attention. We denote the input of each Segment-Correlation module as $H \in \mathbb{R}^{L\times d}$, where $L$ and $d$ are the length and dimension of the input respectively. Formally, for the single head situation, the input series $H$ will be projected by three projection matrices to obtain the query, key and value, i.e., $Q=HW_q, K=HW_k, V=HW_v$. Then all the $Q$, $K$ and $V$ are segmented into several segments having the same length $L_{seg}$: $\{Q_1, Q_2, \dots, Q_m\}$, $\{K_1, K_2, \dots, K_n\}$, $\{V_1, V_2, \dots, V_n\}$, where $Q_i \in \mathbb R^{L_{seg}\times d}$, $K_i \in \mathbb R^{L_{seg}\times d}$, $V_i \in \mathbb R^{L_{seg}\times d}$, $m, n$ denote the number of segments, and $L_{seg}$ is a hyperparameter that determines the computational complexity via controlling the length of segments.

The correlation measurement $c_{ij}$ between any pair of query segment $Q_i$ and key segment $K_j$ can be computed by the function:
\begin{equation}
    c_{ij} = \textup{Correlation}(Q_i,K_j)=\frac{1}{d\times L_{seg}}Q_i \odot K_j,
\end{equation}
where $\odot$ is the dot product operator between two matrices of the same size. For each query segment $Q_i$, whose correlation measurements with all the key segments will be normalized by the $\textup{Softmax}$ function to obtain the aggregation weight:
\begin{equation}
    \hat c_{i1},\hat c_{i2},\dots,\hat c_{in} = \textup{Softmax}(c_{i1}, c_{i2}, \dots, c_{in}).
\end{equation}
The output at the position of the $i$-th segment $Y_i$ is the weighted sum of all the value segments $\{V_j\mid j=1,\dots,n\}$:
\begin{equation}
\label{equ:yi}
    Y_i = \sum_{j=1}^{n} \hat c_{ij} V_j.
\end{equation}
Lastly, output of the Segment-Correlation module can be obtained by concatenating all the $Y_i$ along length dimension: 
\begin{equation}
    \textup{SC}(H;L_{seg}) = \textup{Concat}(Y_1,\dots,Y_m),
\end{equation}
where $\textup{SC}$ is the abbreviation of Segment-Correlation. The output of multi-head Segment-Correlation can be computed by concatenating and projecting the outputs of all heads described above. We omit the formulation since it is similar to the canonical multi-head attention \cite{Transformer}.

\textbf{Multi-Scale Segment-Correlation.}\quad
Since $L_{seg}$ determines the resolution of the smallest unit involved in the Segment-Correlation calculation, a large $L_{seg}$ means that the coarse-grained temporal dependencies in the time series can be captured, while Segment-Correlation with a small $L_{seg}$ can capture fine-grained dependencies. To attenuate the impact of hyperparameter $L_{seg}$ selection on performance, we propose Multi-Scale Segment-Correlation (MSSC) by fusing the output of multiple Segment-Correlation with different $L_{seg}$. Specifically, as the scale level increases, we start with a small initial segment length $L_0$ and increase the segment length exponentially, i.e., $L_{l} = 2^lL_0$, where $l\in \{0,1,\ldots,l_{max}\}$ denotes the scale level and $l_{max}=\lfloor \log_2(\frac{L}{L_0}) \rfloor$. The input $H$ is passed through these Segment-Correlation layers of different scale levels to obtain the outputs of the corresponding scale levels. We aggregate the outputs of all scale levels to get the output of the entire MSSC module, where the weight of the $l$-th level $\alpha_l$ is set to decrease exponentially as the scale level increases. Therefore, MSSC can be formulated as:
\begin{equation}
\begin{aligned}
    \textup{MSSC}(H) &= \sum_{l=0}^{l_{max}} \alpha_l \cdot \textup{SC}(H;2^lL_0),\\
    \alpha_l &= \frac{2^l}{\sum_{l=0}^{l_{max}}2^l}.
\end{aligned}
\end{equation}

\textbf{Predictive paradigm.}\quad
In the decoding phase, the query for the period to be predicted is its preceding segment rather than itself. Therefore, if some segments with respect to keys are highly relevant to the query, their future segments rather than themselves should contribute more to the prediction of the query. Inspired by this intuition, we propose a predictive paradigm for cross-attention by introducing a segment delay between the keys and their corresponding values. The basic idea of the predictive paradigm is shown in \cref{fig:intro}, \cref{intro}. For Segment-Correlation without the predictive paradigm, the output at the position of the $i$-th segment $Y_i$ is obtained according to \cref{equ:yi}. If we reformulate $\hat c_{ij}$ as $\hat c_{(Q_i,K_j)}$, then we have:
\begin{equation}
    Y_i = \sum_{j=1}^{n} \hat c_{(Q_i,K_j)} V_j.
\end{equation}
There are two differences between the predictive paradigm and the non-predictive paradigm. Firstly, to get the output of the current segment $Y_i$, the query of the previous segment $Q_{i-1}$ is used to calculate correlations with the keys of all segments $\{K_1,\ldots,K_{n-1}\}$. Secondly, the correlation $\hat c_{(Q_{i-1},K_j)}$ corresponding to the key of a certain segment $K_j$ is regarded as the weight of the next segment $V_{j+1}$ to aggregate values. That is, the segment of value is delayed by one segment relative to the segment of the corresponding key. Therefore, we can obtain $Y_i$ by the following equation:
\begin{equation}
\begin{aligned}
Y_1 &= \sum_{j=1}^{n-1} \hat c_{(Q_{m},K_j)} V_{j+1},\\
Y_i &= \sum_{j=1}^{n-1} \hat c_{(Q_{i-1},K_j)} V_{j+1}, \quad i>1.
\end{aligned}
\end{equation}

\begin{table*}[ht]
\renewcommand\arraystretch{1}  
\caption{Multivariate time-series forecasting results on six datasets}
\label{tab:multivariate}
\vskip 0.15in
\begin{center}
\begin{scriptsize}
\begin{threeparttable}
\begin{tabular}{cc|cc|cc|cc|cc|cc|cc|cc} 
\toprule[1.3pt]
\multicolumn{2}{c}{Models} & \multicolumn{2}{c}{Preformer} & \multicolumn{2}{c}{Autoformer} & \multicolumn{2}{c}{Informer} & \multicolumn{2}{c}{LogTrans} &\multicolumn{2}{c}{LSTNet} & \multicolumn{2}{c}{LSTM} & \multicolumn{2}{c}{TCN}\\
\cmidrule(lr){3-4}\cmidrule(lr){5-6}\cmidrule(lr){7-8}\cmidrule(lr){9-10}\cmidrule(lr){11-12}\cmidrule(lr){13-14}\cmidrule(lr){15-16}
\multicolumn{2}{c}{Metric} & \multicolumn{1}{c}{MSE} & \multicolumn{1}{c}{MAE} & \multicolumn{1}{c}{MSE} & \multicolumn{1}{c}{MAE} & \multicolumn{1}{c}{MSE} & \multicolumn{1}{c}{MAE} & \multicolumn{1}{c}{MSE} & \multicolumn{1}{c}{MAE} & \multicolumn{1}{c}{MSE} & \multicolumn{1}{c}{MAE} & \multicolumn{1}{c}{MSE} & \multicolumn{1}{c}{MAE} & \multicolumn{1}{c}{MSE} & \multicolumn{1}{c}{MAE}\\
\midrule
\midrule
\multirow{4}{*}{\rotatebox{90}{ETTm2}} 
& 96 & \textbf{0.213} & \textbf{0.295} & \underline{0.255} & \underline{0.339} & 0.365 & 0.453 & 0.768 & 0.642 & 3.142 & 1.365 & 2.041 & 1.073 & 3.041 & 1.330\\
& 192 & \textbf{0.269} & \textbf{0.329} & \underline{0.281} & \underline{0.340} & 0.533 & 0.563 & 0.989 & 0.757 & 3.154 & 1.369 & 2.249 & 1.112 & 3.072 & 1.339\\
& 336 & \textbf{0.324} & \textbf{0.363} & \underline{0.339} & \underline{0.372} & 1.363 & 0.887 & 1.334 & 0.872 & 3.160 & 1.369 & 2.568 & 1.238 & 3.105 & 1.348\\
& 720 & \textbf{0.418} & \textbf{0.416} & \underline{0.422} & \underline{0.419} & 3.379 & 1.388 & 3.048 & 1.328 & 3.171 & 1.368 & 2.720 & 1.287 & 3.135 & 1.354\\
\midrule
\multirow{4}{*}{\rotatebox{90}{Electricity}} 
& 96 & \textbf{0.180} & \textbf{0.297} & \underline{0.201} & \underline{0.317} & 0.274 & 0.368 & 0.258 & 0.357 & 0.680 & 0.645 & 0.375 & 0.437 & 0.985 & 0.813 \\
& 192 & \textbf{0.189} & \textbf{0.302} & \underline{0.222} & \underline{0.334} & 0.296 & 0.386 & 0.266 & 0.368 & 0.725 & 0.676 & 0.442 & 0.473 & 0.996 & 0.821\\
& 336 & \textbf{0.201} & \textbf{0.319} & \underline{0.231} & \underline{0.338} & 0.300 & 0.394 & 0.280 & 0.380 & 0.828 & 0.727 & 0.439 & 0.473 & 1.000 & 0.824\\
& 720 & \textbf{0.232} & \textbf{0.342} & \underline{0.254} & \underline{0.361} & 0.373 & 0.439 & 0.283 & 0.376 & 0.957 & 0.811 & 0.980 & 0.814 & 1.438 & 0.784\\
\midrule
\multirow{4}{*}{\rotatebox{90}{Exchange}} 
& 96 & \textbf{0.148} & \textbf{0.282} & \underline{0.197} & \underline{0.323} & 0.847 & 0.752 & 0.968 & 0.812 & 1.551 & 1.058 & 1.453 & 1.049 & 3.004 & 1.432\\
& 192 & \textbf{0.268} & \underline{0.378} & \underline{0.300} & \textbf{0.369} & 1.204 & 0.895 & 1.040 & 0.851 & 1.477 & 1.028 & 1.846 & 1.179 & 3.048 & 1.444\\
& 336 & \textbf{0.447} & \textbf{0.499} & \underline{0.509} & \underline{0.524} & 1.672 & 1.036 & 1.659 & 1.081 & 1.507 & 1.031 & 2.136 & 1.231 & 3.113 & 1.459\\
& 720 & \textbf{1.092} & \textbf{0.812} & \underline{1.447} & \underline{0.941} & 2.478 & 1.310 & 1.941 & 1.127 & 2.285 & 1.243 & 2.984 & 1.427 & 3.150 & 1.458\\
\midrule
\multirow{4}{*}{\rotatebox{90}{Traffic}} 
& 96 & \textbf{0.560} & \textbf{0.349} & \underline{0.613} & \underline{0.388} & 0.719 & 0.391 & 0.684 & 0.384 & 1.107 & 0.685 & 0.843 & 0.453 & 1.438 & 0.784\\
& 192 & \textbf{0.565} & \textbf{0.349} & \underline{0.616} & \underline{0.382} & 0.696 & 0.379 & 0.685 & 0.390 & 1.157 & 0.706 & 0.847 & 0.453 & 1.463 & 0.794\\
& 336 & \textbf{0.577} & \underline{0.351} & \underline{0.622} & \textbf{0.337} & 0.777 & 0.420 & 0.733 & 0.408 & 1.216 & 0.730 & 0.853 & 0.455 & 1.479 & 0.799\\
& 720 & \textbf{0.597} & \textbf{0.358} & \underline{0.660} & \underline{0.408} & 0.864 & 0.472 & 0.717 & 0.396 & 1.481 & 0.805 & 1.500 & 0.805 & 1.499 & 0.804\\
\midrule
\multirow{4}{*}{\rotatebox{90}{Weather}} 
& 96 & \textbf{0.227} & \textbf{0.292} & \underline{0.266} & \underline{0.336} & 0.300 & 0.384 & 0.458 & 0.490 & 0.594 & 0.587 & 0.369 & 0.406 & 0.615 & 0.589\\
& 192 & \textbf{0.275} & \textbf{0.322} & \underline{0.307} & \underline{0.367} & 0.598 & 0.544 & 0.658 & 0.589 & 0.560 & 0.565 & 0.416 & 0.435 & 0.629 & 0.600\\
& 336 & \textbf{0.324} & \textbf{0.352} & \underline{0.359} & \underline{0.395} & 0.578 & 0.523 & 0.797 & 0.652 & 0.597 & 0.587 & 0.455 & 0.454 & 0.639 & 0.608\\
& 720 & \textbf{0.394} & \textbf{0.393} & \underline{0.419} & \underline{0.428} & 1.059 & 0.741 & 0.869 & 0.675 & 0.618 & 0.599 & 0.535 & 0.520 & 0.639 & 0.610\\
\midrule
\multirow{4}{*}{\rotatebox{90}{ILI}} 
& 24 & \textbf{3.143} & \textbf{1.185} & \underline{3.483} & \underline{1.287} & 5.764 & 1.677 & 4.480 & 1.444 & 6.026 & 1.770 & 5.914 & 1.734 & 6.624 & 1.830\\
& 36 & \textbf{2.793} & \textbf{1.054} & \underline{3.103} & \underline{1.148} & 4.755 & 1.467 & 4.799 & 1.467 & 5.340 & 1.668 & 6.631 & 1.845 & 6.858 & 1.879\\
& 48 & \underline{2.845} & \underline{1.090} & \textbf{2.669} & \textbf{1.085} & 4.763 & 1.469 & 4.800 & 1.468 & 6.080 & 1.787 & 6.736 & 1.857 & 6.968 & 1.892\\
& 60 & \underline{2.957} & \textbf{1.124} & \textbf{2.770} & \underline{1.125} & 5.264 & 1.564 & 5.278 & 1.560 & 5.548 & 1.720 & 6.870 & 1.879 & 7.127 & 1.918\\
\bottomrule[1.3pt]
\end{tabular}
\begin{tablenotes}
\item[1] Reported metrics except Preformer all come from the Autoformer paper \cite{autoformer}.
\end{tablenotes}
\end{threeparttable}
\end{scriptsize}
\end{center}
\vskip -0.1in
\end{table*}

\subsection{Complexity Analysis}
For Single-Scale Segment-Correlation, if the segment length $L_{seg}=L_0$, the computational complexity is $\mathcal{O}(L^2/L_{0})$. For Multi-Scale Segment-Correlation, the computational complexity is the sum of all scales, which is still at the same computational level as $\mathcal{O}(L^2/L_{0})$ benefiting from the exponential increase in segment length. We demonstrate this in \cref{sec:msscproof} due to space limitations.

Most other sparse mechanisms introduce some extra operations, such as selections of dominant queries in Informer and fast Fourier transform in Autoformer. Although their theoretical complexity is $\mathcal{O}(L\log L)$, the actual operation efficiency is not even as good as our Multi-Scale Segment-Correlation. Please see \cref{sec:eff} for more details.
\section{Experiments}

\subsection{Experimental Setup}

\begin{table*}[ht]
\renewcommand\arraystretch{1}  
\caption{Univariate time-series forecasting results on two typical datasets}
\label{tab:univariate}
\vskip 0.15in
\begin{center}
\begin{scriptsize}
\begin{threeparttable}
\setlength{\tabcolsep}{1.7mm}{
\begin{tabular}{cc|cc|cc|cc|cc|cc|cc|cc|cc} 
\toprule[1.3pt]
\multicolumn{2}{c}{Models} & \multicolumn{2}{c}{Preformer} & \multicolumn{2}{c}{Autoformer} & \multicolumn{2}{c}{N-BEATS} & \multicolumn{2}{c}{Informer} & \multicolumn{2}{c}{LogTrans} &\multicolumn{2}{c}{DeepAR} & \multicolumn{2}{c}{Prophet} & \multicolumn{2}{c}{ARIMA}\\
\cmidrule(lr){3-4}\cmidrule(lr){5-6}\cmidrule(lr){7-8}\cmidrule(lr){9-10}\cmidrule(lr){11-12}\cmidrule(lr){13-14}\cmidrule(lr){15-16}\cmidrule(lr){17-18}
\multicolumn{2}{c}{Metric} & \multicolumn{1}{c}{MSE} & \multicolumn{1}{c}{MAE} & \multicolumn{1}{c}{MSE} & \multicolumn{1}{c}{MAE} & \multicolumn{1}{c}{MSE} & \multicolumn{1}{c}{MAE} & \multicolumn{1}{c}{MSE} & \multicolumn{1}{c}{MAE} & \multicolumn{1}{c}{MSE} & \multicolumn{1}{c}{MAE} & \multicolumn{1}{c}{MSE} & \multicolumn{1}{c}{MAE} & \multicolumn{1}{c}{MSE} & \multicolumn{1}{c}{MAE} & \multicolumn{1}{c}{MSE} & \multicolumn{1}{c}{MAE}\\
\midrule
\midrule
\multirow{4}{*}{\rotatebox{90}{ETTm2}} 
& 96 & \underline{0.072} & \underline{0.205} & \textbf{0.065} & \textbf{0.189} & 0.082 & 0.219 & 0.088 & 0.225 & 0.082 & 0.217 & 0.099 & 0.237 & 0.287 & 0.456 & 0.211 & 0.362\\
& 192 & \textbf{0.109} & \textbf{0.255} & \underline{0.118} & \underline{0.256} & 0.120 & 0.268 & 0.132 & 0.283 & 0.133 & 0.284 & 0.154 & 0.310 & 0.312 & 0.483 & 0.261 & 0.406\\
& 336 & \textbf{0.139} & \textbf{0.295} & \underline{0.154} & \underline{0.305} & 0.226 & 0.370 & 0.180 & 0.336 & 0.201 & 0.361 & 0.277 & 0.428 & 0.331 & 0.474 & 0.317 & 0.448\\
& 720 & \textbf{0.167} & \textbf{0.327} & \underline{0.182} & \underline{0.335} & 0.188 & 0.338 & 0.300 & 0.435 & 0.268 & 0.407 & 0.332 & 0.468 & 0.534 & 0.593 & 0.366 & 0.487\\
\midrule
\multirow{4}{*}{\rotatebox{90}{Exchange}} 
& 96 & \underline{0.141} & \underline{0.293} & 0.241 & 0.387 & 0.156 & 0.299 & 0.591 & 0.615 & 0.279 & 0.441 & 0.417 & 0.515 & 0.828 & 0.762 & \textbf{0.112} & \textbf{0.245}\\
& 192 & \textbf{0.252} & \textbf{0.389} & \underline{0.273} & \underline{0.403} & 0.669 & 0.665 & 1.183 & 0.912 & 1.950 & 1.048 & 0.813 & 0.735 & 0.909 & 0.974 & 0.304 & 0.404\\
& 336 & \textbf{0.426} & \textbf{0.512} & \underline{0.508} & \underline{0.539} & 0.611 & 0.605 & 1.367 & 0.984 & 2.438 & 1.262 & 1.331 & 0.962 & 1.304 & 0.988 & 0.736 & 0.598\\
& 720 & \textbf{0.790} & \textbf{0.712} & \underline{0.991} & \underline{0.768} & 1.111 & 0.860 & 1.872 & 1.072 & 2.010 & 1.247 & 1.894 & 1.181 & 3.238 & 1.566 & 1.871 & 0.935
\\
\bottomrule[1.3pt]
\end{tabular}}
\begin{tablenotes}
\item[1] Reported metrics except Preformer all come from the Autoformer paper \cite{autoformer}.
\end{tablenotes}
\end{threeparttable}
\end{scriptsize}
\end{center}
\vskip -0.1in
\end{table*}

\textbf{Datasets}\quad
We conduct experiments on the following six datasets as in \citet{autoformer}. (1) {\em ETT} \footnote{\url{https://github.com/zhouhaoyi/ETDataset.}} contains data related to electricity which is collected from two Chinese stations in two years. In order to explore the model's performance on data with different granularities, we use different sampling frequencies to get hourly data $\{${\em ETTh1, ETTh2}$\}$ and 15-minutes data $\{${\em ETTm1, ETTm2}$\}$.  (2) {\em Electricity} \footnote{\url{https://archive.ics.uci.edu/ml/datasets/ElectricityLoadDiagrams20112014.}} contains the hourly electricity consumption of 321 clients in 2 years. (3) {\em Exchange} \cite{LSTNet} collects the daily exchange rates of eight countries ranging from 1990 to 2016. (4) {\em Traffic} \footnote{\url{http://pems.dot.ca.gov/.}} collects the hourly road occupancy rates from the California Department of Transportation in two years. (5) {\em Weather} \footnote{\url{https://www.bgc-jena.mpg.de/wetter/.}} records climatological data of the Max-Planck-Institute every 10 minutes in 2020 year, which contains 21 climate features including air pressure and temperature etc. (6) {\em ILI} \footnote{\url{https://gis.cdc.gov/grasp/fluview/fluportaldashboard.html.}} collects the weekly outpatient visits for influenza-like illness (ILI) from 2002 to 2021. We split the datasets following \citet{autoformer}. The train/val/test contains 12/4/4 months of data for the ETT datasets, while we split other datasets into train/val/test by the ratio of 7:1:2.

\textbf{Implementation details}\quad
We use the ADAM \cite{adam} optimizer with an initial learning rate 1e-4 and the learning rate decay to train our model. Early stop training strategy is utilized to avoid overfitting. The number of training epochs is set to 10 and the batch size is set to 32. All experiments are implemented with PyTorch \cite{pytorch} and conducted on single NVIDIA TITAN RTX 24GB GPU. For all experimental settings, Preformer contains 2 encoder layers and 1 decoder layer. 

\textbf{Baselines}\quad
Several models are selected to compare with Preformer in multivariate time-series forecasting, including three transformer-based models: Autoformer \cite{autoformer}, Informer \cite{informer}, LogTrans \cite{ConvTransformer}, two RNN-based models: LSTM \cite{LSTM}, LSTNet \cite{LSTNet} and one TCN-based model: TCN \cite{TCN}. Since there are some models designed specifically for univariate forecasting, we compare Preformer with more competitive baselines, including two deep learning models: N-BEATS \cite{nbeats} and DeepAR \cite{deepar}, two traditional meaching learning methods: Prophet \cite{Prophet} and ARMIA \cite{ARIMA}.

\subsection{Main Results}

To comprehensively compare Preformer and baselines, we conduct thorough multivariate and univariate forecasting experiments on various datasets under multiple settings. For the ILI dataset, the input length is fixed to 24 and the prediction length includes $\{24, 36, 48, 60\}$. For other datasets, the input length is fixed to 96 and the prediction length is chosen from $\{96, 192, 336, 720\}$.

\textbf{Multivariate Time-series Forecasting}\quad
From \cref{tab:multivariate}, we find that Preformer is better than other models except Autoformer in all cases and performs slightly worse than Autoformer in only a few settings. For example, under the input-96-predict-192 setting, compared to previous state-of-the-art results, Preformer has achieved \bm{$4.3\%$} ($0.281 \to 0.269$) relative MSE improvement in ETTm2, \bm{$14.9\%$} ($0.222\to0.189$) in Electricity, \bm{$10.7\%$} ($0.300\to0.268$) in Exchange, \bm{$8.3\%$} ($0.616\to0.565$) in Traffic, \bm{$10.4\%$} ($0.307\to0.275$) in Weather, and \bm{$10\%$} ($3.103\to2.793$) in ILI. Moreover, Preformer shows stable performance in cases of long prediction horizons, which shows that it is suitable for long-term forecasting. Besides, the overall performance of Transformer-based models is better than RNN-based models and TCN-based models, proving the potential of the Transformer-based models in time-series forecasting. 

\textbf{Univariate Time-series Forecasting}\quad
We show the results on two typical datasets, i.e., the ETTm2 dataset with obvious periodicity and the Exchange dataset without obvious periodicity in \cref{tab:univariate}. We observe that our Preformer achieves the best results in all long-term forecasting cases whose prediction length is not less than 192. For example, under the input-96-predict-336 setting, compared to previous state-of-the-art results, Preformer has achieved \bm{$9.7\%$} ($0.154 \to 0.139$) relative improvement on MSE in ETTm2 and \bm{$16\%$} ($0.508 \to 0.426$) in Exchange. Also, Autoformer performs best in the input-96-predict-96 setting of the ETTm2 dataset while ARIMA performs best in the same setting of the Exchange dataset, which illustrates their advantages in shorter-term prediction.

\subsection{Ablation Study}

\textbf{Performance of the Segment-Correlation mechanism}\quad
We conduct experiments on the ETTh1 dataset to compare several different sparse attention mechanisms. Segment-correlation in \cref{tab:attention} refers to the Preformer model with the multi-scale structure and predictive paradigm, and the initial segment length $L_0$ is set to 4. For a fair comparison, we replace MSSC and PreMSSC with other sparse attention mechanisms and set hyperparameters such as the hidden dimension and the head numbers to be consistent. From \cref{tab:attention}, we observe that compared with other sparse mechanisms, Segment-Correlation achieves the best performance. Full attention and LogSparse attention fail in long-term forecasting due to high memory complexity. 

\begin{table}[ht]
\caption{Ablations of the Segment-Correlation mechanism}
\label{tab:attention}
\vskip 0.15in
\begin{center}
\begin{scriptsize}
\begin{threeparttable}
\setlength{\tabcolsep}{1.6mm}{
\begin{tabular}{cc|ccc|ccc} 
\toprule[1.3pt]
\multicolumn{2}{c}{Input Length} & \multicolumn{3}{c}{96} & \multicolumn{3}{c}{336}\\
\cmidrule(lr){3-5}\cmidrule(lr){6-8}
\multicolumn{2}{c}{Prediction Length} & 336 & 720 & 1440 & 336 & 720 & 1440 \\
\midrule
\midrule
Segment- & MSE & \textbf{0.628} & \textbf{0.632} & \textbf{0.769} & \textbf{0.560} & \textbf{0.581} & \textbf{0.755}\\
Correlation & MAE & \textbf{0.551} & \textbf{0.571} & \textbf{0.650} & \textbf{0.531} & \textbf{0.558} & \textbf{0.657}\\
\midrule
Auto- & MSE & 0.630 & 0.687 & 0.821 & 0.597 & 0.619 & 0.859\\
Correlation & MAE & 0.552 & 0.593 & 0.671 & 0.539 & 0.578 & 0.692\\
\midrule
Full & MSE & 0.673 & 0.679 & - & 0.661 & 0.678 & -\\
Attention & MAE & 0.564 & 0.585 & - & 0.566 & 0.592 & -\\
\midrule
LogSparse & MSE & 0.679 & 0.688 & - & 0.646 & 0.668 & -\\
Attention & MAE & 0.567 & 0.591 & - & 0.565 & 0.592 & -\\
\midrule
ProbSparse & MSE & 0.702 & 0.705 & 0.831 & 0.689 & 0.701 & 0.861\\
Attention & MAE & 0.577 & 0.599 & 0.673 & 0.579 & 0.604 & 0.694\\
\bottomrule[1.3pt]
\end{tabular}}
\begin{tablenotes}
\item[1] The "-" indicates the out-of-memory.
\end{tablenotes}
\end{threeparttable}
\end{scriptsize}
\end{center}
\vskip -0.1in
\end{table}

\textbf{Impact of the multi-scale structure}\quad
The multi-scale structure can effectively extract dependencies at different temporal resolutions, which is important for time series forecasting. To illustrate this, we remove the multi-scale structure from all PreMSSC and MSSC modules in Preformer to get the model without multi-scale structure. As shown in \cref{tab:ablationMSPre}, the prediction performance of Preformer with the multi-scale structure ({\em \textbf{MS+Predictive}}) is better than that without the multi-scale structure ({\em \textbf{Only Predictive}}) in all cases. Especially on the ETTh1 and Exchange datasets, the multi-scale structure can lead to considerable performance improvements.

\textbf{Impact of the predictive paradigm}\quad
To explore whether the predictive paradigm is really effective for prediction tasks, we conduct ablation studies of the predictive paradigm on three datasets with different settings. As shown in \cref{tab:ablationMSPre}, {\em \textbf{MS+predictive}} represents the standard Preformer model, while {\em \textbf{Only MS}} means that replacing the PreMSSC module in Preformer's decoder with the MSSC module. The experimental results show that Preformer with the predictive paradigm achieves better performance in almost all cases, which proves that the proposed predictive paradigm is helpful for the prediction tasks.

\begin{table}[ht]
\caption{Ablations of predictive paradigm and multi-scale structure}
\label{tab:ablationMSPre}
\vskip 0.15in
\begin{center}
\begin{scriptsize}
\begin{tabular}{cc|cc|cc|cc}
\toprule[1.3pt]
\multicolumn{2}{c}{\multirow{2}{*}{Models}}
& \multicolumn{2}{c}{{\em \textbf{MS+Predictive}}} & \multicolumn{2}{c}{{\em Only Predictive}} & \multicolumn{2}{c}{{\em Only MS}}
\\
\multicolumn{2}{c}{} & \multicolumn{2}{c}{(\textbf{ours})} & \multicolumn{2}{c}{(without MS)} & \multicolumn{2}{c}{(without Predictive)}
\\
\cmidrule(lr){3-4}\cmidrule(lr){5-6}\cmidrule(lr){7-8}
\multicolumn{2}{c}{Metric} & \multicolumn{1}{c}{MSE} & \multicolumn{1}{c}{MAE} & \multicolumn{1}{c}{MSE} & \multicolumn{1}{c}{MAE} & \multicolumn{1}{c}{MSE} & \multicolumn{1}{c}{MAE}  
\\
\midrule
\midrule
\multirow{4}{*}{\rotatebox{90}{ETTh1}} 
& 96 & \textbf{0.414} & 0.439 & 0.480 & 0.472 & 0.416 & \textbf{0.436} \\
& 192 & \textbf{0.445} & \textbf{0.455} & 0.493 & 0.484 & 0.472 & 0.472 \\
& 336 & \textbf{0.466} & \textbf{0.468} & 0.511 & 0.497 & 0.471 & 0.475\\
& 720 & \textbf{0.471} & \textbf{0.487} & 0.605 & 0.553 & 0.484 & 0.499\\
\midrule
\multirow{4}{*}{\rotatebox{90}{Exchange}} 
& 96 & \textbf{0.148} & \textbf{0.282} & 0.186 & 0.305 & 0.187 & 0.305\\
& 192 & \textbf{0.268} & \textbf{0.378} & 0.359 & 0.457 & 0.280 & 0.386\\
& 336 & \textbf{0.447} & \textbf{0.499} & 0.704 & 0.659 & 0.520 & 0.542\\
& 720 & \textbf{1.092} & \textbf{0.812} & 1.400 & 0.931 & 1.232 & 0.883\\
\midrule
\multirow{4}{*}{\rotatebox{90}{Traffic}} 
& 96 & \textbf{0.560} & \textbf{0.349} & 0.561 & 0.352 & 0.567 & 0.351\\
& 192 & \textbf{0.565} & \textbf{0.349} & 0.573 & 0.356 & 0.583 & 0.358\\
& 336 & \textbf{0.577} & \textbf{0.351} & \textbf{0.577} & 0.353 & 0.581 & 0.354\\
& 720 & \textbf{0.597} & \textbf{0.358} &0.599 & 0.363 & 0.598 & 0.362\\
\bottomrule[1.3pt]
\end{tabular}
\end{scriptsize}
\end{center}
\vskip -0.1in
\end{table}

\begin{figure}[t]
\vskip 0.2in
\begin{center}
    \subfigure[MSE scores]{
    \includegraphics[width=0.465\columnwidth]{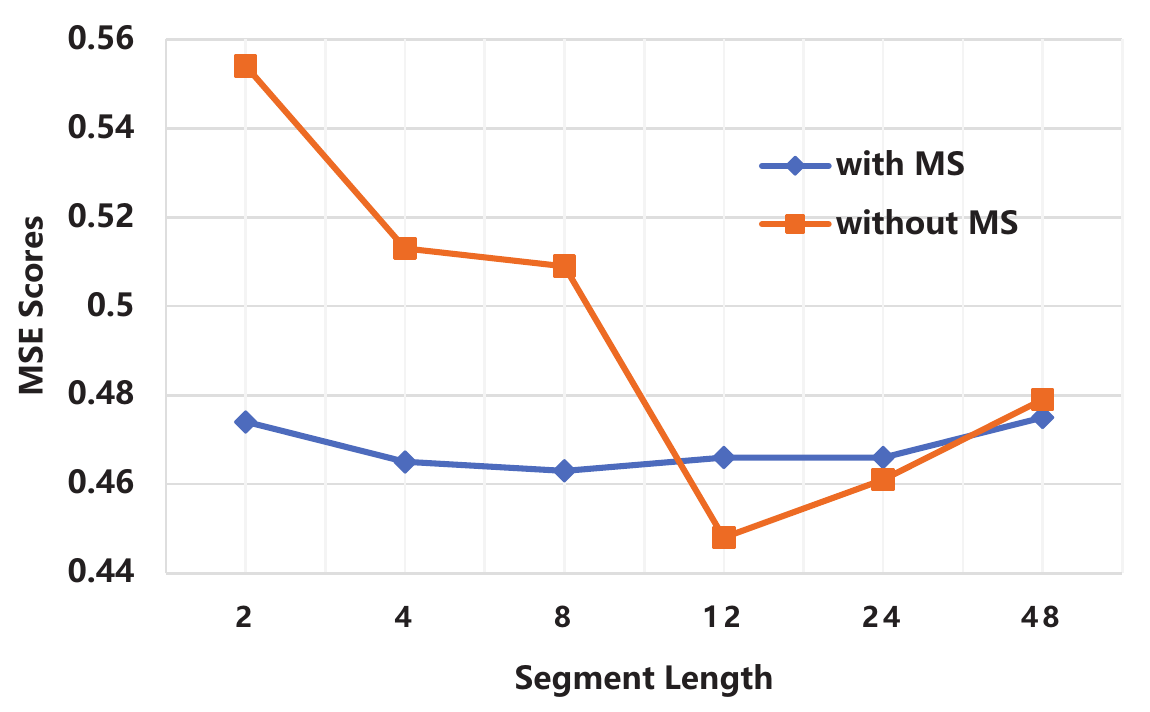}
    }
    \subfigure[MAE scores]{
    \includegraphics[width=0.475\columnwidth]{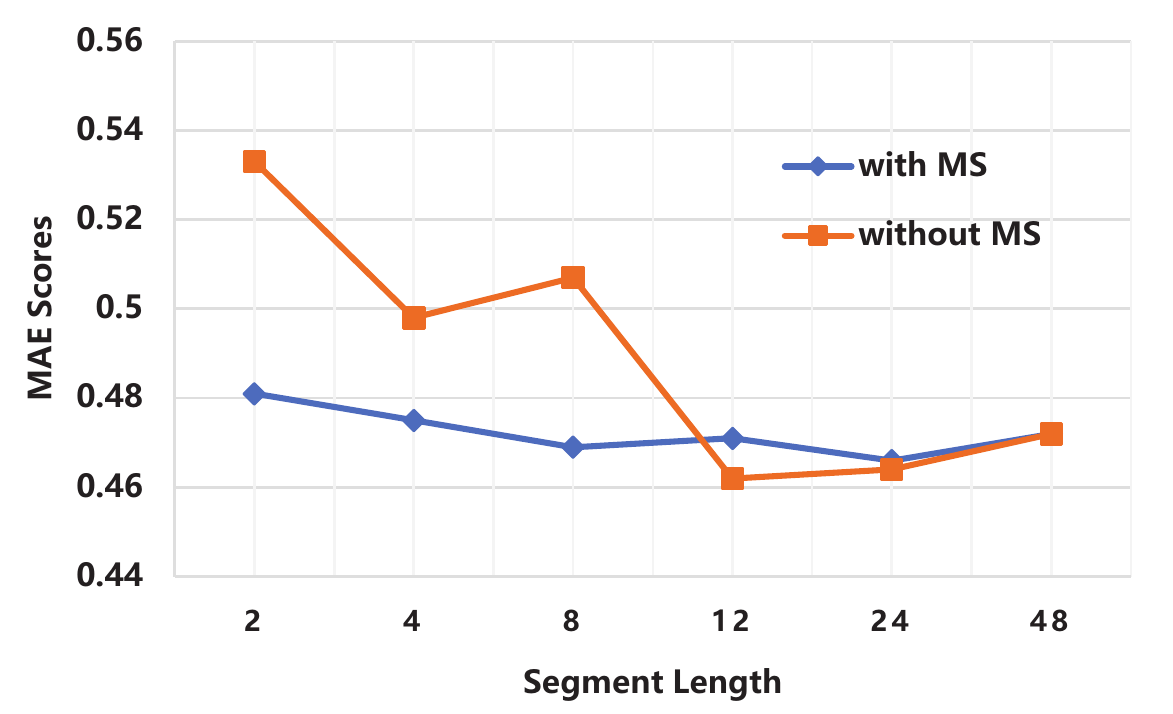}
    }
  \caption{Effects of the multi-scale structure on sensitivity of segment length hyperparameter.}
\label{fig:segmentlength}
\end{center}
\vskip -0.2in
\end{figure}

\textbf{The sensitivity analysis of the segment length}\quad
The segment length $L_{seg}$ is a critical hyperparameter in Segment-Correlation. The multi-scale structure can facilitate the selection of segment length. We conduct multivariate forecasting experiments (input-96 to predict-336) on the ETTh1 dataset to explore the sensitivity of segment length using our Preformer with or without the multi-scale structure. As shown in \cref{fig:segmentlength}, Preformer without the multi-scale structure is very sensitive to the choice of segment length, while the performance of Preformer with the multi-scale structure does not change much with the segment length. More detailed results can be found in \cref{tab:segmentlength}. These results all demonstrate the effectiveness of the multi-scale structure.

\subsection{Efficiency analysis}
\label{sec:eff}

\begin{figure}[ht]
\vskip 0.2in
\begin{center}
    \subfigure[Memory efficiency]{
    \includegraphics[width=0.47\columnwidth]{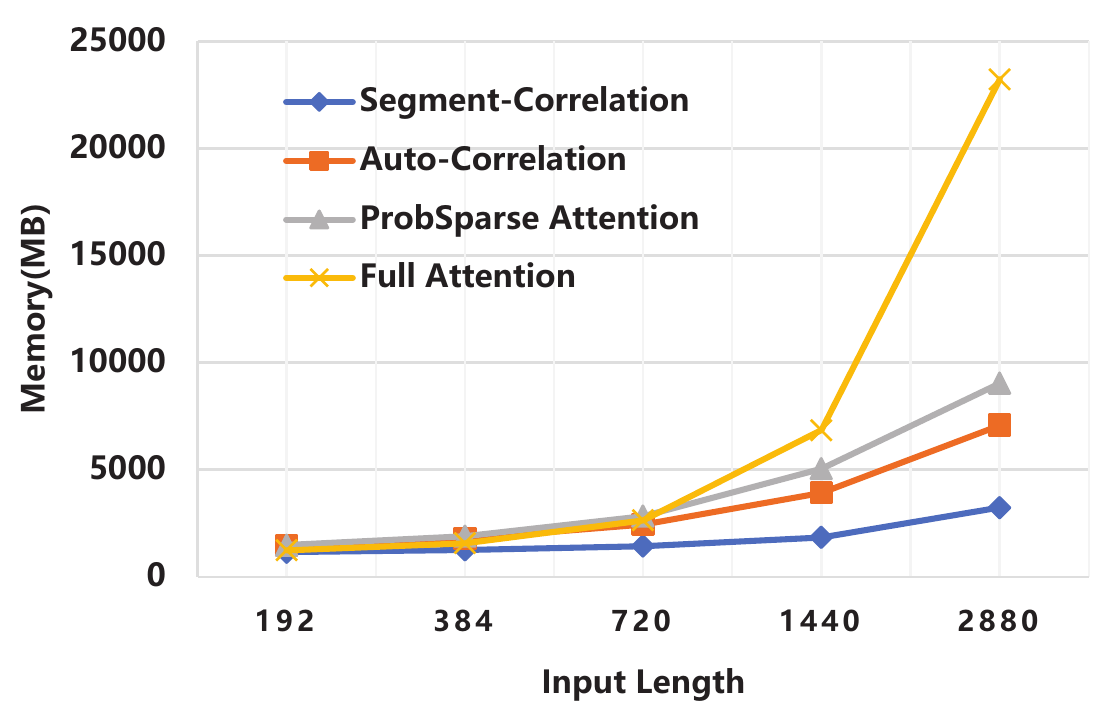}
    }
    \subfigure[Time efficiency]{
    \includegraphics[width=0.47\columnwidth]{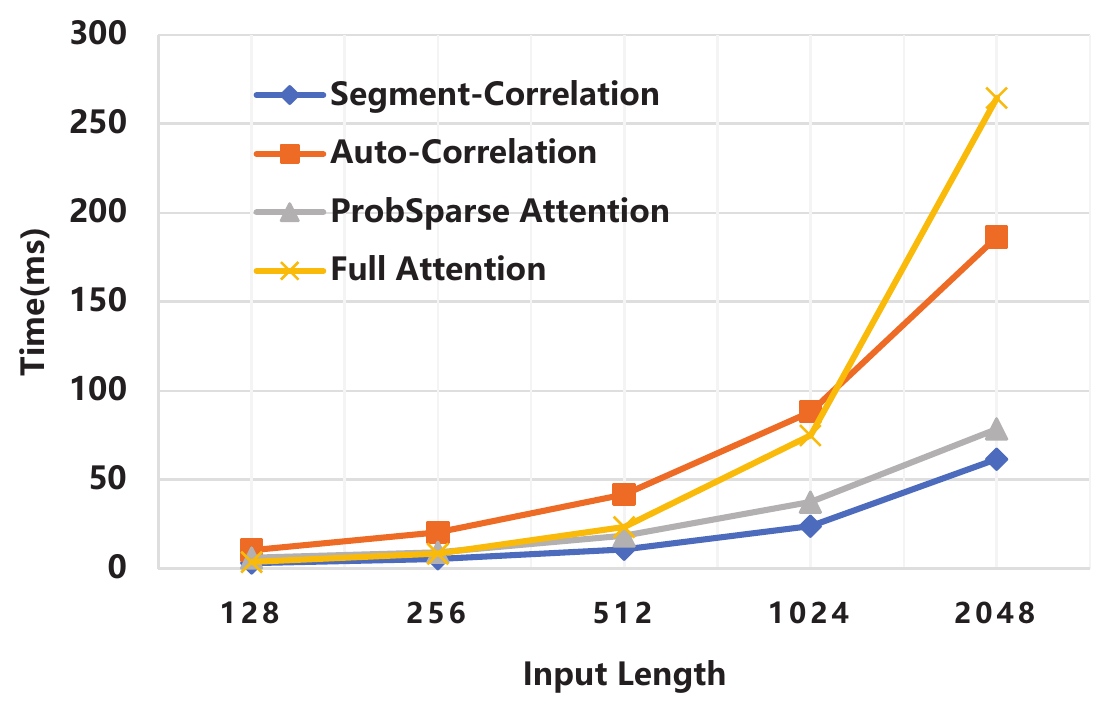}
    }
  \caption{Memory and time consumption during the training phase.}
\label{fig:eff}
\end{center}
\vskip -0.2in
\end{figure}

As shown in \cref{fig:eff}, we compare the running memory and time of models with different sparse attention mechanisms during the training phase. Segment-Correlation in the figure means MSSC with the multi-scale structure and $L_0$ is set to 4. For memory efficiency analysis, we set the batch size to 16 in order to avoid out-of-memory. For time efficiency analysis, we run each attention mechanism 1000 times to get the average training time. It is worth noting that our Multi-Scale Segment-Correlation is more efficient in time and memory than other sparse mechanisms whose theoretical complexity $\mathcal{O}(L\log L)$ in practical applications.

\section{Conclusion}
In this paper, we propose a Transformer-based model called Preformer for long-term time series forecasting. In Preformer, we introduce a sparse and efficient attention mechanism called Multi-Scale Segment-Correlation (MSSC) which utilizes correlations between segment pairs to discover dependencies and aggregate information in time series. Further, we design a predictive paradigm and combine it with MSSC to get Predictive Multi-Scale Segment-Correlation (PreMSSC), which can discover predictive dependencies from contexts. Under various experimental settings in different datasets, Preformer with MSSC and PreMSSC can yield state-of-the-art prediction performance, which demonstrates the effectiveness of our Preformer.



\bibliography{example_paper}
\bibliographystyle{icml2022}

\newpage
\appendix
\onecolumn
\section{Evalution Metrics}
On each prediction window, we use two metrics which are commonly used in prediction task to evaluate performances, i.e., Mean Square Error (MSE) and Mean Absolute Error (MAE). They are defined as follows: 
\begin{equation}
\begin{aligned}
\textup{MSE} &= \frac{1}{n}\sum_{i=1}^{n}(y_i-\hat y_i)^2,\\
\textup{MAE} &= \frac{1}{n}\sum_{i=1}^{n}|y_i-\hat y_i|,
\end{aligned}
\end{equation}
where $y_i$ is ground truth and $\hat{y}_i$ is prediction result. For multivariate prediction, the metrics can be calculated and averaged on each single variable.

\section{The Computational Complexity of Multi-Scale Segment-Correlation}
\label{sec:msscproof}
The overall computational complexity of the MSSC moduel is the sum of all scales, which can be formulated as:
\begin{equation}
\begin{split}
    \mathcal{O}(MSSC) &= \mathcal{O}(\frac{L^2}{2^0 L_{0}}) + \mathcal{O}(\frac{L^2}{2^1 L_{0}}) + \dots + \mathcal{O}(\frac{L^2}{2^{\lfloor \log_2(\frac{L}{L_0}) \rfloor}} L_{0})\\
    &= \left[2- (\frac 1 2)^{\lfloor \log_2(\frac{L}{L_0}) \rfloor}\right] \mathcal{O}(\frac{L^2}{L_{0}})\\
    &\leq \left[2-(\frac 1 2)^{\log_2(\frac{L}{L_0})}\right] \mathcal{O}(\frac{L^2}{L_{0}})\\
    &=\left(2-\frac{L_0}{L}\right) \mathcal{O}(\frac{L^2}{L_{0}}).
\end{split}
\end{equation}
We find that the computational complexity of MSSC does not exceed twice the computational complexity of the Segment-Correlation without the multi-scale structure (i.e., Single-Scale Segment-Correlation). 

\section{Sensitivity of the Segment Length}

We explore the impact of the multi-scale structure by comparing the effect of choice at different segment lengths on MSE scores with or without the multi-scale structure. We use the standard deviation of the MSE scores to evaluate the stability of forecasting performance. As shown in \cref{tab:segmentlength}, models with multi-scale structure have more stable performance on all datasets, i.e., the MSE scores vary less with different choices of segment length.

\begin{table}[H]
\renewcommand\arraystretch{1.1}  
\caption{In the cases of the Preformer model with or without multi-scale structure, the MSE scores vary with the segment length on four datasets}
\label{tab:segmentlength}
\vskip 0.15in
\begin{center}
\begin{small}
\setlength{\tabcolsep}{2.5mm}{
\begin{tabular}{cc|cccccc|c} 
\toprule[1.3pt]
\multicolumn{2}{c}{Segment length} & \multicolumn{1}{c}{2} & \multicolumn{1}{c}{4} & \multicolumn{1}{c}{8} & \multicolumn{1}{c}{12} & \multicolumn{1}{c}{24} & \multicolumn{1}{c}{48} & \multicolumn{1}{c}{std ($\times$ 1000)}\\
\midrule
\midrule
\multirow{2}{*}{ETTm2} 
& with MS & 0.326 & 0.327 & 0.326 & 0.327 & 0.327 & 0.327 & \textbf{0.47} \\
& without MS & 0.328 & 0.327 & 0.326 & 0.325 & 0.327 & 0.324 & 1.34\\
\midrule
\multirow{2}{*}{Electricity} 
& with MS & 0.201 & 0.195 & 0.199 & 0.196 & 0.211 & 0.206 & \textbf{5.62}\\
& without MS & 0.207 & 0.200 & 0.198 & 0.200 & 0.216 & 0.207 & 6.16\\
\midrule
\multirow{2}{*}{Traffic} 
& with MS & 0.577 & 0.578 & 0.580 & 0.583 & 0.585 & 0.597 & \textbf{6.7}\\
& without MS & 0.596 & 0.583 & 0.584 & 0.581 & 0.590 & 0.600 & 7.02\\
\midrule
\multirow{2}{*}{Weather} 
& with MS & 0.325 & 0.325 & 0.326 & 0.327 & 0.332 & 0.327 & \textbf{2.38} \\
& without MS & 0.322 & 0.326 & 0.342 & 0.344 & 0.337 & 0.326 & 8.53 \\
\bottomrule[1.3pt]
\end{tabular}}
\end{small}
\end{center}
\vskip -0.1in
\end{table}

\section{Full Benchmark on the ETT Datasets}
We conduct more experiments on the four ETT datasets, and the experimental results are shown in \cref{tab:multivariateETT}. We find that our Preformer achieves state-of-the-art on different forecasting horizon. 
\begin{table}[H]
\renewcommand\arraystretch{1}  
\caption{Multivariate time-series forecasting results on the ETT datasets}
\label{tab:multivariateETT}
\vskip 0.15in
\begin{center}
\begin{footnotesize}
\begin{threeparttable}
\begin{tabular}{cc|cc|cc|cc|cc|cc|cc} 
\toprule[1.3pt]
\multicolumn{2}{c}{Models} & \multicolumn{2}{c}{Preformer} & \multicolumn{2}{c}{Autoformer} & \multicolumn{2}{c}{Informer} & \multicolumn{2}{c}{LogTrans} &\multicolumn{2}{c}{LSTNet} & \multicolumn{2}{c}{LSTMa} \\
\cmidrule(lr){3-4}\cmidrule(lr){5-6}\cmidrule(lr){7-8}\cmidrule(lr){9-10}\cmidrule(lr){11-12}\cmidrule(lr){13-14}
\multicolumn{2}{c}{Metric} & \multicolumn{1}{c}{MSE} & \multicolumn{1}{c}{MAE} & \multicolumn{1}{c}{MSE} & \multicolumn{1}{c}{MAE} & \multicolumn{1}{c}{MSE} & \multicolumn{1}{c}{MAE} & \multicolumn{1}{c}{MSE} & \multicolumn{1}{c}{MAE} & \multicolumn{1}{c}{MSE} & \multicolumn{1}{c}{MAE} & \multicolumn{1}{c}{MSE} & \multicolumn{1}{c}{MAE}\\
\midrule
\midrule
\multirow{5}{*}{\rotatebox{90}{ETTh1}} 
& 24 & \textbf{0.357} & \textbf{0.411} & \underline{0.384} & \underline{0.425} & 0.577 & 0.549 & 0.686 & 0.604 & 1.293 & 0.901 & 0.650 & 0.624\\
& 48 & \textbf{0.378} & \textbf{0.417} & \underline{0.392} & \underline{0.419} & 0.685 & 0.625 & 0.766 & 0.757 & 1.456 & 0.960 & 0.702 & 0.675\\
& 168 & \textbf{0.438} & \textbf{0.455} & \underline{0.490} & \underline{0.481} & 0.931 & 0.752 & 1.002 & 0.846 & 1.997 & 1.214 & 1.212 & 0.867\\
& 336 & \textbf{0.463} & \textbf{0.467} & \underline{0.505} & \underline{0.484} & 1.128 & 0.873 & 1.362 & 0.952 & 2.655 & 1.369 & 1.424 & 0.994\\
& 720 & \textbf{0.474} & \textbf{0.486} & \underline{0.498} & \underline{0.500} & 1.215 & 0.896 & 1.397 & 1.291 & 2.143 & 1.380 & 1.960 & 1.322\\
\midrule
\multirow{5}{*}{\rotatebox{90}{ETTh2}} 
& 24 & \textbf{0.255} & \textbf{0.340} & \underline{0.261} & \underline{0.341} & 0.720 & 0.665 & 0.828 & 0.750 & 2.742 & 1.457 & 1.143 & 0.813 \\
& 48 & \textbf{0.300} & \textbf{0.367} & \underline{0.312} & \underline{0.373} & 1.457 & 1.001 & 1.806 & 1.034 & 3.567 & 1.687 & 1.671 & 1.221\\
& 168 & \textbf{0.398} & \textbf{0.418} & \underline{0.457} & \underline{0.455} & 3.489 & 1.515 & 4.070 & 1.681 & 3.242 & 2.513 & 4.117 & 1.674\\
& 336 & \textbf{0.440} & \textbf{0.457} & \underline{0.471} & \underline{0.475} & 2.723 & 1.340 & 3.875 & 1.763 & 2.544 & 2.591 & 3.434 & 1.549\\
& 720 & \textbf{0.458} & \textbf{0.472} & \underline{0.474} & \underline{0.484} & 3.467 & 1.473 & 3.913 & 1.552 & 4.625 & 3.709 & 3.963 & 1.788\\
\midrule
\multirow{5}{*}{\rotatebox{90}{ETTm1}} 
& 24 & \underline{0.345} & \underline{0.402} & 0.383 & 0.403 & \textbf{0.323} & \textbf{0.369} & 0.419 & 0.412 & 1.968 & 1.170 & 0.621 & 0.629\\
& 48 & \textbf{0.422} & \textbf{0.427} & \underline{0.454} & \underline{0.453} & 0.494 & 0.503 & 0.507 & 0.583 & 1.999 & 1.215 & 1.392 & 0.939\\
& 96 & \textbf{0.443} & \textbf{0.450} & \underline{0.481} & \underline{0.463} & 0.678 & 0.614 & 0.768 & 0.792 & 2.762 & 1.542 & 1.339 & 0.913\\
& 288 & \textbf{0.502} & \textbf{0.493} & \underline{0.634} & \underline{0.528} & 1.056 & 0.786 & 1.462 & 1.320 & 1.257 & 2.076 & 1.740 & 1.124\\
& 672 & \textbf{0.583} & \textbf{0.538} & \underline{0.606} & \underline{0.542} & 1.192 & 0.926 & 1.669 & 1.461 & 1.917 & 2.941 & 2.736 & 1.555\\
\midrule
\multirow{5}{*}{\rotatebox{90}{ETTm2}} 
& 24 & \textbf{0.143} & \textbf{0.252} & \underline{0.153} & \underline{0.261} & 0.173 & 0.301 & 0.211 & 0.332 & 1.101 & 0.831 & 0.580 & 0.572\\
& 48 & \textbf{0.171} & \textbf{0.272} & \underline{0.178} & \underline{0.280} & 0.303 & 0.409 & 0.427 & 0.487 & 2.619 & 1.393 & 0.747 & 0.630\\
& 96 & \textbf{0.213} & \textbf{0.295} & \underline{0.255} & \underline{0.339} & 0.365 & 0.453 & 0.768 & 0.642 & 3.142 & 1.365 & 2.041 & 1.073\\
& 288 & \textbf{0.309} & \textbf{0.353} & \underline{0.342} & \underline{0.378} & 1.047 & 0.804 & 1.090 & 0.806 & 2.856 & 1.329 & 0.969 & 0.742\\
& 672 & \textbf{0.407} & \textbf{0.409} & \underline{0.434} & \underline{0.430} & 3.126 & 1.302 & 2.397 & 1.214 & 3.409 & 1.420 & 2.541 & 1.239\\
\midrule
\bottomrule[1.3pt]
\end{tabular}
\begin{tablenotes}
\item[1] Reported metrics except Preformer all come from the Autoformer paper \cite{autoformer}.
\end{tablenotes}
\end{threeparttable}
\end{footnotesize}
\end{center}
\vskip -0.1in
\end{table}

\section{Dataset Statistics and Hyperparameters}
\cref{tab:datasets} lists the length, granularity and series number of all datasets. And optimal hyperparameter settings for the models on these datasets are also included, where $d_{model}$, $d_{ff}$, $n_{heads}$, $L_0$, $e_{layers}$, $d_{layers}$ represent the dimension of hidden representation, the dimension of hidden state in the FFN, the initial segment length, the number of encoder layers, the number of decoder layers respectively.
\begin{table}[H]
\renewcommand\arraystretch{1.1}  
\caption{Details and model hyperparameters of all the datasets}
\label{tab:datasets}
\vskip 0.15in
\begin{center}
\begin{footnotesize}
\begin{tabular}{c|ccccccccc} 
\toprule[1.3pt]
\multicolumn{1}{c}{Datasets} & \multicolumn{1}{c}{ETTh1} & \multicolumn{1}{c}{ETTm1} & \multicolumn{1}{c}{ETTh2} & \multicolumn{1}{c}{ETTm2} & \multicolumn{1}{c}{Electricity} & \multicolumn{1}{c}{Exchange} & \multicolumn{1}{c}{Traffic} & \multicolumn{1}{c}{Weather} & \multicolumn{1}{c}{ILI}\\
\midrule
\midrule
\underline{\textbf{Dataset Statistics}} \\[0.23cm]
series number & 7 & 7 & 7 & 7 & 321 & 8 & 862 & 21 & 7 \\
length & 17420 & 69680 & 17420 & 69680 & 26304 & 7588 & 17544 & 52696 & 966\\
sample rate & 1 hour & 15 mins & 1 hour & 15 mins & 1 hour & 1 day & 1 hour & 10 mins & 1 week\\
\midrule
\underline{\textbf{Model Parameters}} \\[0.23cm]
$d_{model}$ & 64 & 512 & 64 & 8 & 512 & 512 & 512 & 8 & 512\\
$d_{ff}$ & 256 & 2048 & 256 & 32 & 2048 & 2048 & 2048 & 32 & 2048\\
$n_{heads}$ & 8 & 8 & 8 & 1 & 8 & 8 & 8 & 1 & 8\\
$L_0$ & 3 & 6 & 3 & 16 & 4 & 4 & 4 & 4 & 3 \\
$e_{layers}$ & 2 & 2 & 2 & 2 & 2 & 2 & 2 & 2 & 2 \\
$d_{layers}$ & 1 & 1 & 1 & 1 & 1 & 1 & 1 & 1 & 1\\
\bottomrule[1.3pt]
\end{tabular}
\end{footnotesize}
\end{center}
\vskip -0.1in
\end{table}

\clearpage
\section{Visualization of Multivariate Time-Series Forecasting}
\subsection{Comparison of Transformer-based Models}
\begin{figure}[H]
\vskip 0.2in
\begin{center}
    \subfigure[Preformer]{
    \includegraphics[width=0.22\columnwidth]{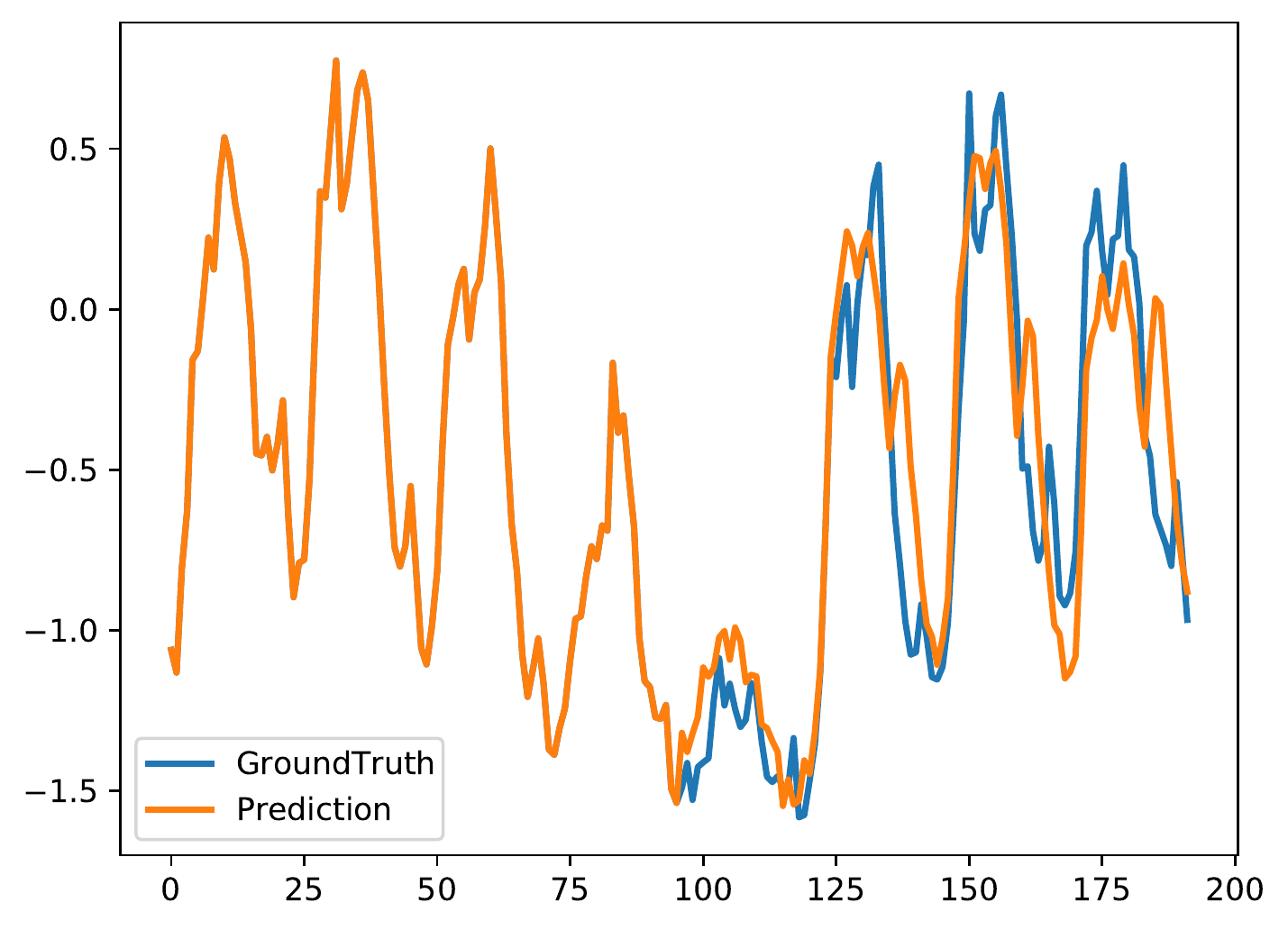}
    }
    \subfigure[Autoformer]{
    \includegraphics[width=0.22\columnwidth]{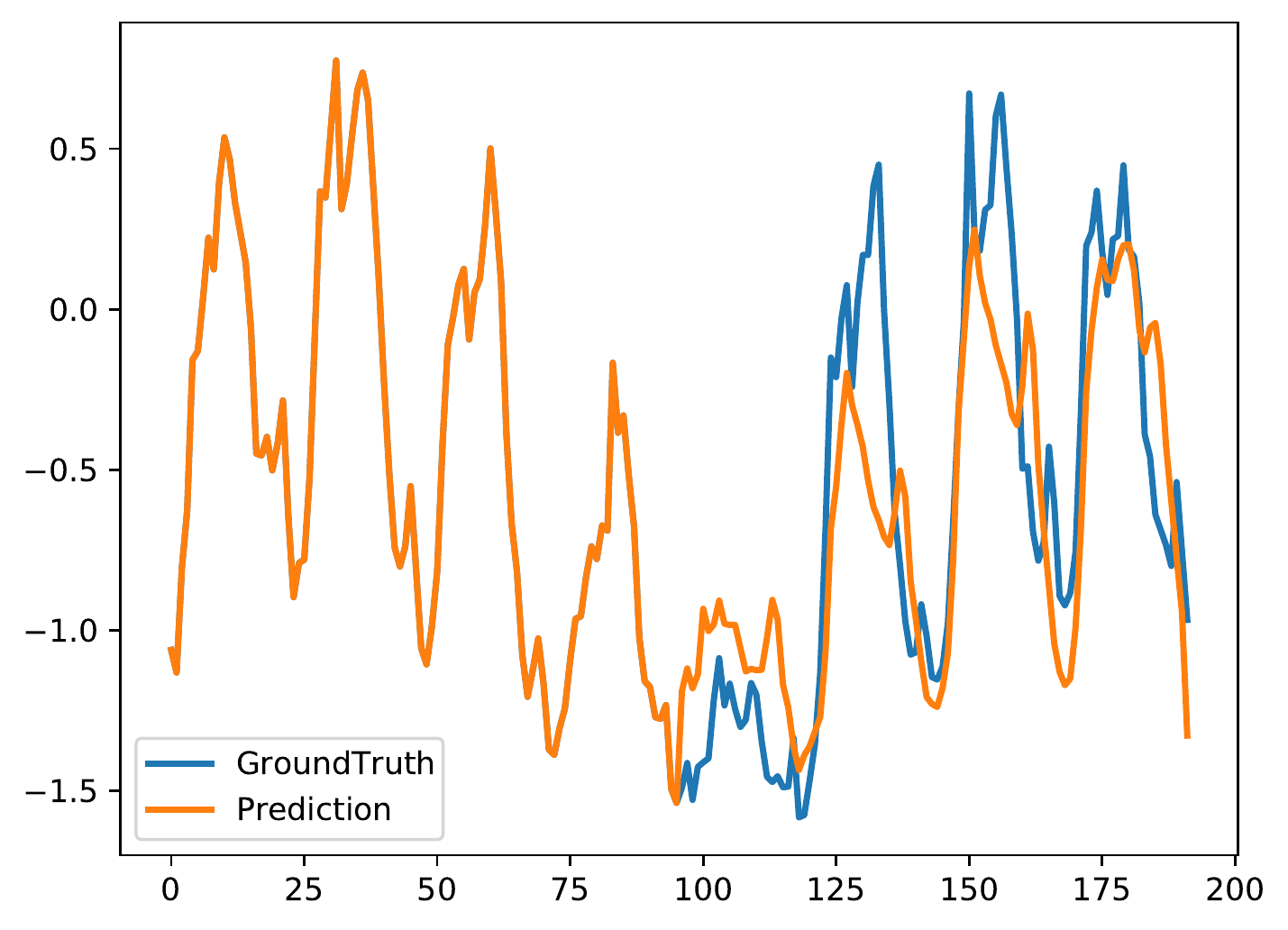}
    }
    \subfigure[Informer]{
    \includegraphics[width=0.22\columnwidth]{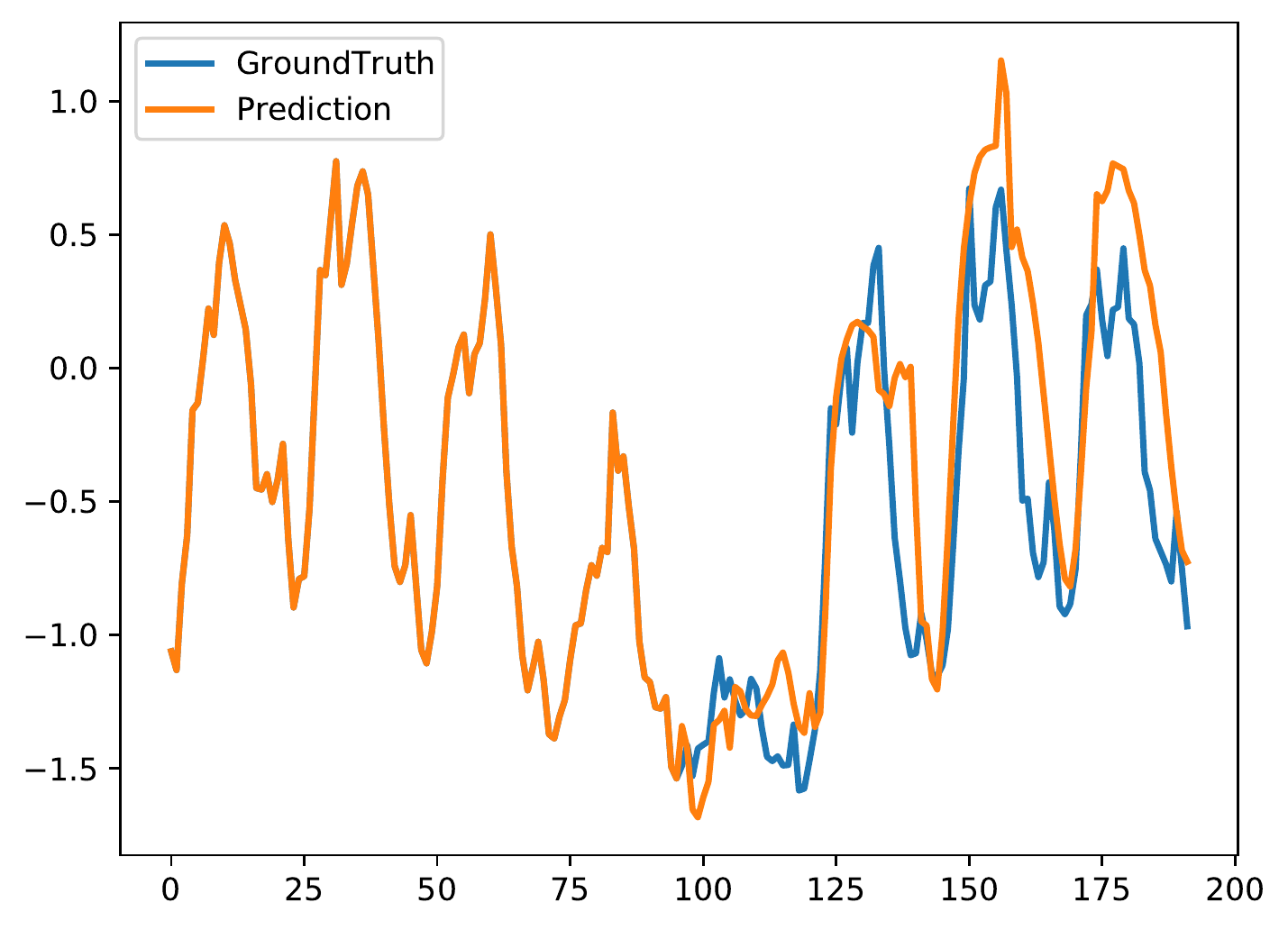}
    }
    \subfigure[LogTrans]{
    \includegraphics[width=0.22\columnwidth]{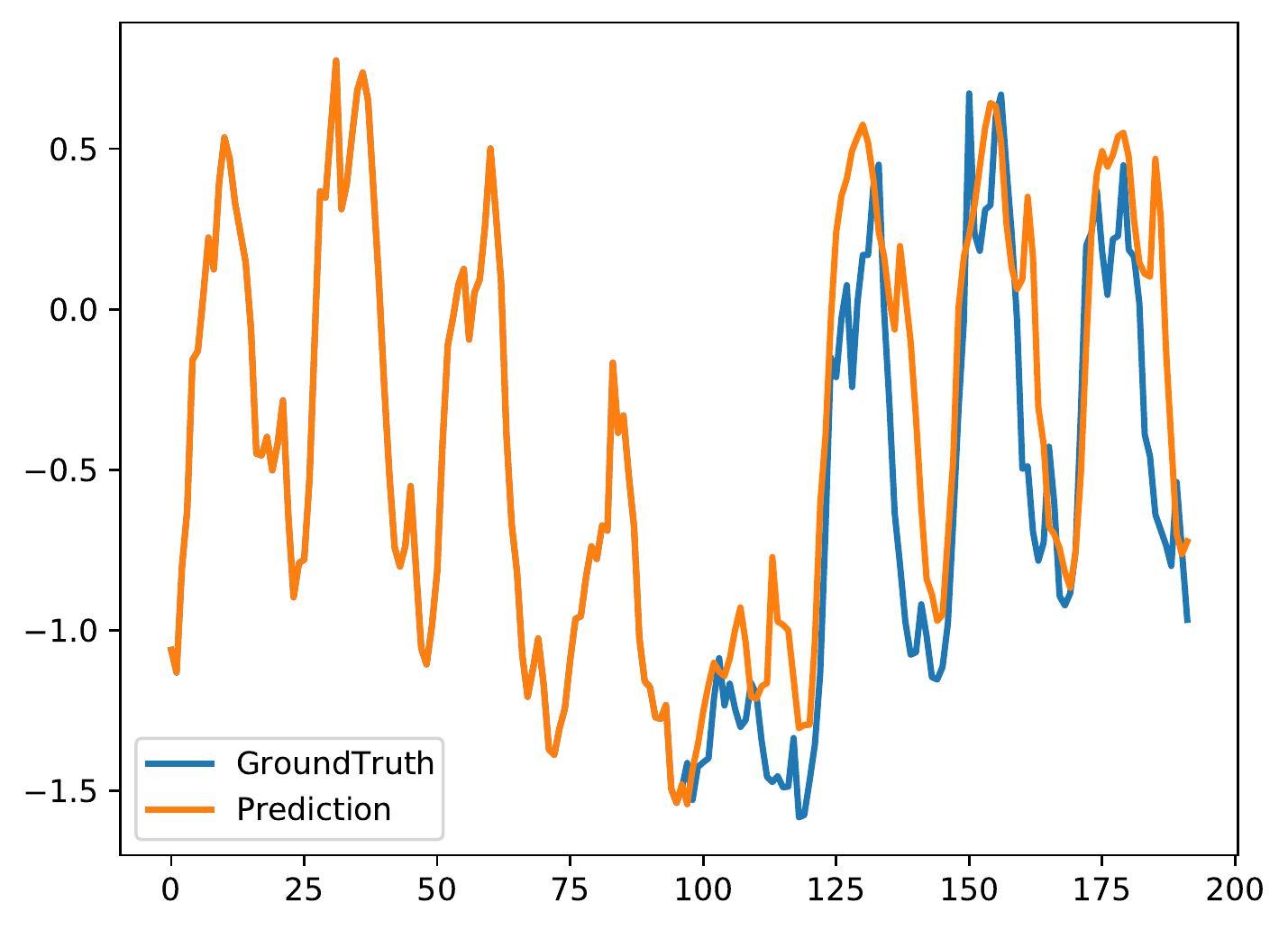}
    }
  \caption{The prediction results on the Electricity dataset under the input-96-predict-96 setting. }
  \label{fig:6}
\end{center}
\vskip -0.2in
\end{figure}

\begin{figure}[H]
\vskip 0.2in
\begin{center}
    \subfigure[Preformer]{
    \includegraphics[width=0.22\columnwidth]{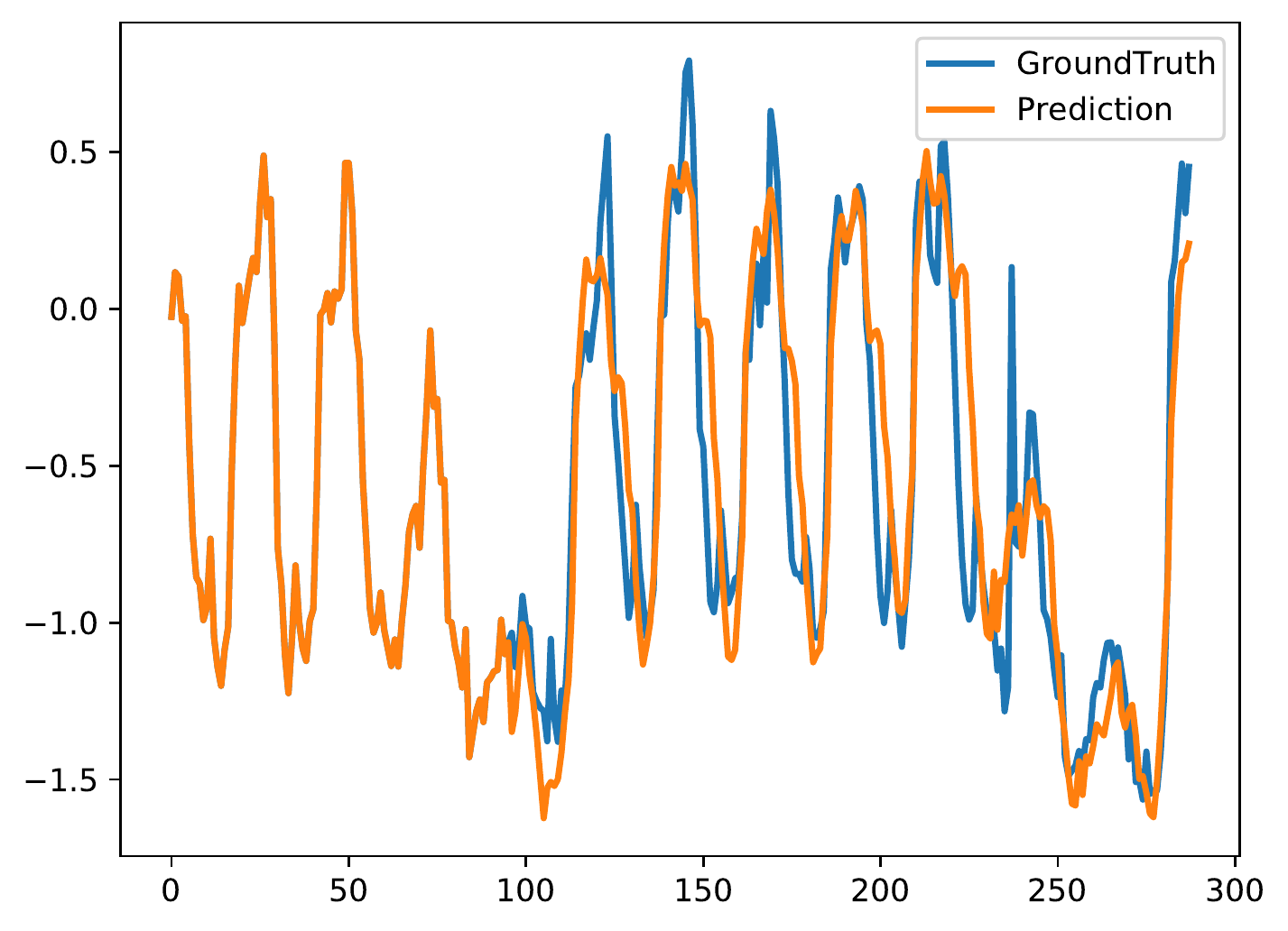}
    }
    \subfigure[Autoformer]{
    \includegraphics[width=0.22\columnwidth]{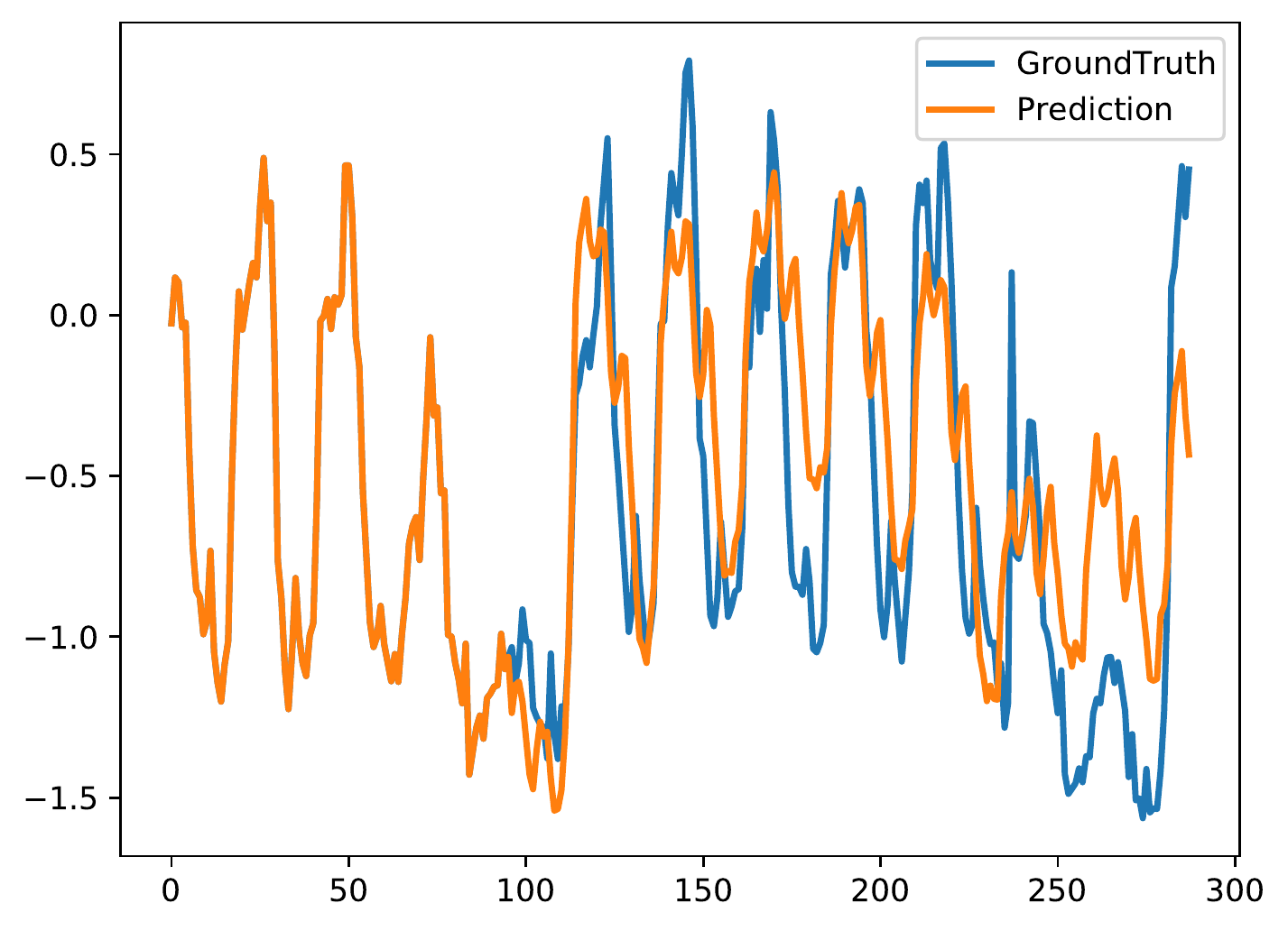}
    }
    \subfigure[Informer]{
    \includegraphics[width=0.22\columnwidth]{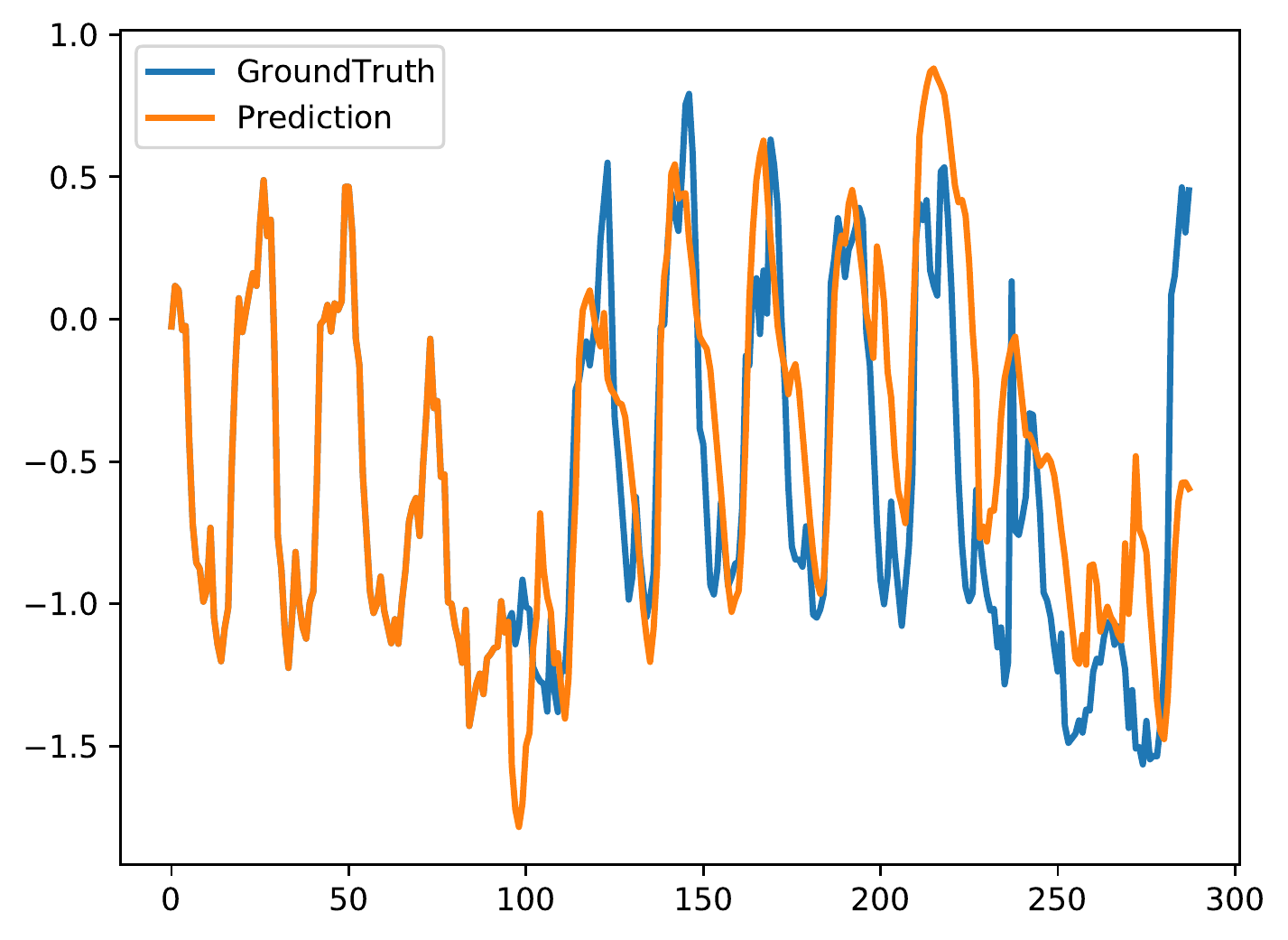}
    }
    \subfigure[LogTrans]{
    \includegraphics[width=0.22\columnwidth]{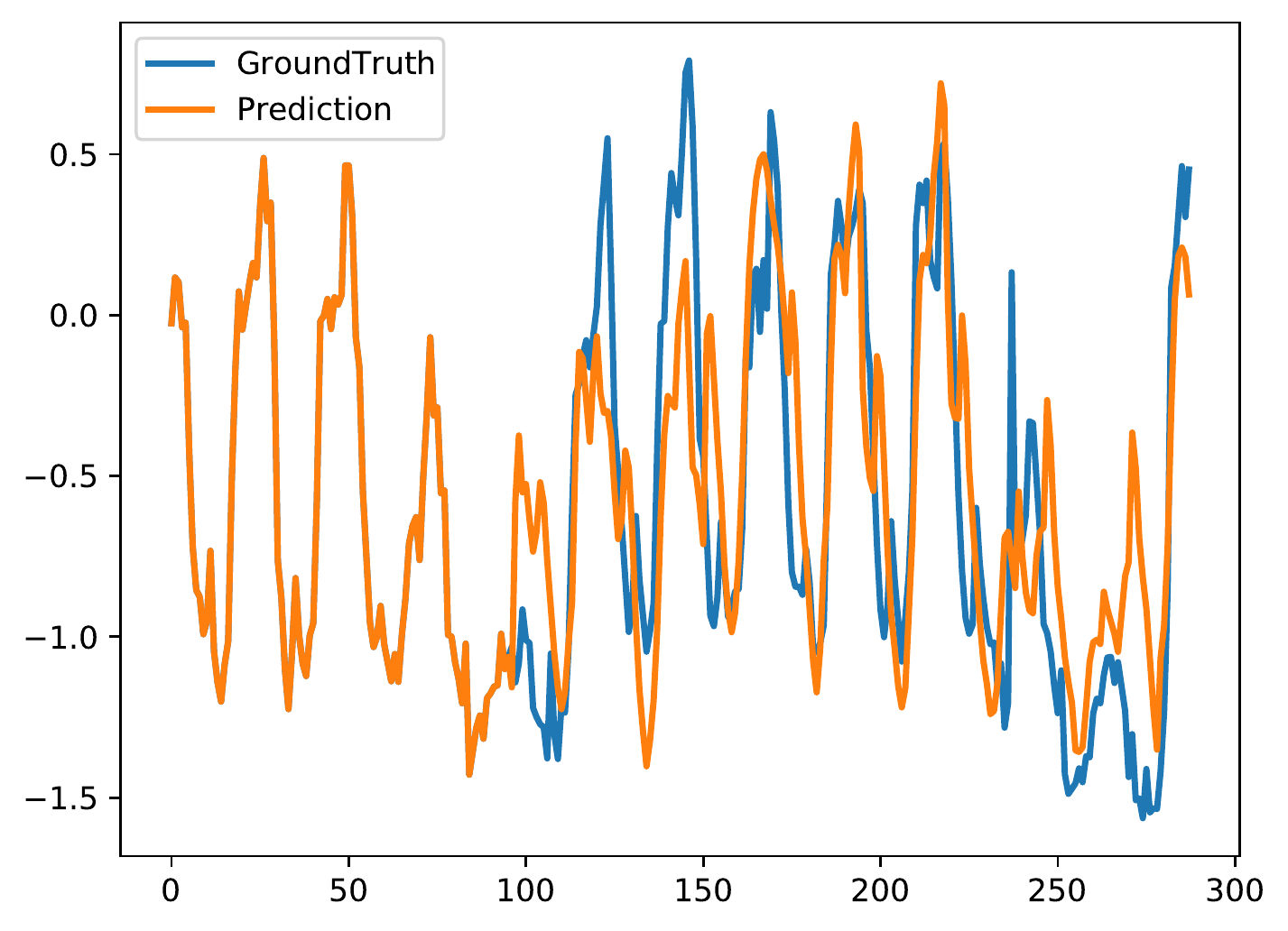}
    }
  \caption{The prediction results on the Electricity dataset under the input-96-predict-192 setting. }
\label{fig:7}
\end{center}
\vskip -0.2in
\end{figure}

\begin{figure}[H]
\vskip 0.2in
\begin{center}
    \subfigure[Preformer]{
    \includegraphics[width=0.22\columnwidth]{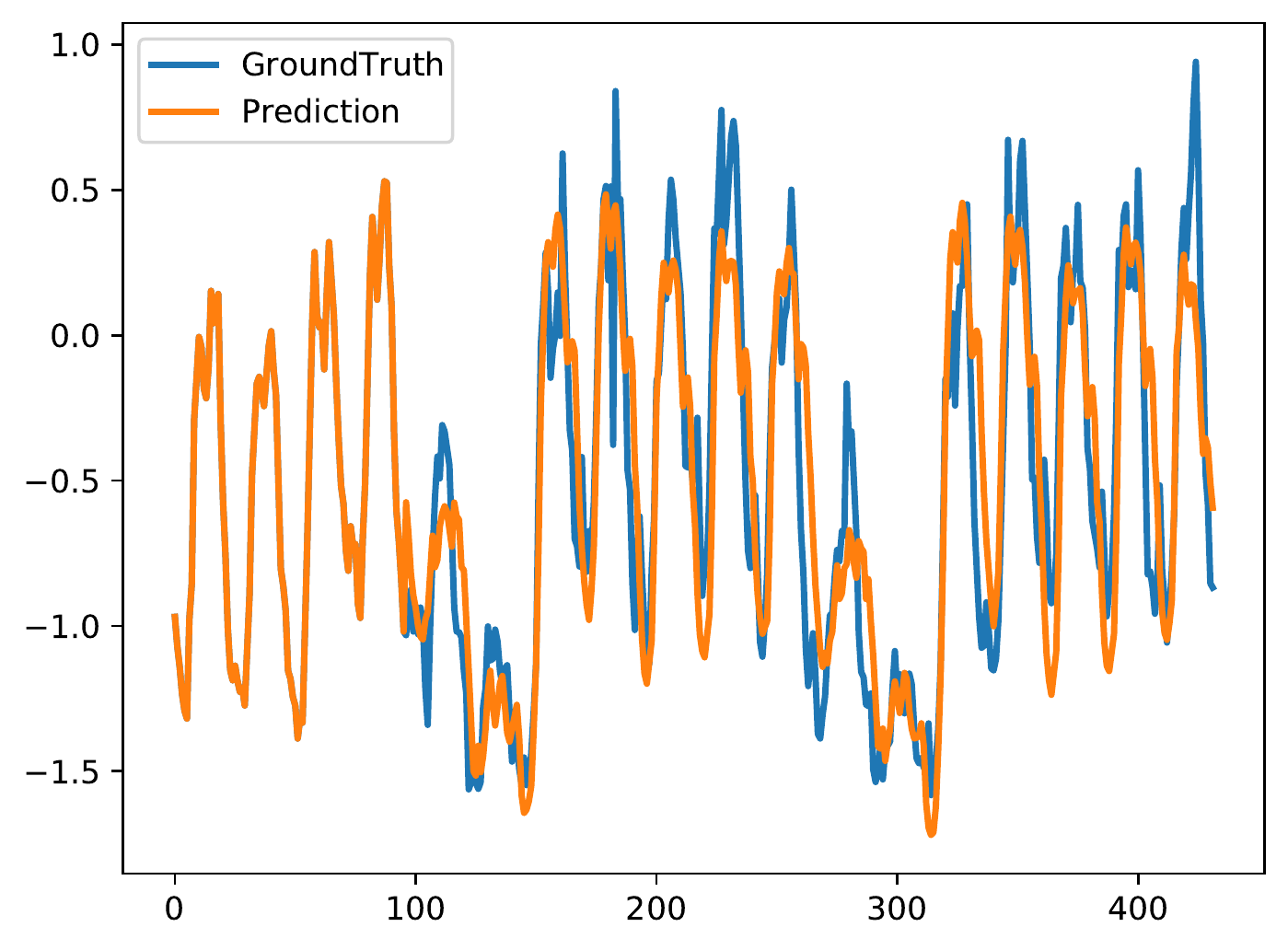}
    }
    \subfigure[Autoformer]{
    \includegraphics[width=0.22\columnwidth]{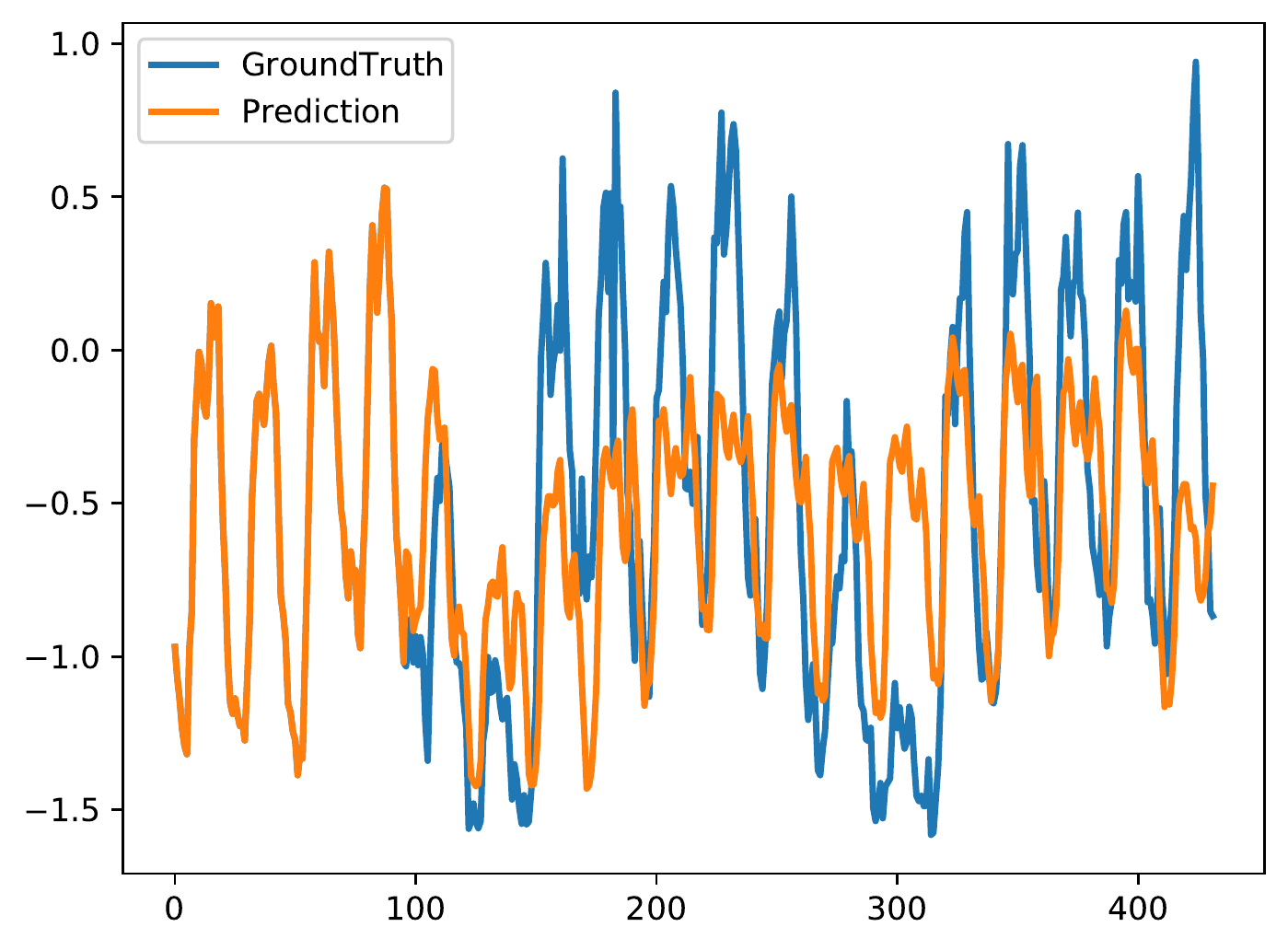}
    }
    \subfigure[Informer]{
    \includegraphics[width=0.22\columnwidth]{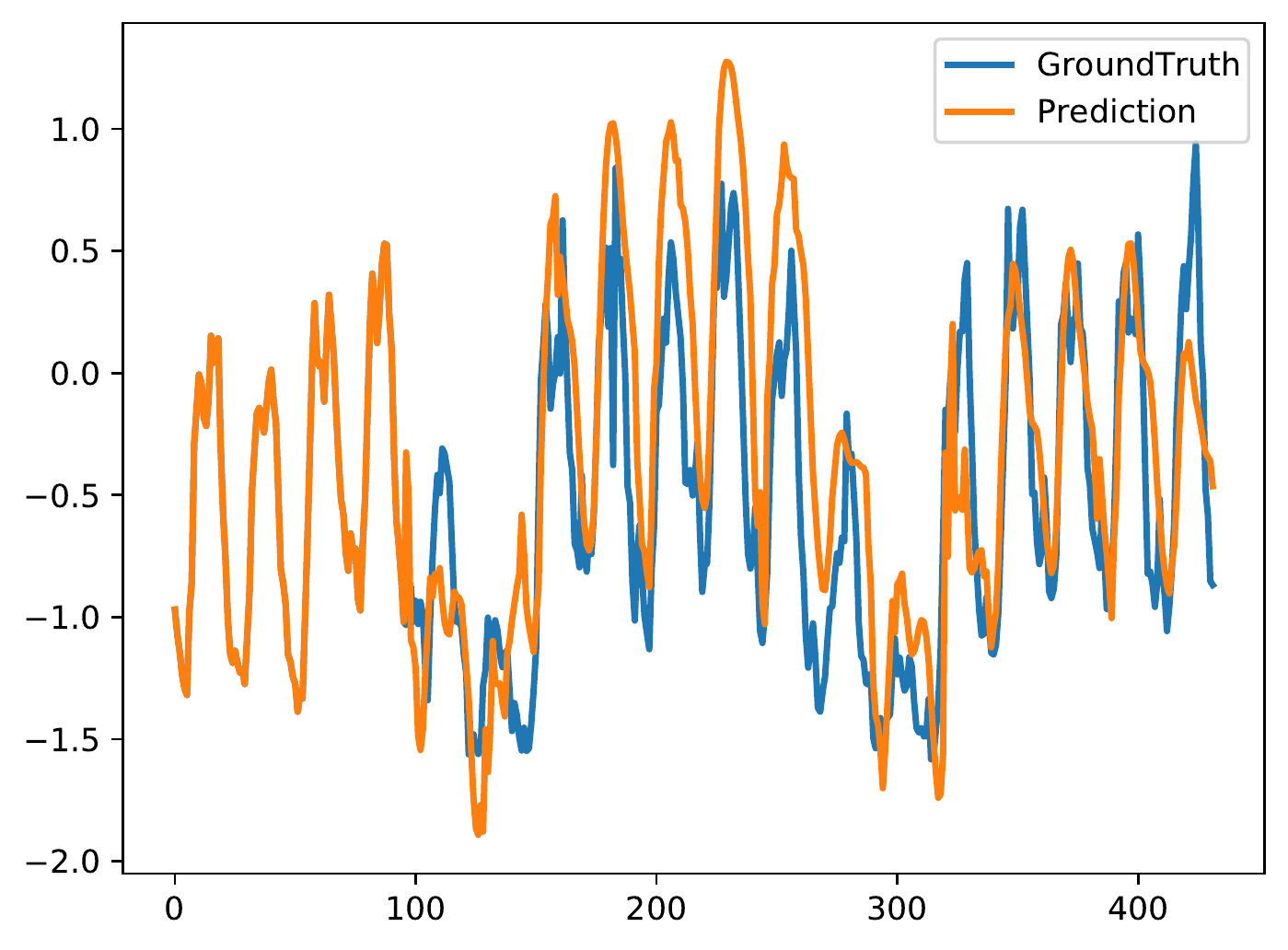}
    }
    \subfigure[LogTrans]{
    \includegraphics[width=0.22\columnwidth]{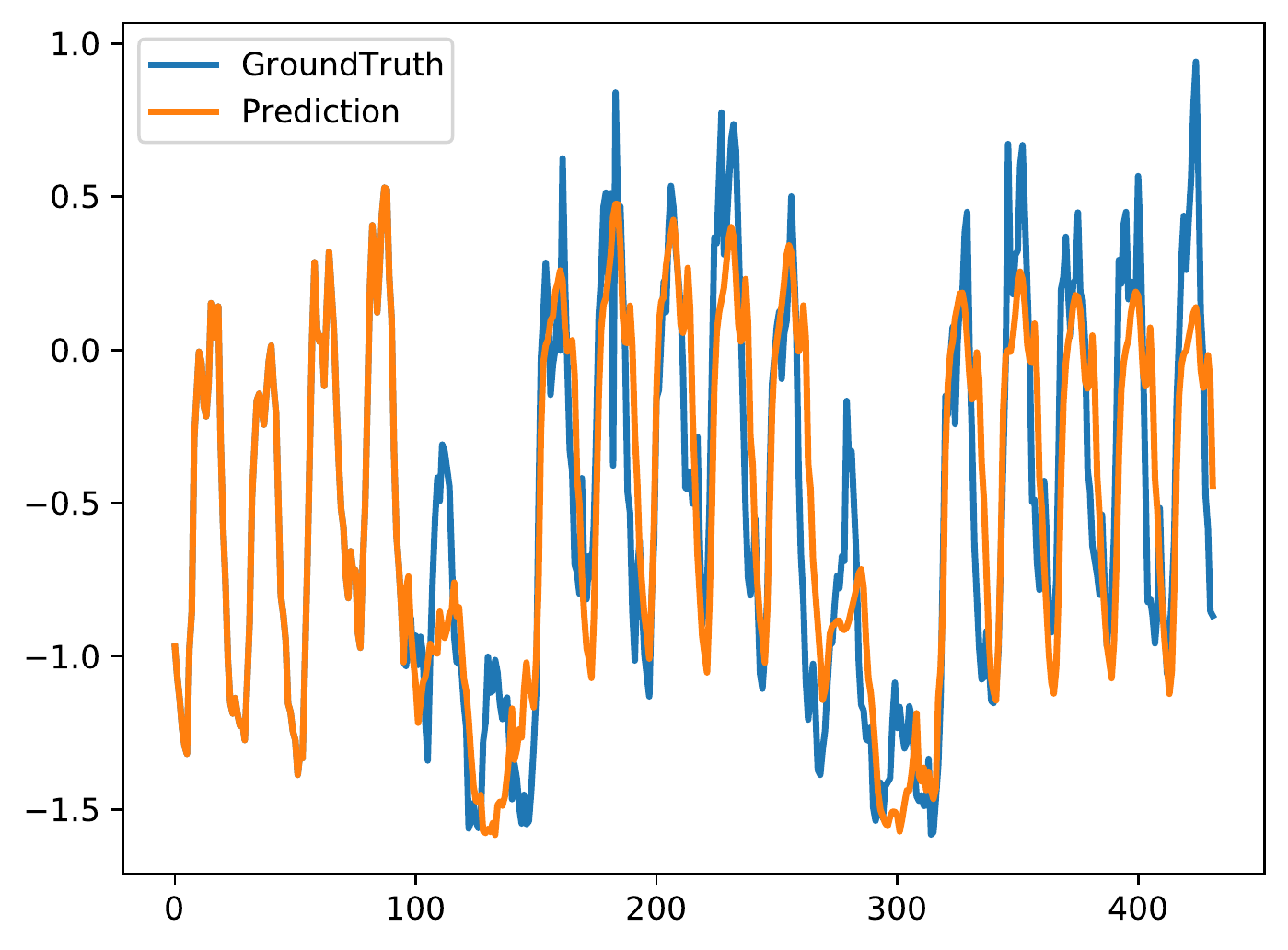}
    }
  \caption{The prediction results on the Electricity dataset under the input-96-predict-336 setting. }
\label{fig:8}
\end{center}
\vskip -0.2in
\end{figure}

\begin{figure}[H]
\vskip 0.2in
\begin{center}
    \subfigure[Preformer]{
    \includegraphics[width=0.22\columnwidth]{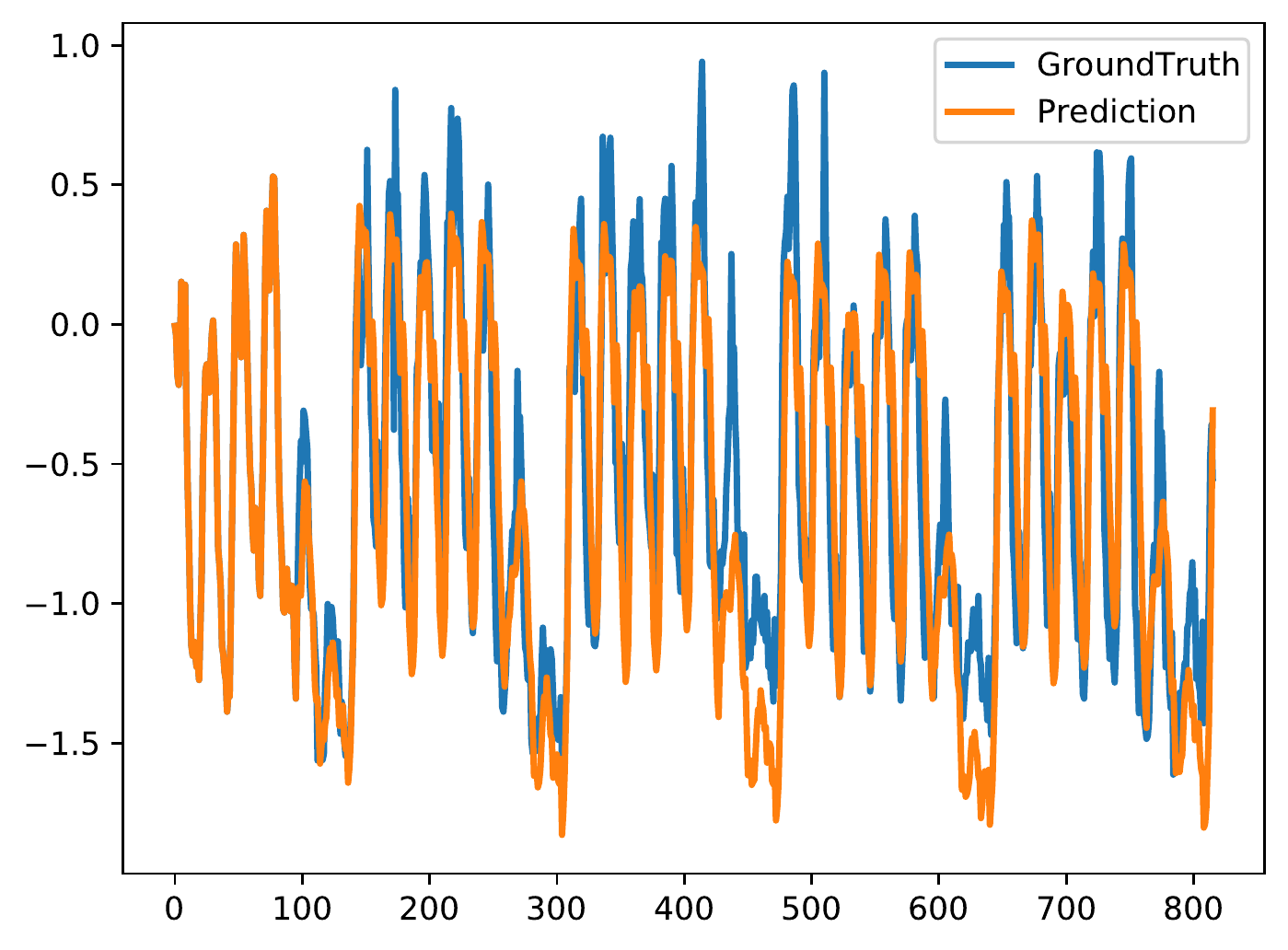}
    }
    \subfigure[Autoformer]{
    \includegraphics[width=0.22\columnwidth]{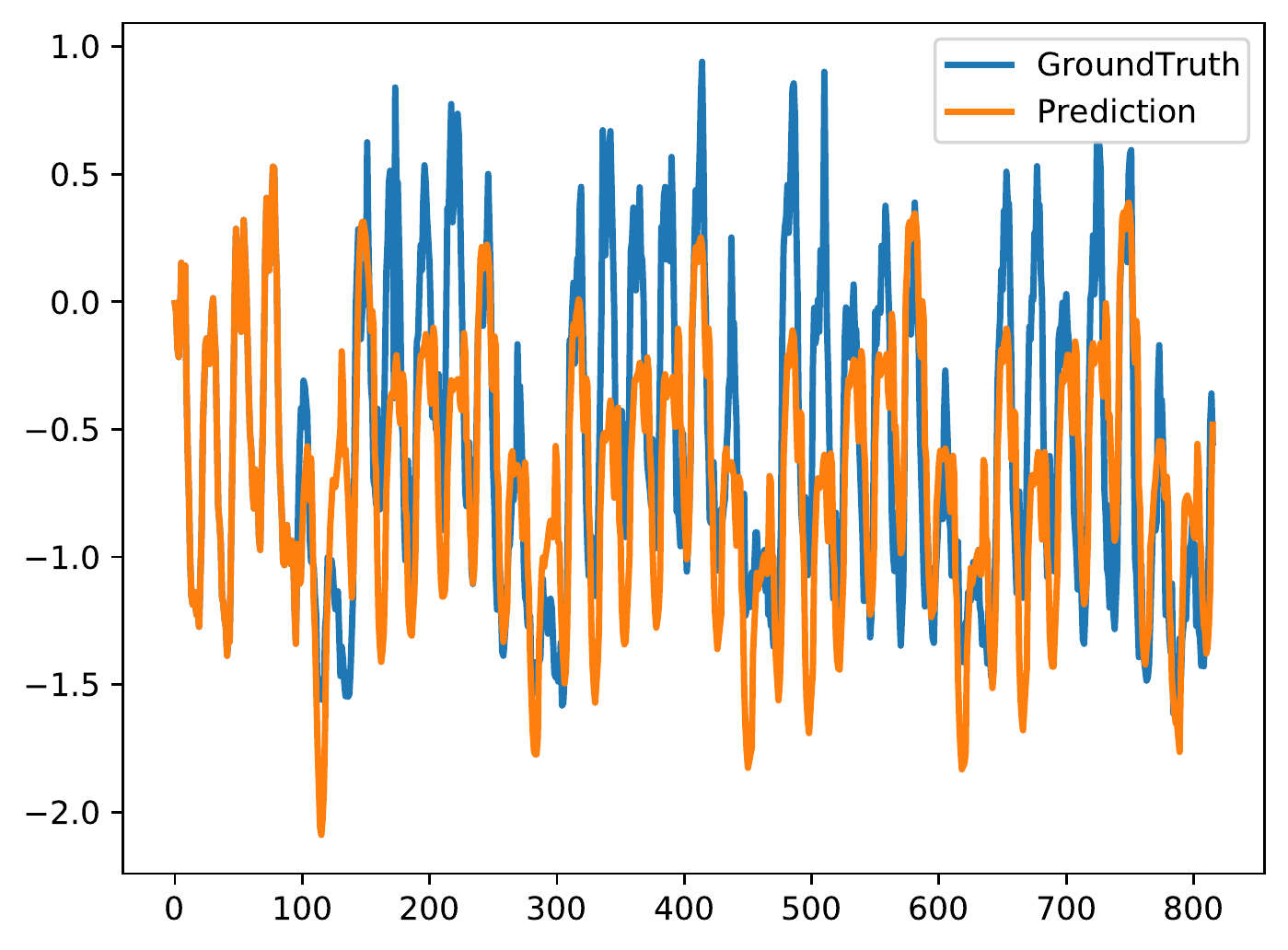}
    }
    \subfigure[Informer]{
    \includegraphics[width=0.22\columnwidth]{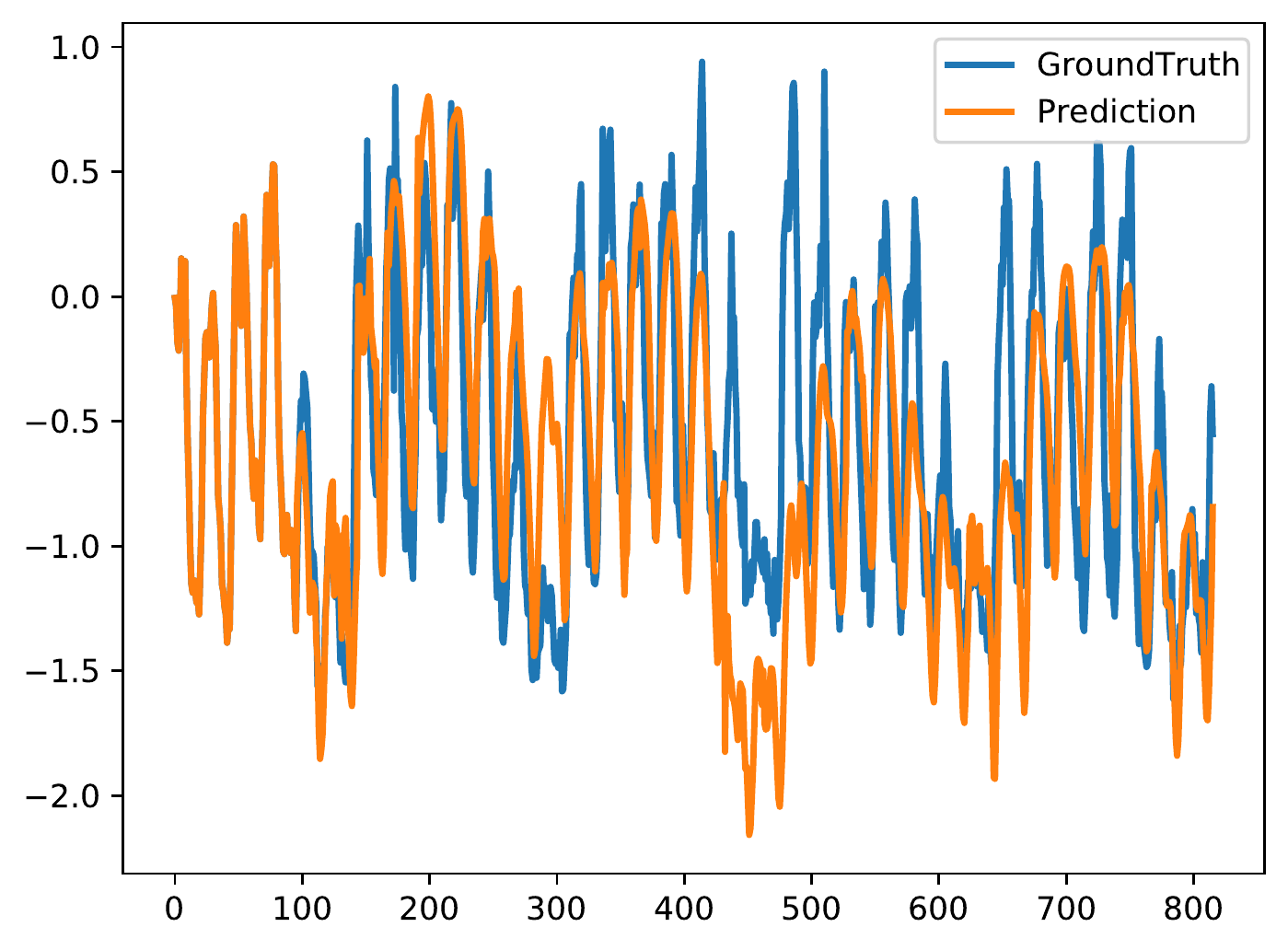}
    }
    \subfigure[LogTrans]{
    \includegraphics[width=0.22\columnwidth]{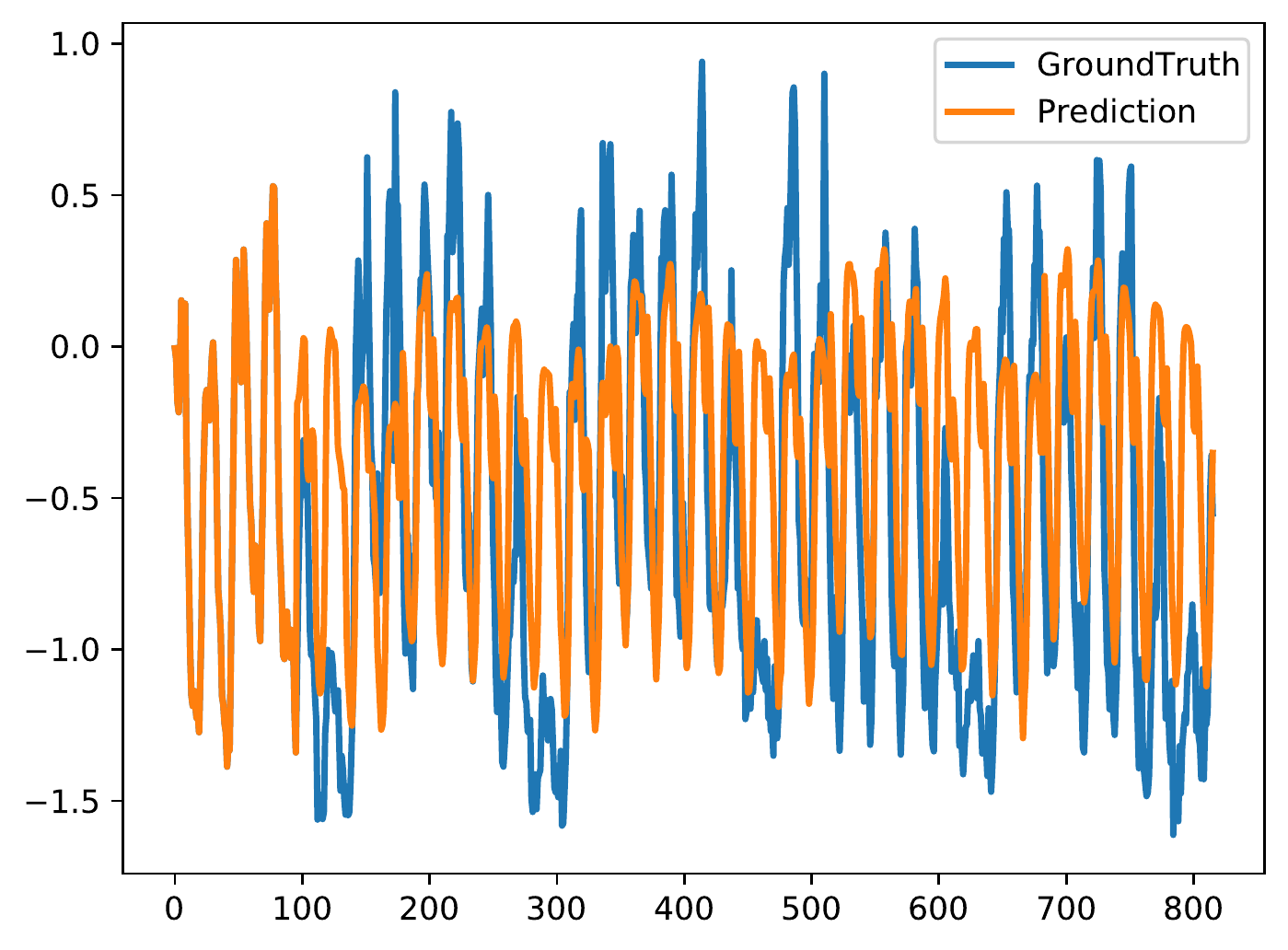}
    }
  \caption{The prediction results on the Electricity dataset under the input-96-predict-720 setting. }
\label{fig:9}
\end{center}
\vskip -0.2in
\end{figure}

As shown in \cref{fig:6,fig:7,fig:8,fig:9}, we plot the multivariate forecasting results of several Transformer-based models on the Electricity dataset. Blue lines are the ground truth and orange lines are the prediction results. Our Preformer can accurately predict the periodicity, trend and even some small fluctuations. Though under a very long prediction horizon, i.e., under the input-96-predict-720 setting, our Preformer can also perform well.

\subsection{Results of Preformer on various datasets}
\begin{figure}[H]
\vskip 0.2in
\begin{center}
    \subfigure[predict-96]{
    \includegraphics[width=0.22\columnwidth]{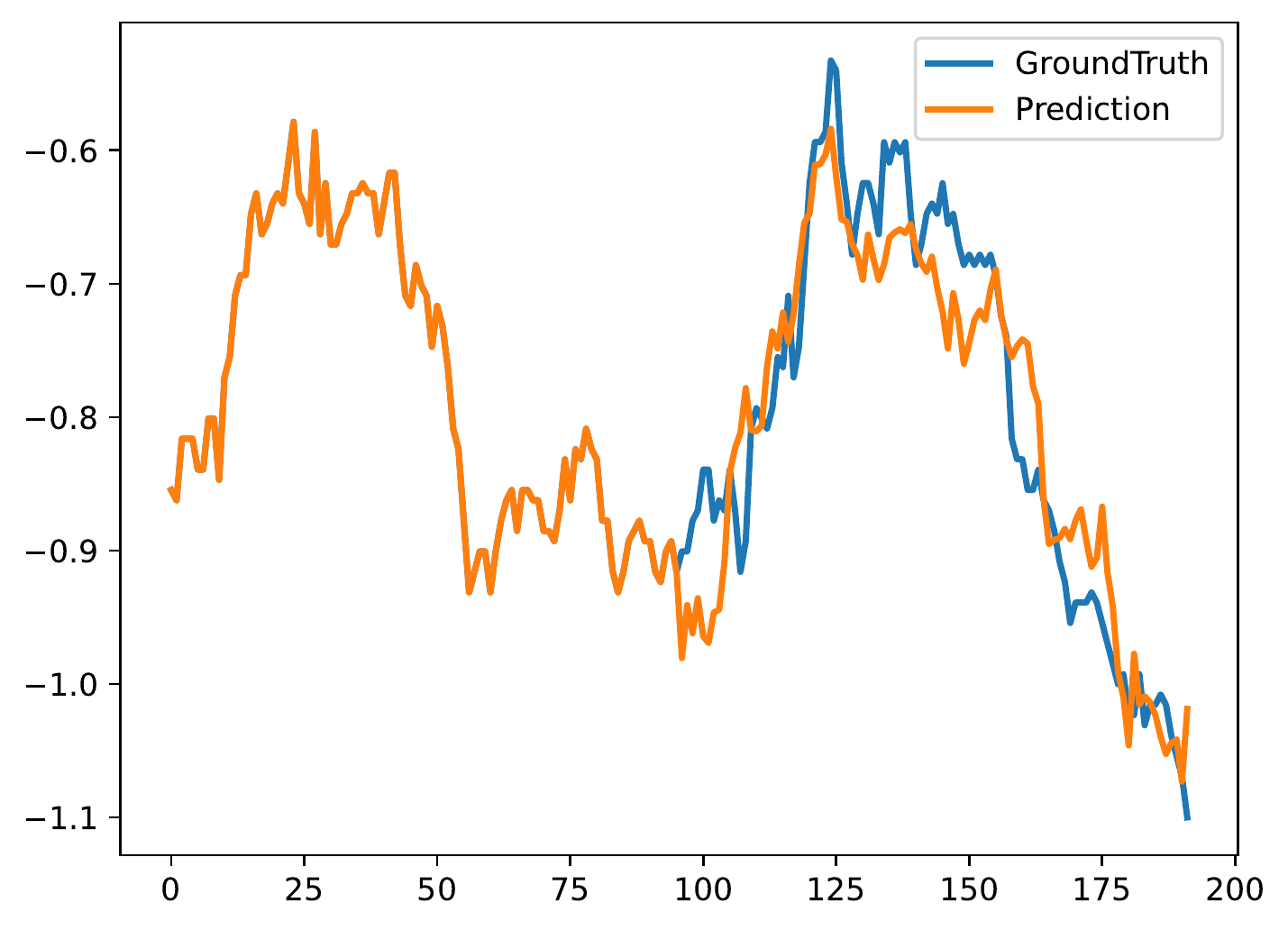}
    }
    \subfigure[predict-192]{
    \includegraphics[width=0.22\columnwidth]{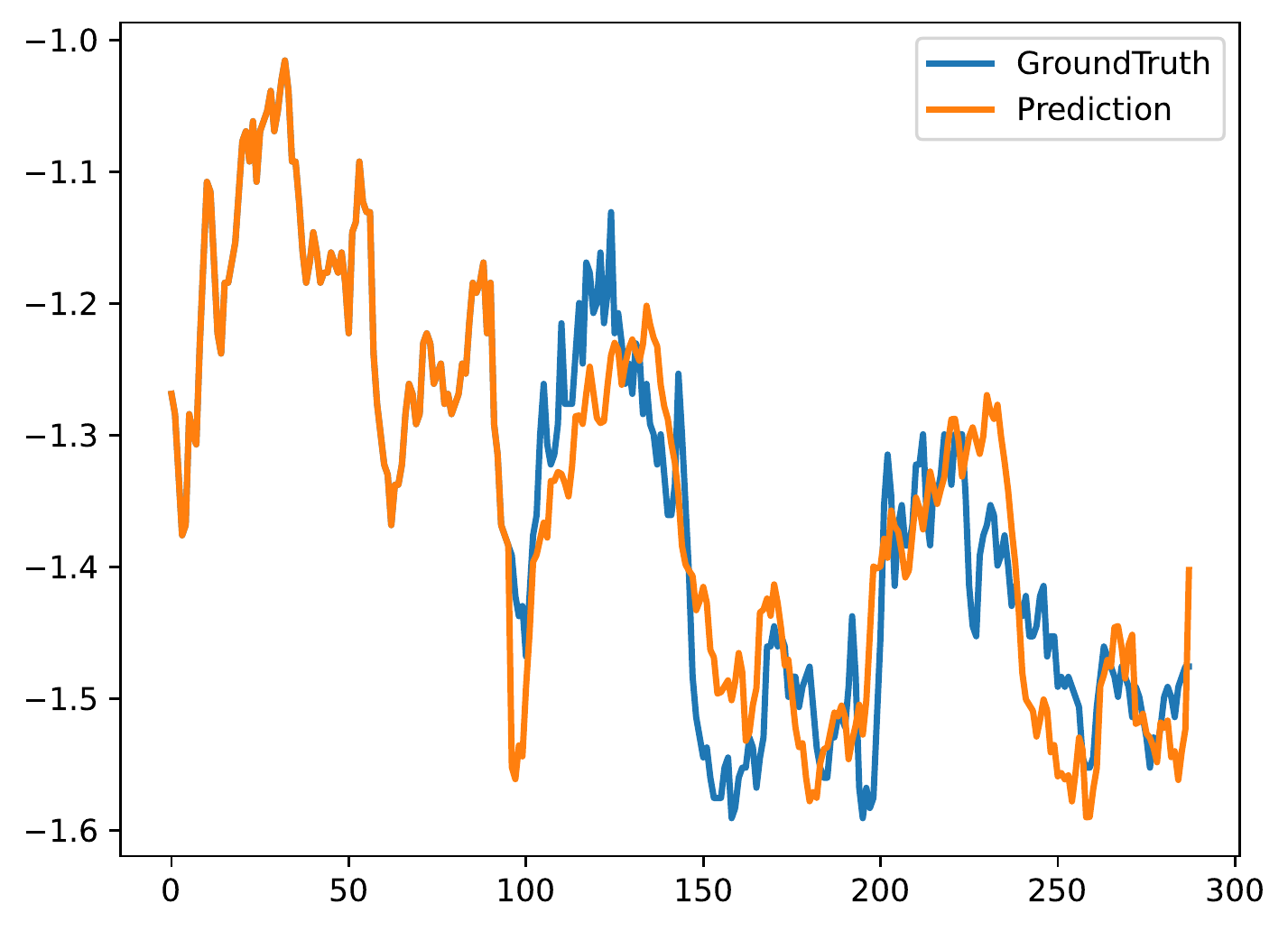}
    }
    \subfigure[predict-336]{
    \includegraphics[width=0.22\columnwidth]{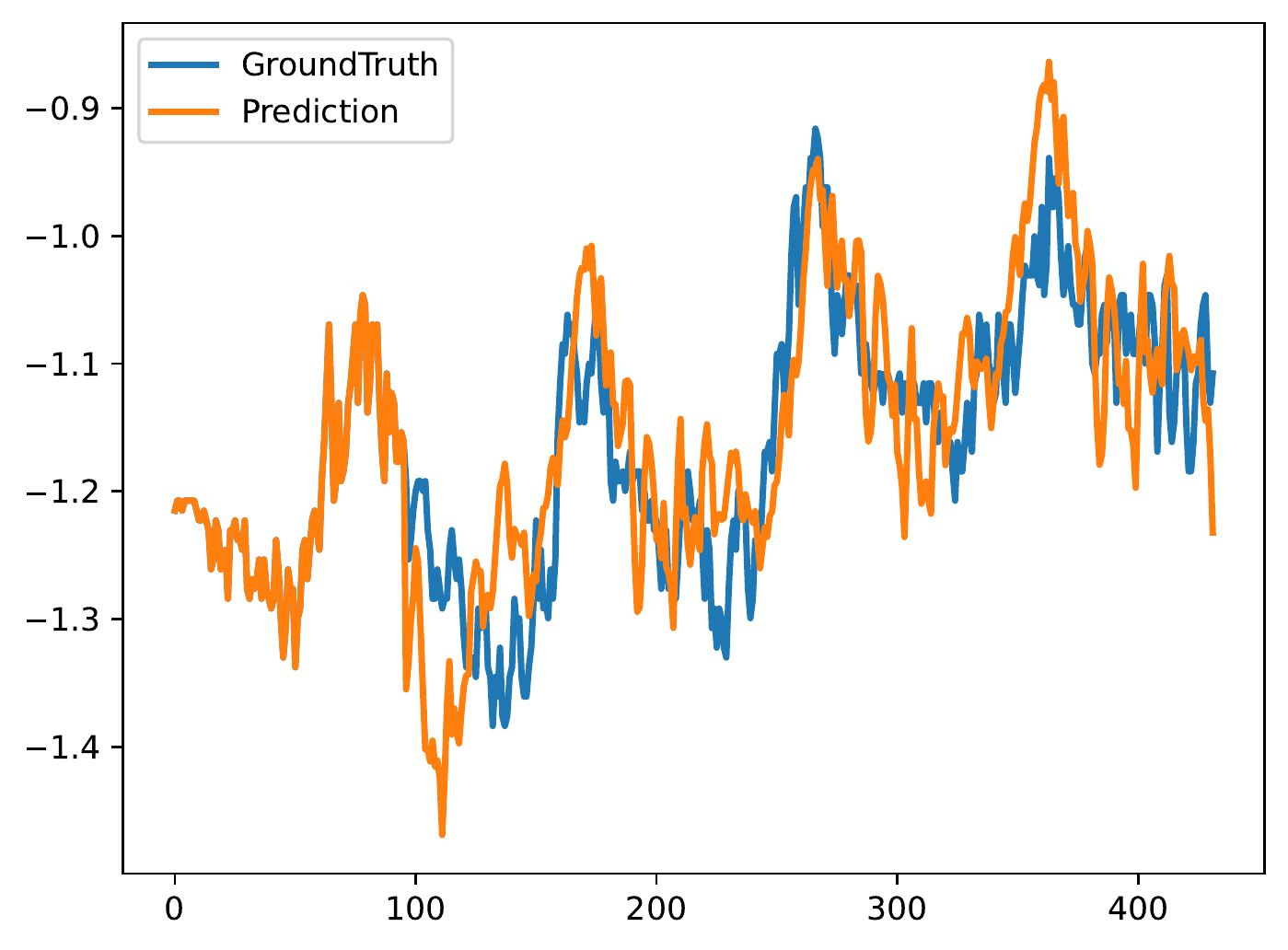}
    }
    \subfigure[predict-720]{
    \includegraphics[width=0.22\columnwidth]{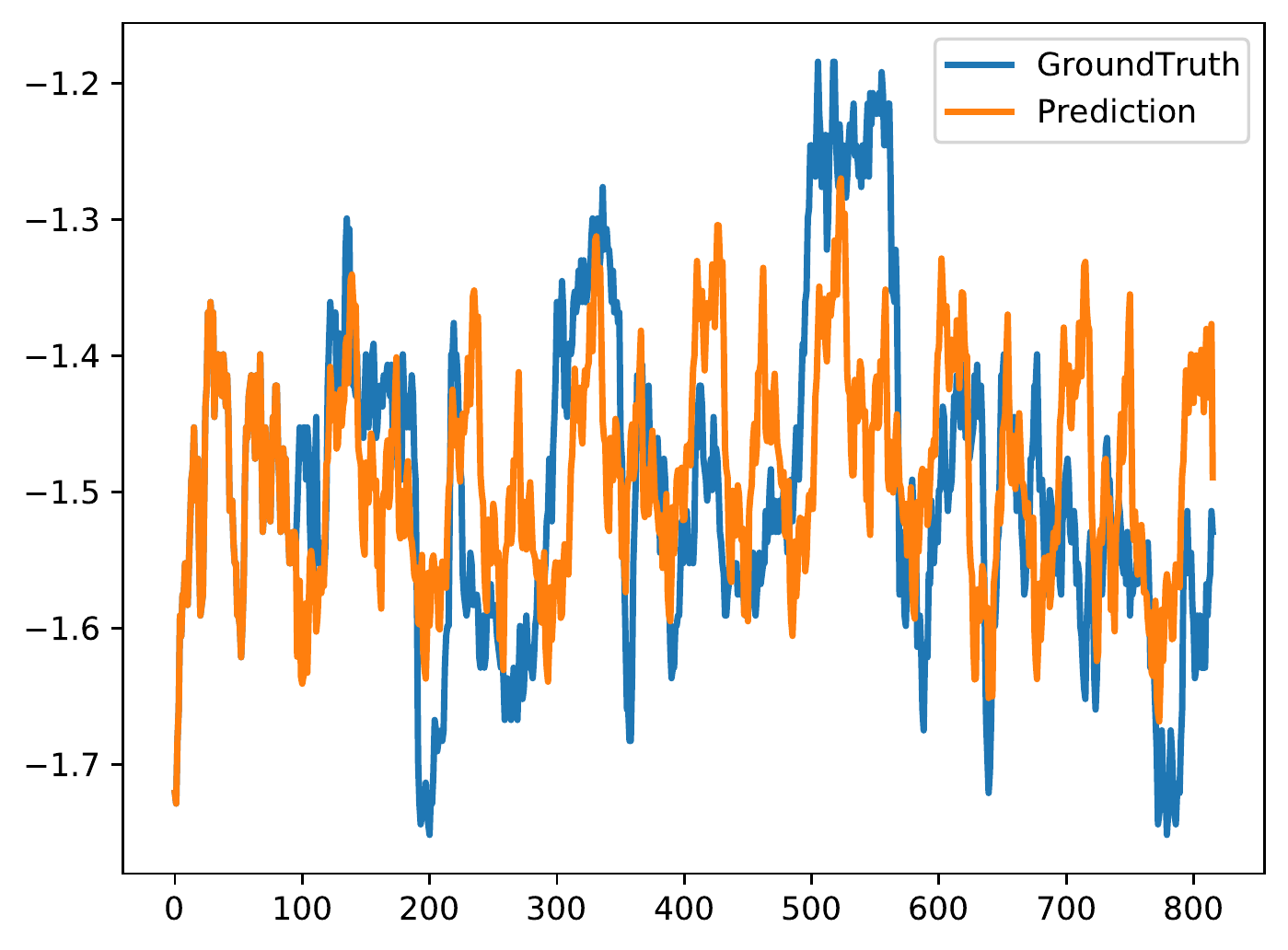}
    }
  \caption{Some prediction results of Preformer on the ETTm1 dataset under different prediction lengths.}
  \label{fig:10}
\end{center}
\vskip -0.2in
\end{figure}

\begin{figure}[H]
\vskip 0.2in
\begin{center}
    \subfigure[predict-96]{
    \includegraphics[width=0.22\columnwidth]{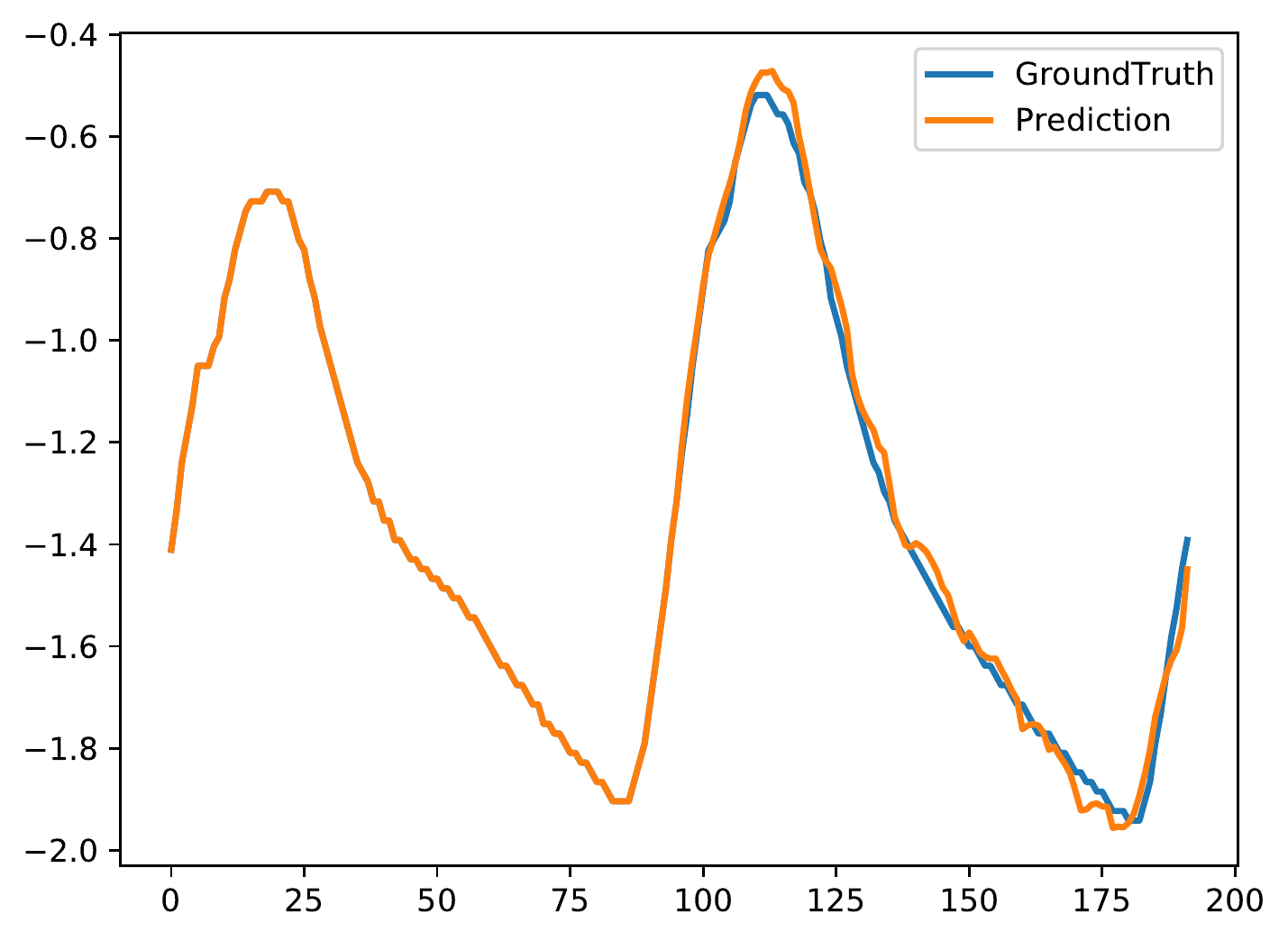}
    }
    \subfigure[predict-192]{
    \includegraphics[width=0.22\columnwidth]{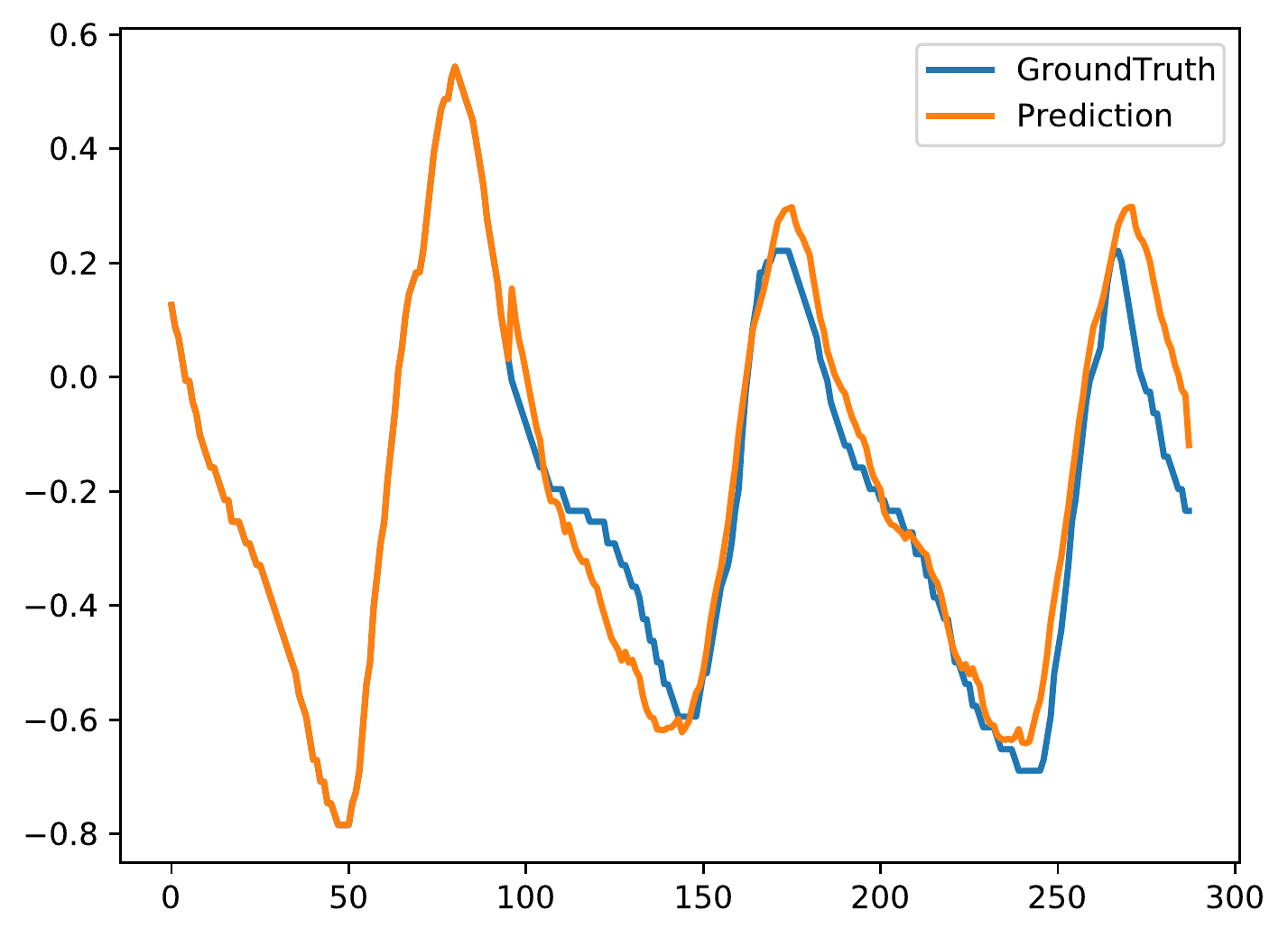}
    }
    \subfigure[predict-336]{
    \includegraphics[width=0.22\columnwidth]{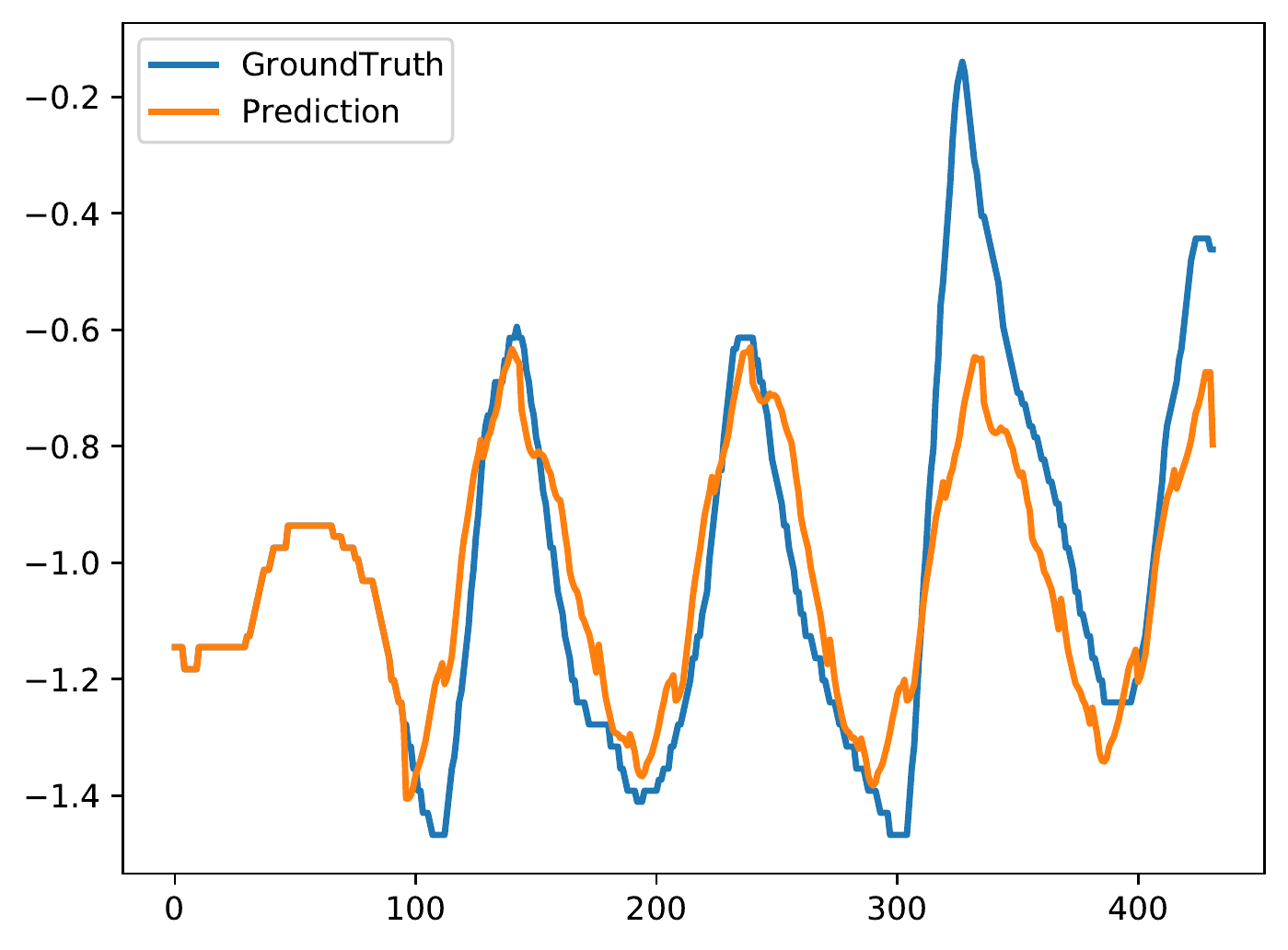}
    }
    \subfigure[predict-720]{
    \includegraphics[width=0.22\columnwidth]{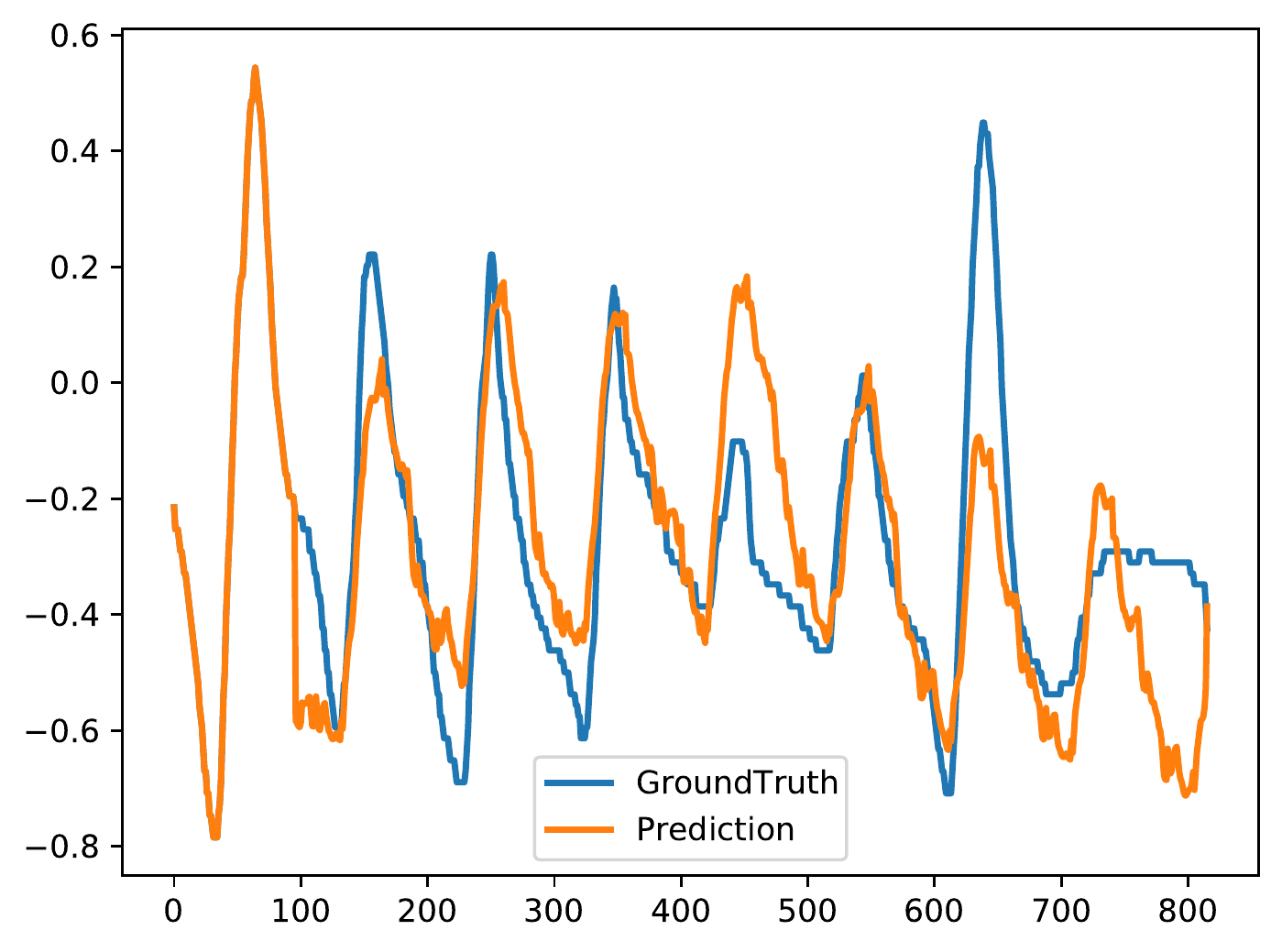}
    }
  \caption{Some prediction results of Preformer on the ETTm2 dataset under different prediction lengths.}
    \label{fig:11}
\end{center}
\vskip -0.2in
\end{figure}

\begin{figure}[H]
\vskip 0.2in
\begin{center}
    \subfigure[predict-96]{
    \includegraphics[width=0.22\columnwidth]{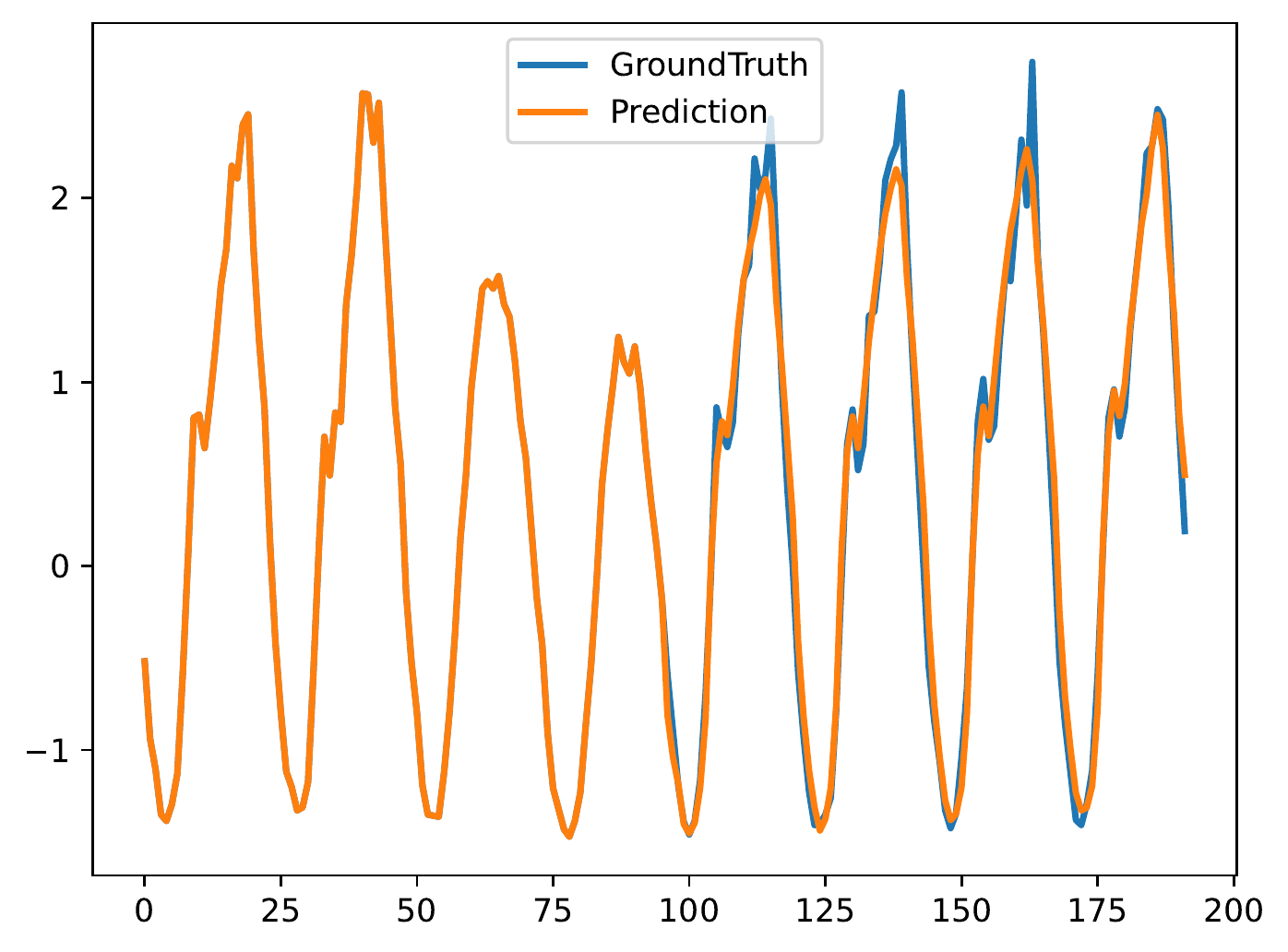}
    }
    \subfigure[predict-192]{
    \includegraphics[width=0.22\columnwidth]{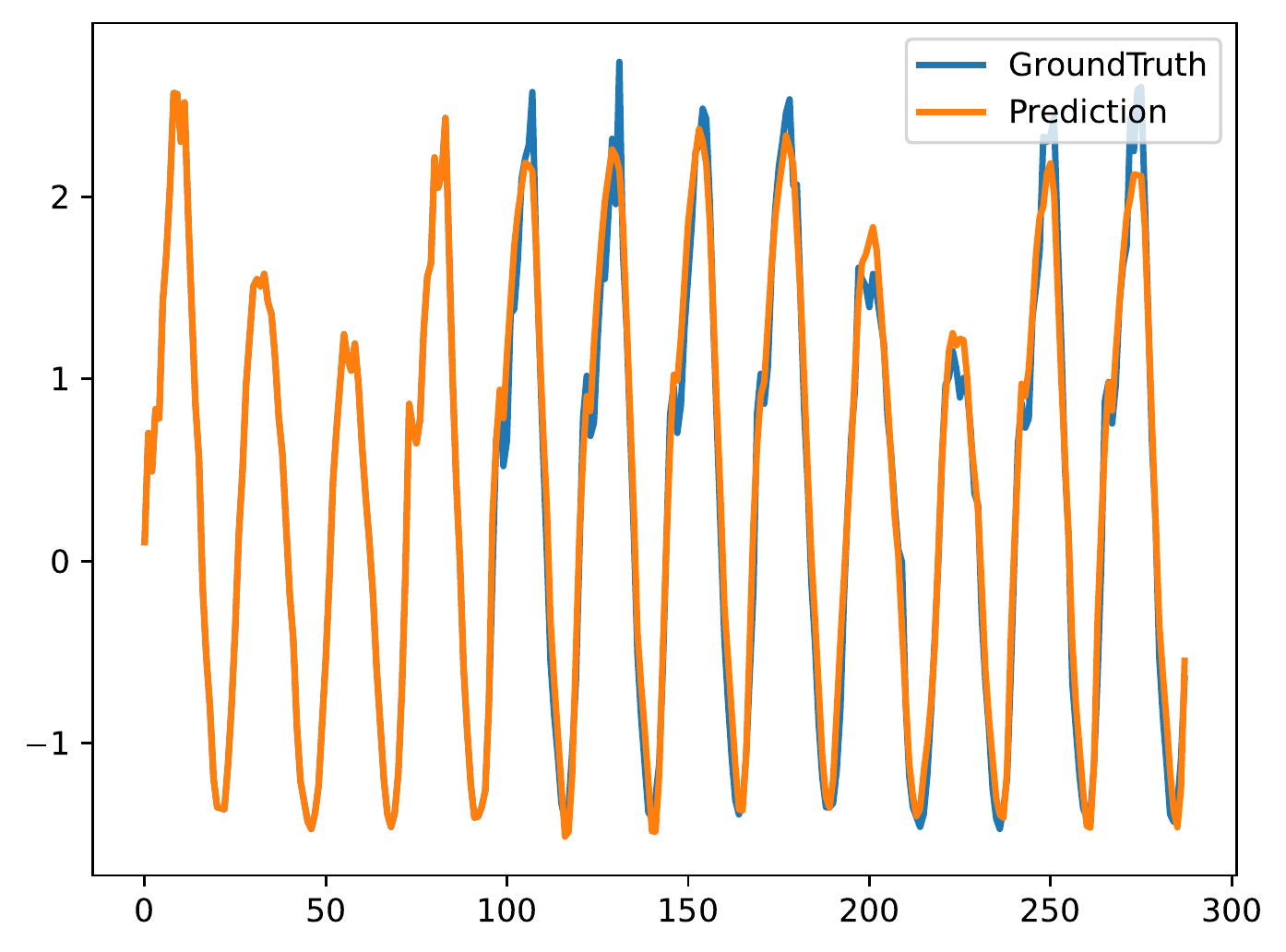}
    }
    \subfigure[predict-336]{
    \includegraphics[width=0.22\columnwidth]{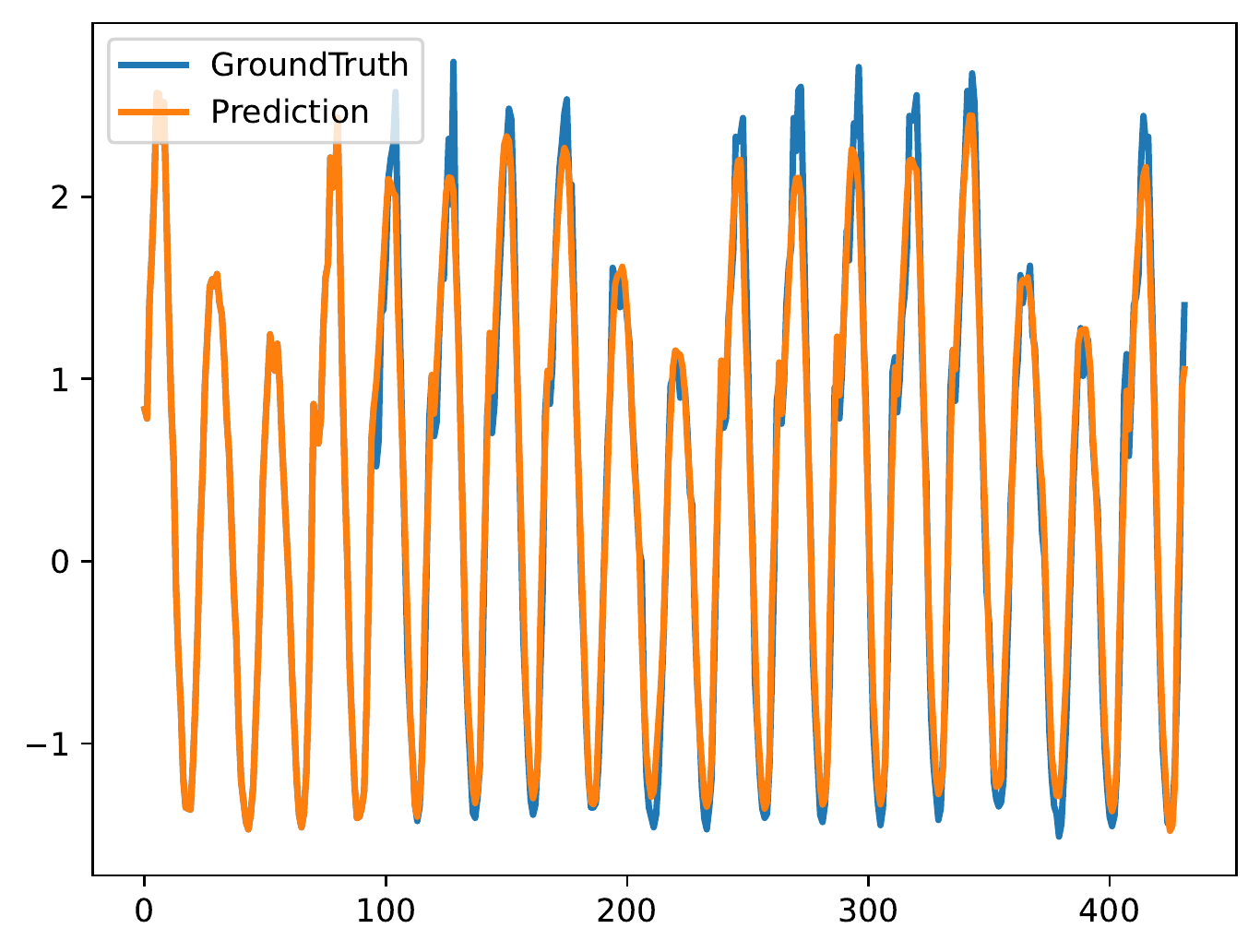}
    }
    \subfigure[predict-720]{
    \includegraphics[width=0.22\columnwidth]{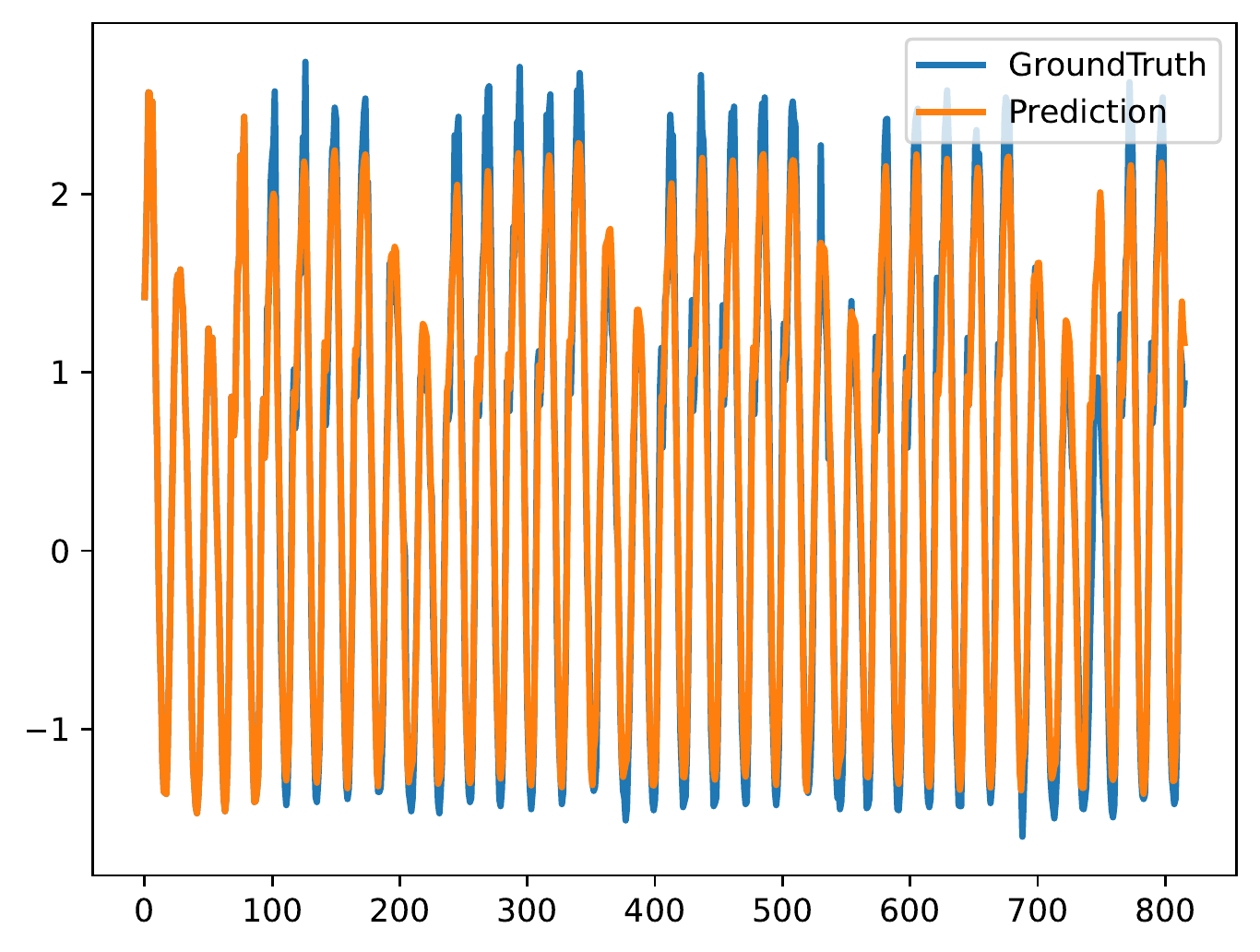}
    }
  \caption{Some prediction results of Preformer on the Traffic dataset under different prediction lengths.}
    \label{fig:12}
\end{center}
\vskip -0.2in
\end{figure}

We plot forecasting results of our Preformer on more datasets in \cref{fig:10,fig:11,fig:12}. We find that Preformer can obtain satisfactory prediction results whether on datasets with strong periodicity or on datasets with strong noise. 

\clearpage
\section{Visualization of Univariate Time-Series Forecasting}
\subsection{Comparison of Transformer-based Models}
\begin{figure}[H]
\vskip 0.2in
\begin{center}
    \subfigure[Preformer]{
    \includegraphics[width=0.22\columnwidth]{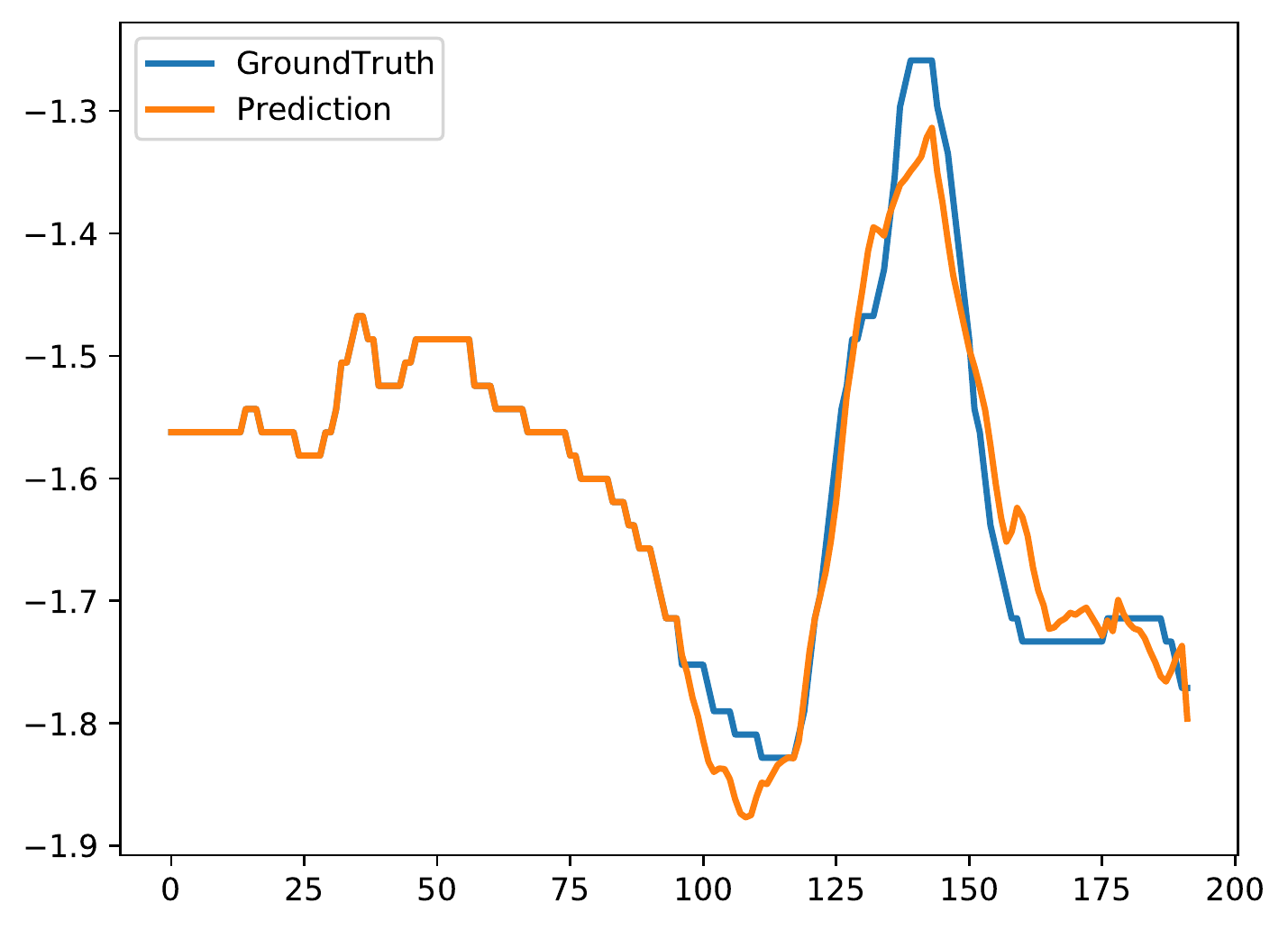}
    }
    \subfigure[Autoformer]{
    \includegraphics[width=0.22\columnwidth]{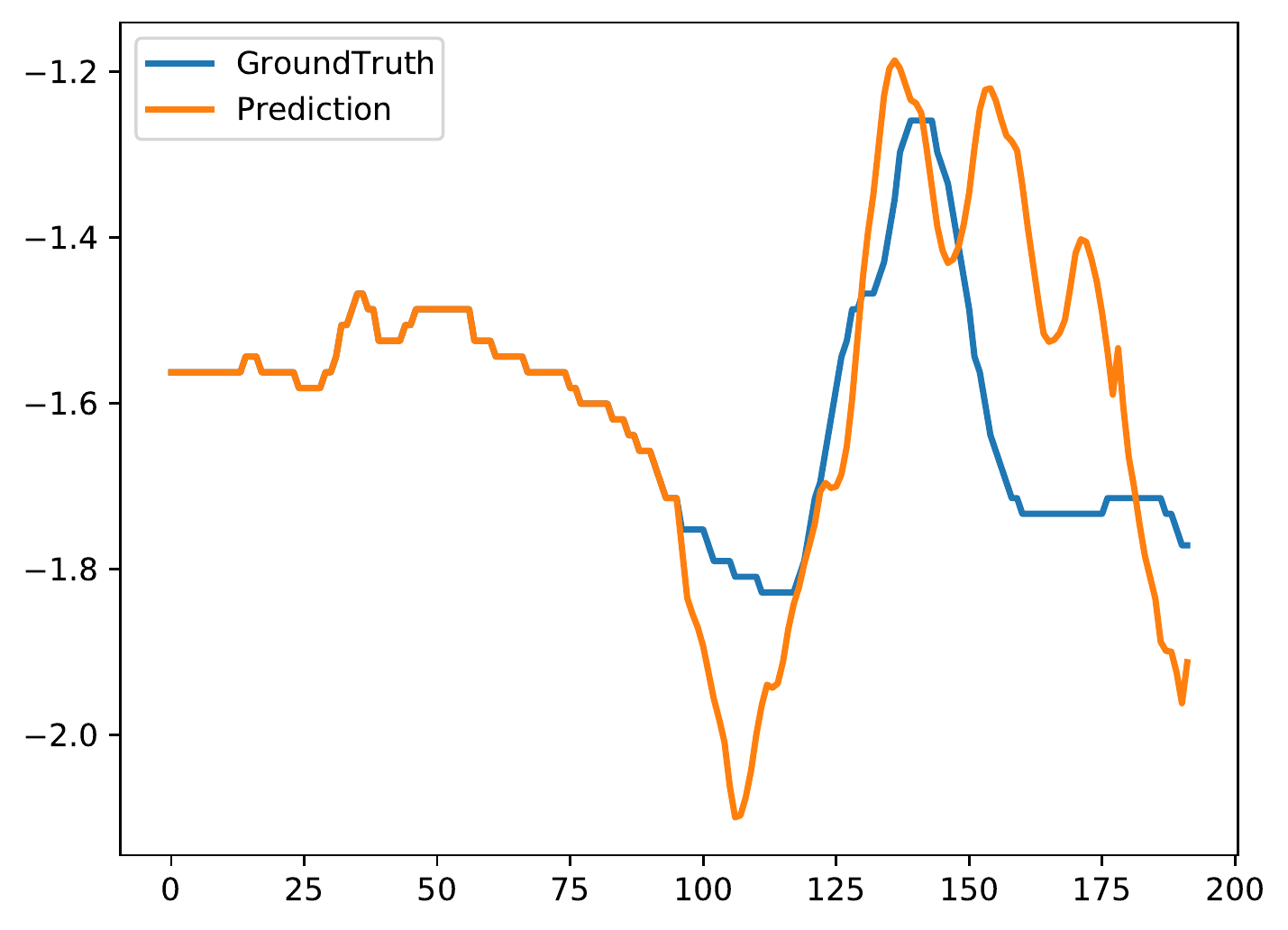}
    }
    \subfigure[Informer]{
    \includegraphics[width=0.22\columnwidth]{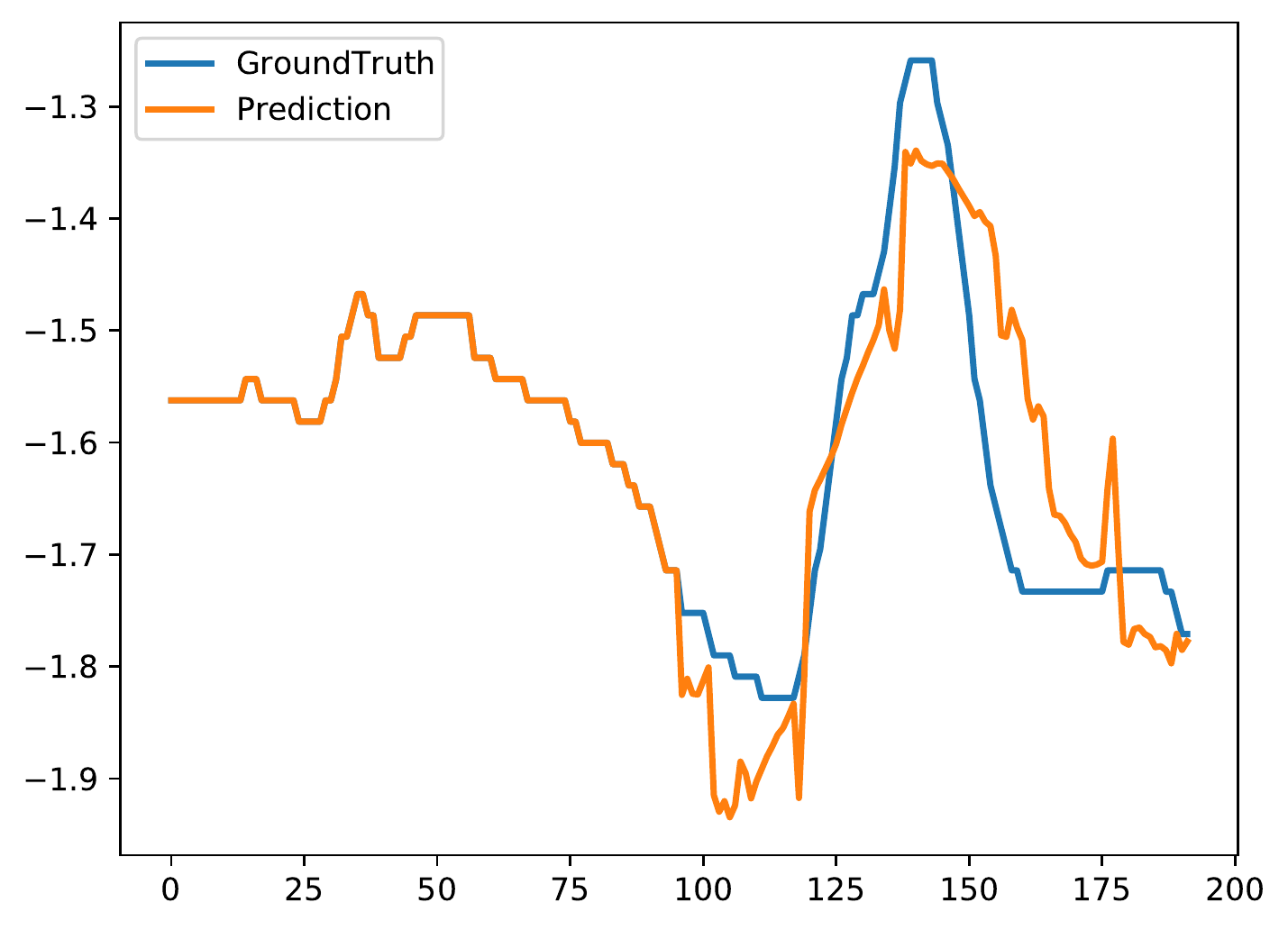}
    }
    \subfigure[LogTrans]{
    \includegraphics[width=0.22\columnwidth]{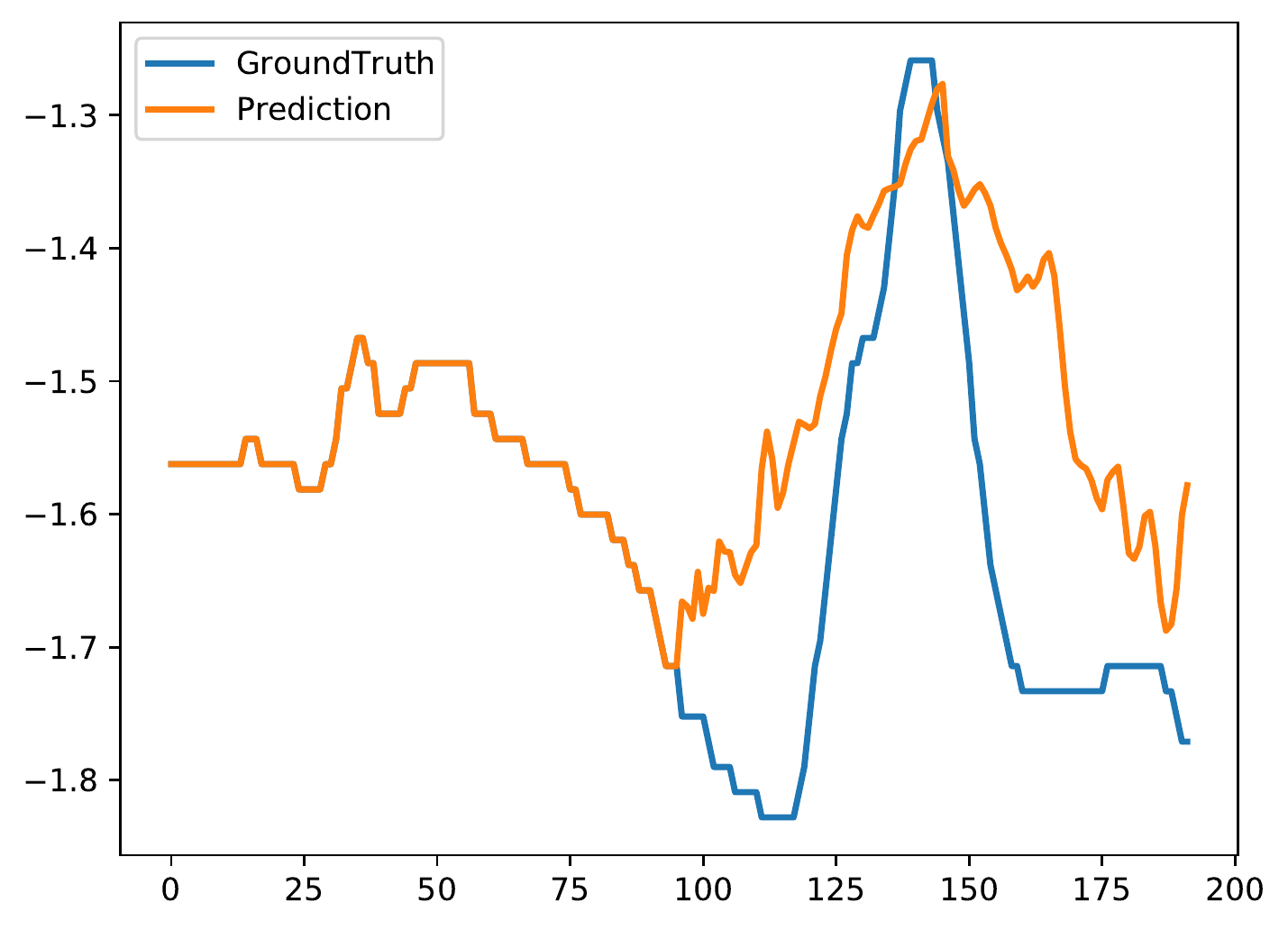}
    }
  \caption{The prediction results on the ETTm2 dataset under the input-96-predict-96 setting. }
  \label{fig:13}
\end{center}
\vskip -0.2in
\end{figure}

\begin{figure}[H]
\vskip 0.2in
\begin{center}
    \subfigure[Preformer]{
    \includegraphics[width=0.22\columnwidth]{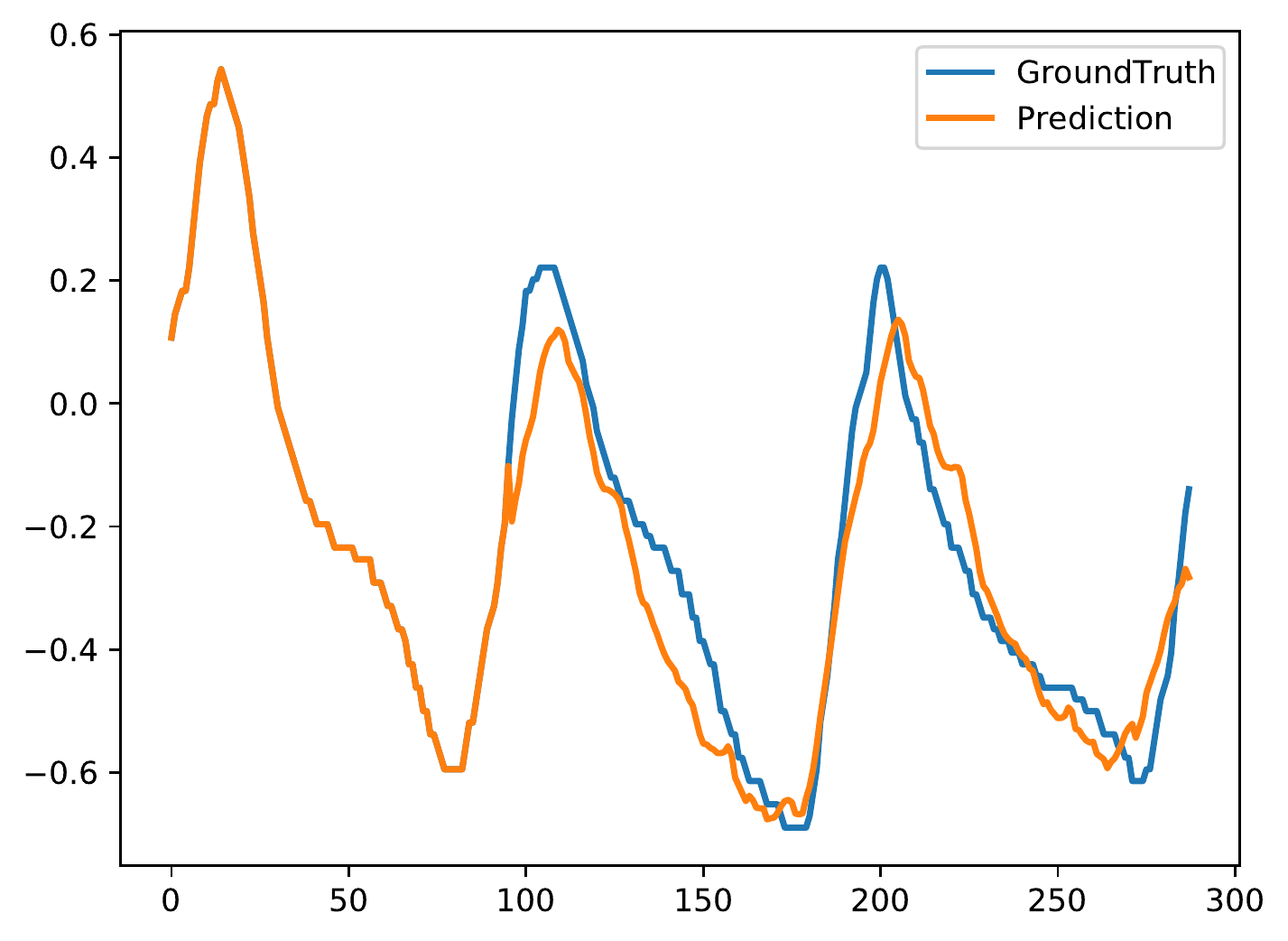}
    }
    \subfigure[Autoformer]{
    \includegraphics[width=0.22\columnwidth]{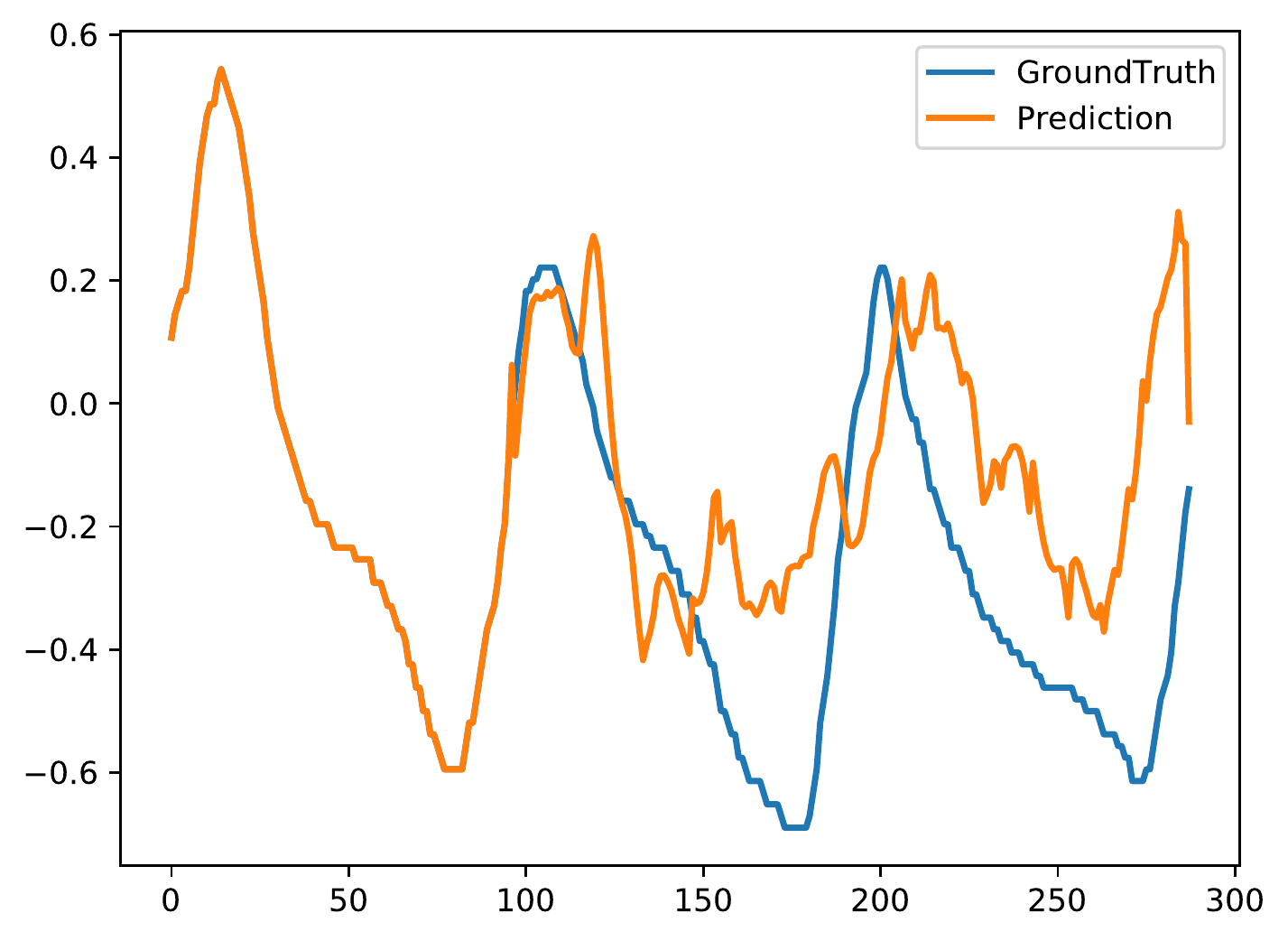}
    }
    \subfigure[Informer]{
    \includegraphics[width=0.22\columnwidth]{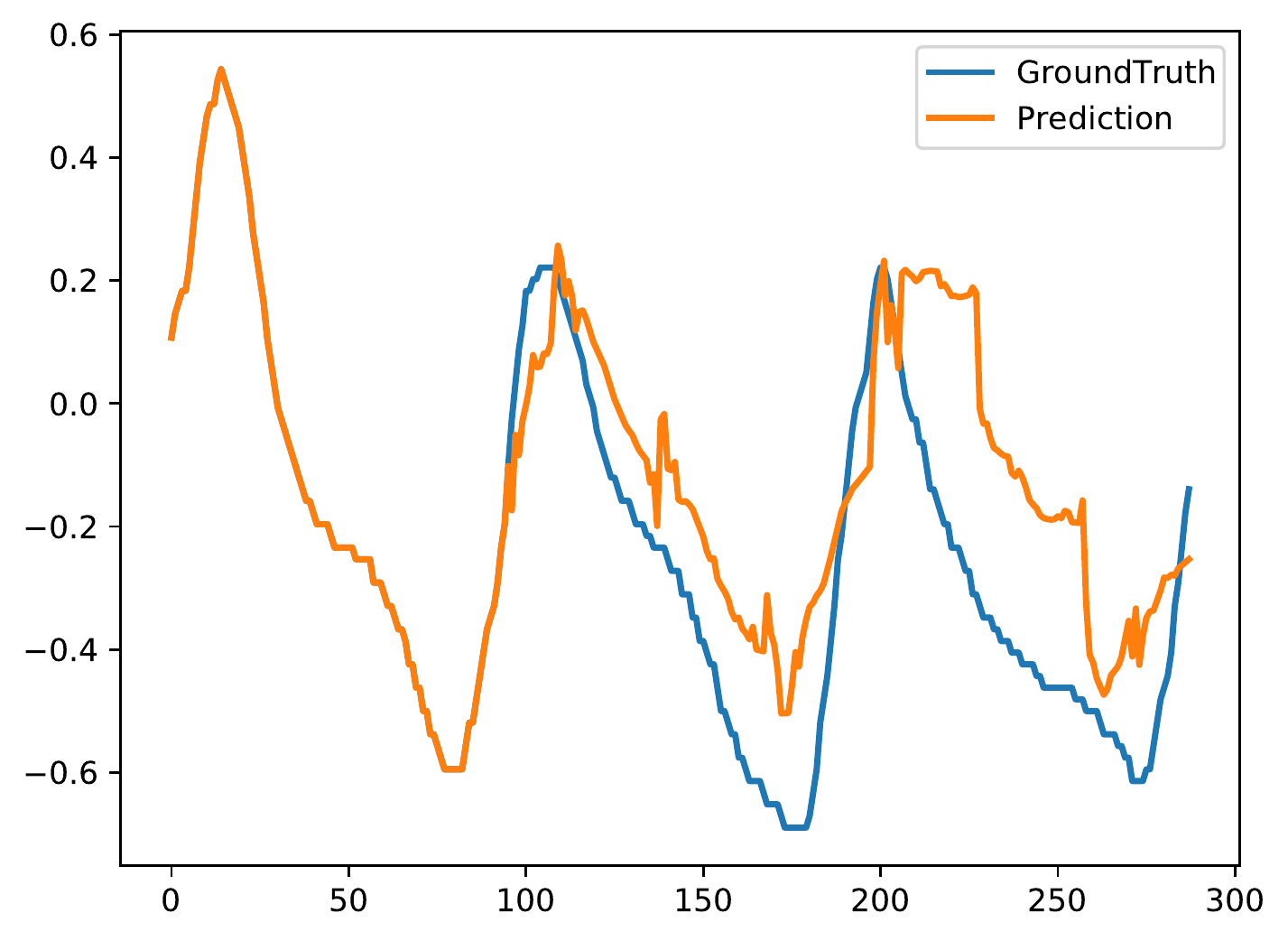}
    }
    \subfigure[LogTrans]{
    \includegraphics[width=0.22\columnwidth]{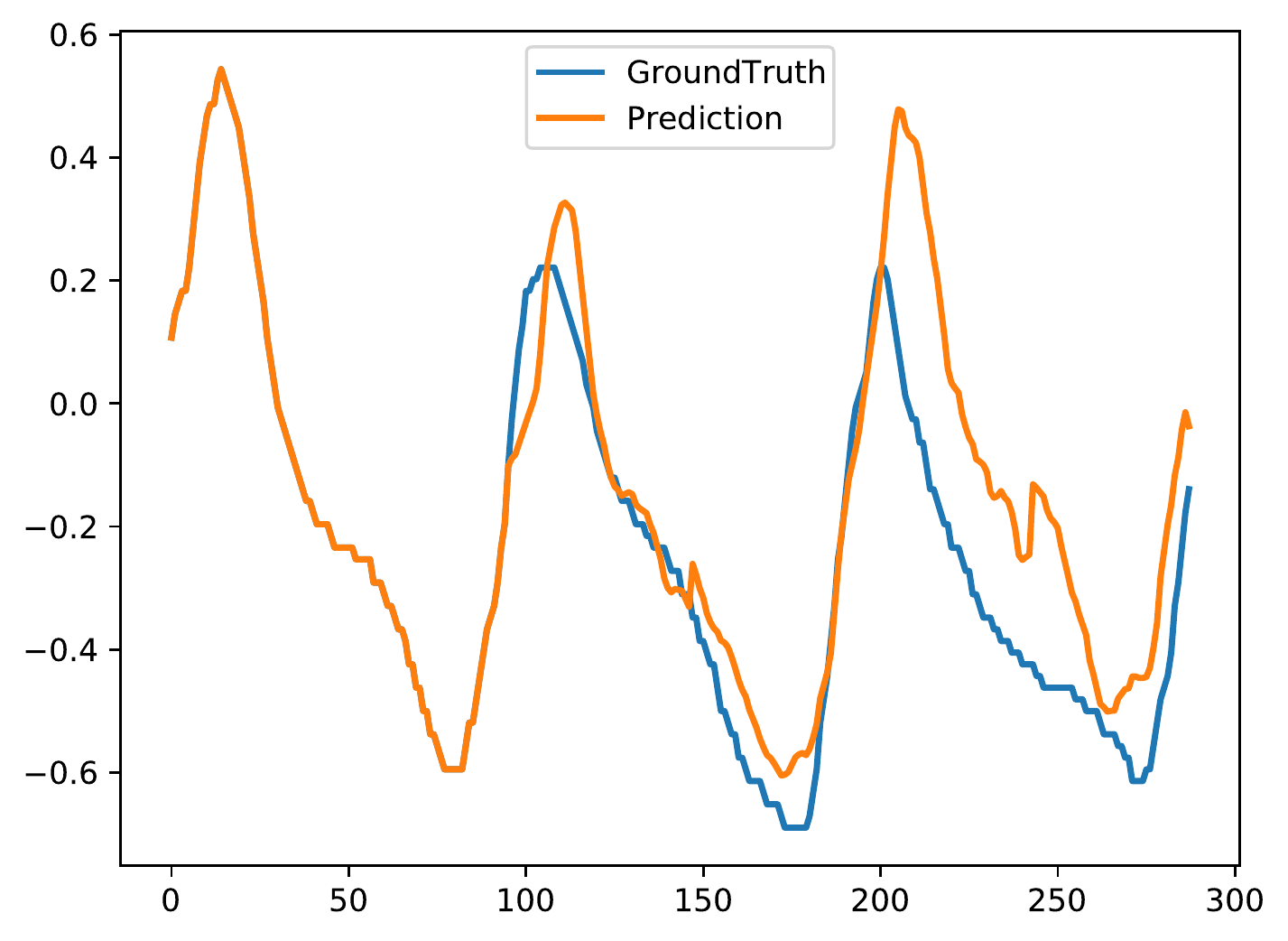}
    }
  \caption{The prediction results on the ETTm2 dataset under the input-96-predict-192 setting. }
    \label{fig:14}
\end{center}
\vskip -0.2in
\end{figure}

\begin{figure}[H]
\vskip 0.2in
\begin{center}
    \subfigure[Preformer]{
    \includegraphics[width=0.22\columnwidth]{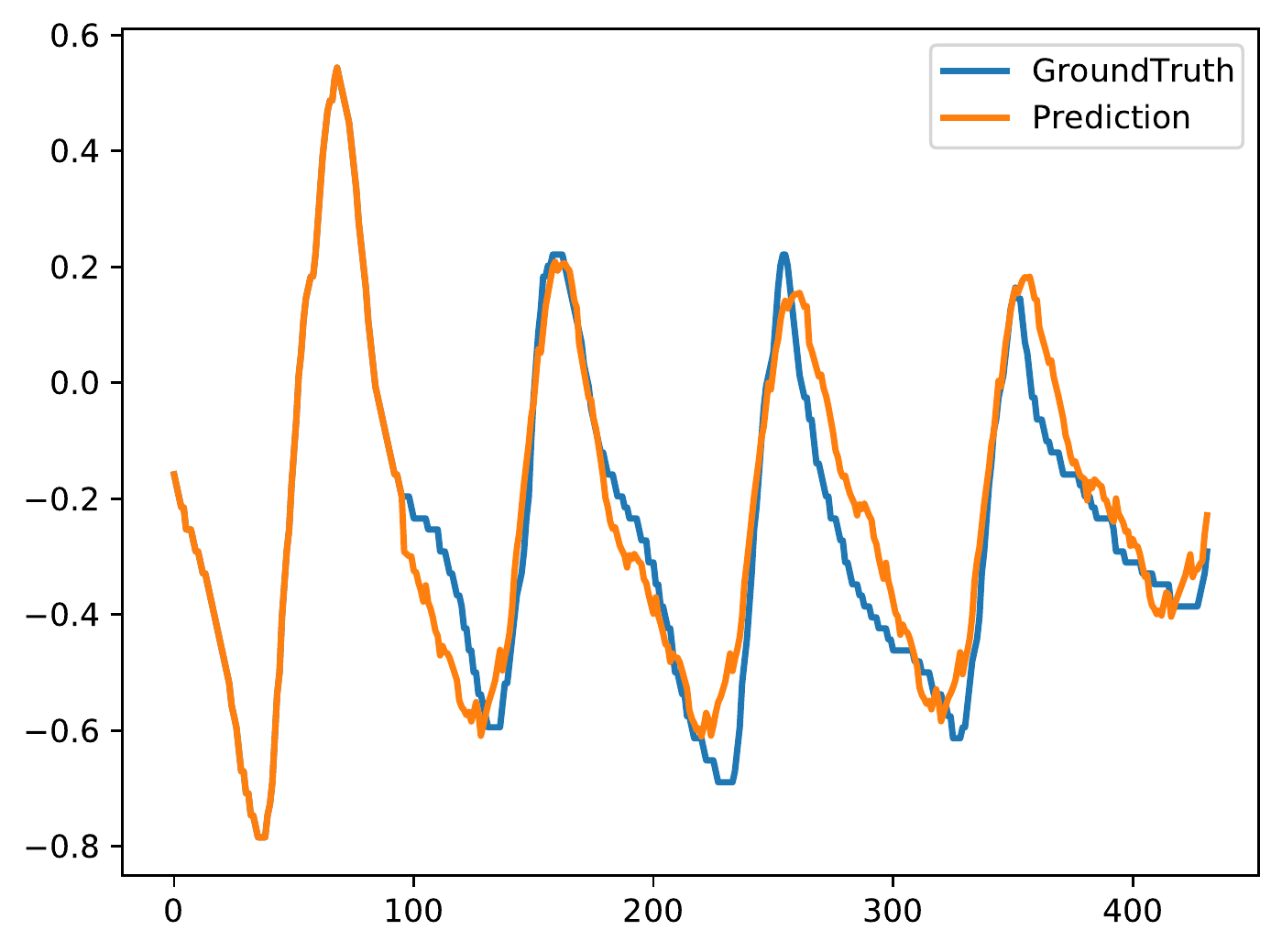}
    }
    \subfigure[Autoformer]{
    \includegraphics[width=0.22\columnwidth]{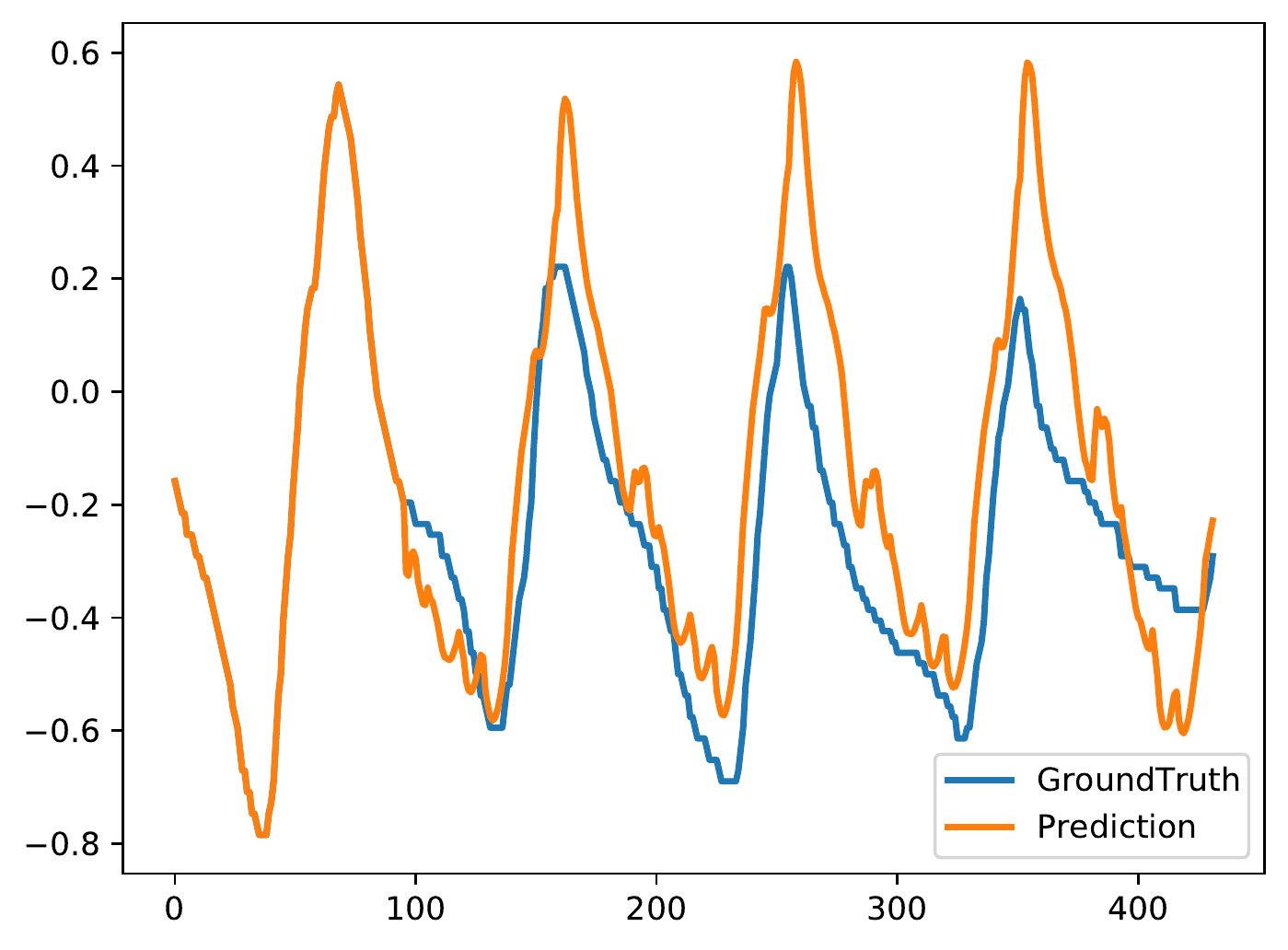}
    }
    \subfigure[Informer]{
    \includegraphics[width=0.22\columnwidth]{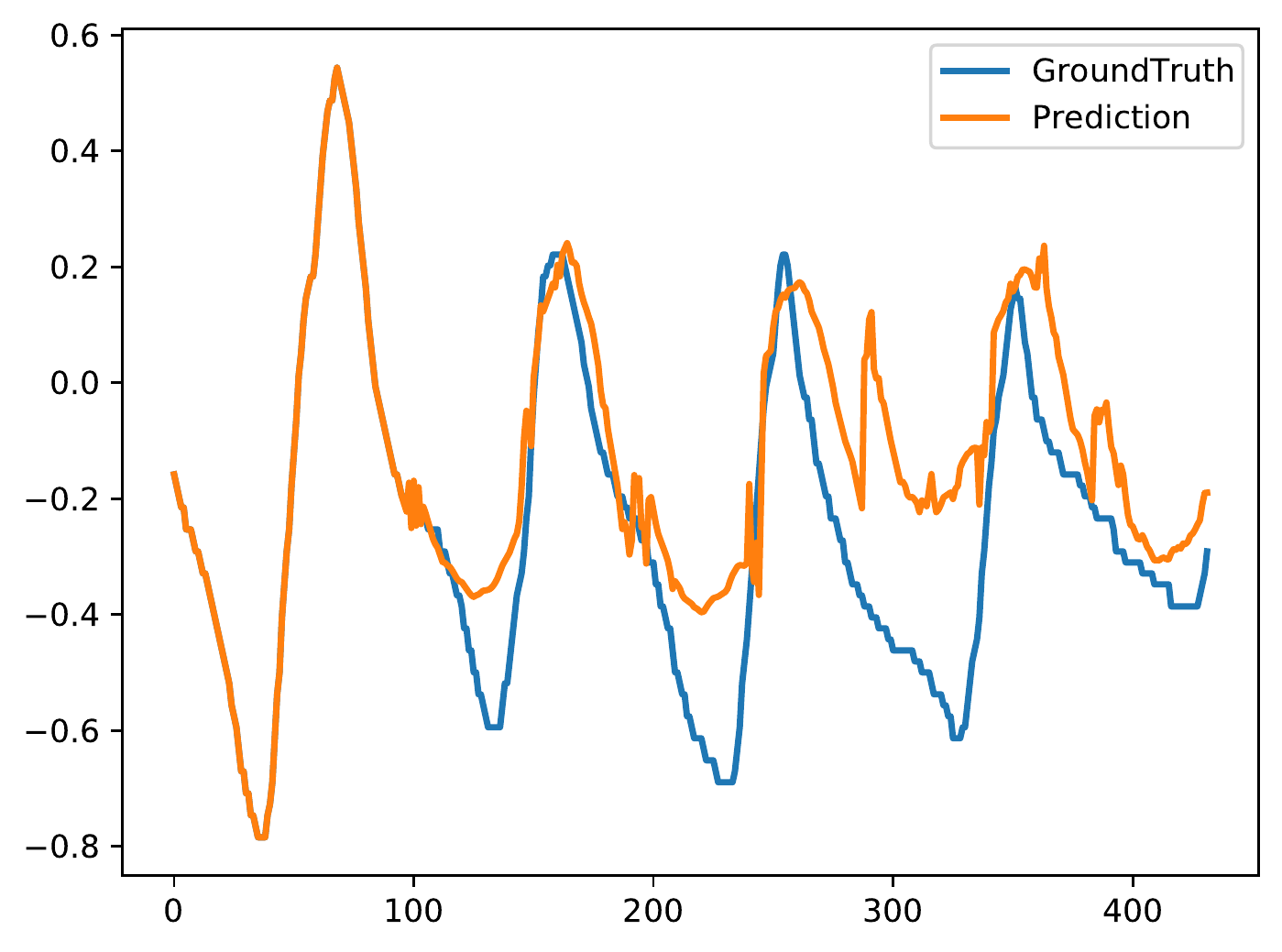}
    }
    \subfigure[LogTrans]{
    \includegraphics[width=0.22\columnwidth]{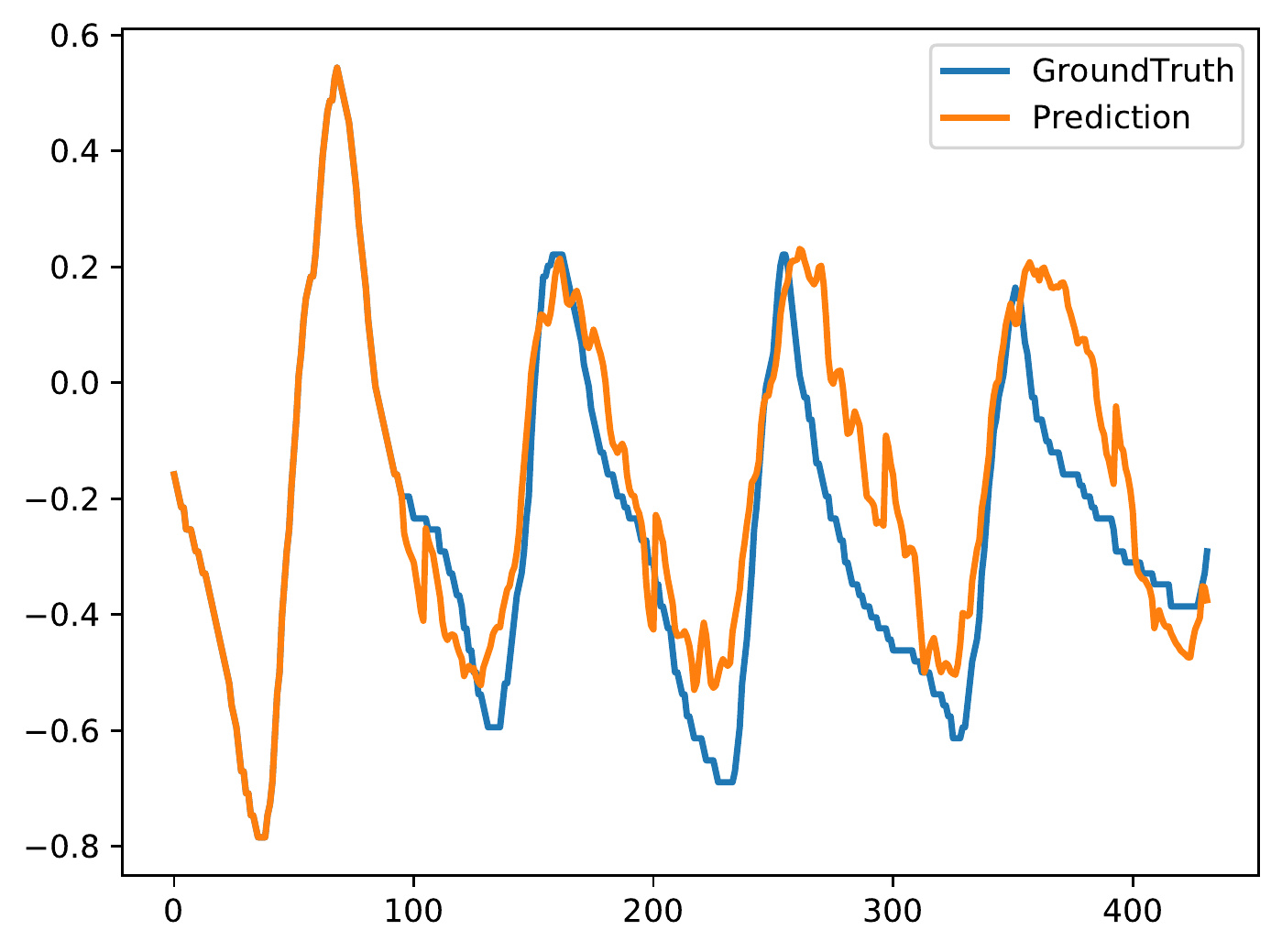}
    }
  \caption{The prediction results on the ETTm2 dataset under the input-96-predict-336 setting. }
    \label{fig:15}
\end{center}
\vskip -0.2in
\end{figure}

\begin{figure}[H]
\vskip 0.2in
\begin{center}
    \subfigure[Preformer]{
    \includegraphics[width=0.22\columnwidth]{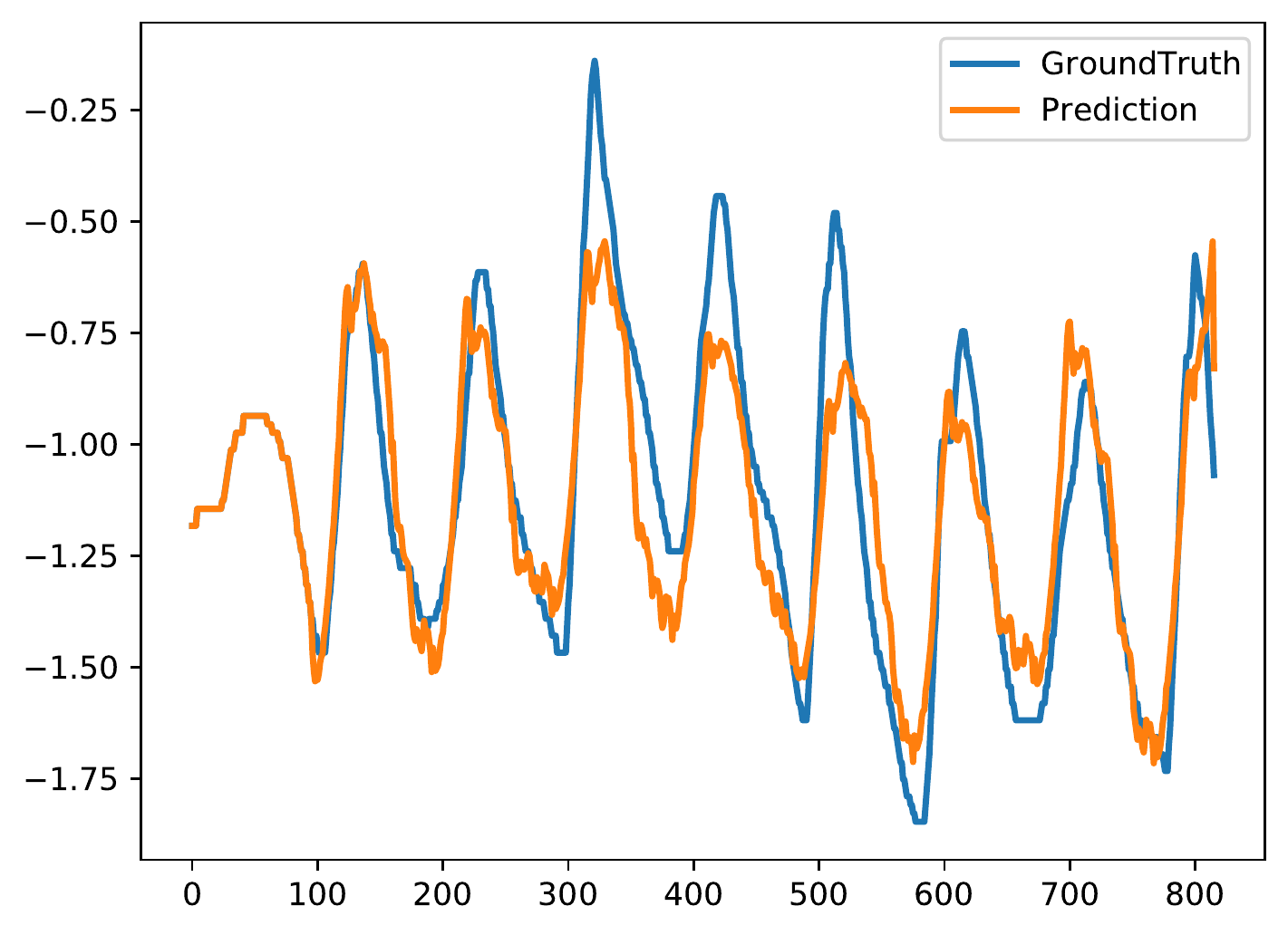}
    }
    \subfigure[Autoformer]{
    \includegraphics[width=0.22\columnwidth]{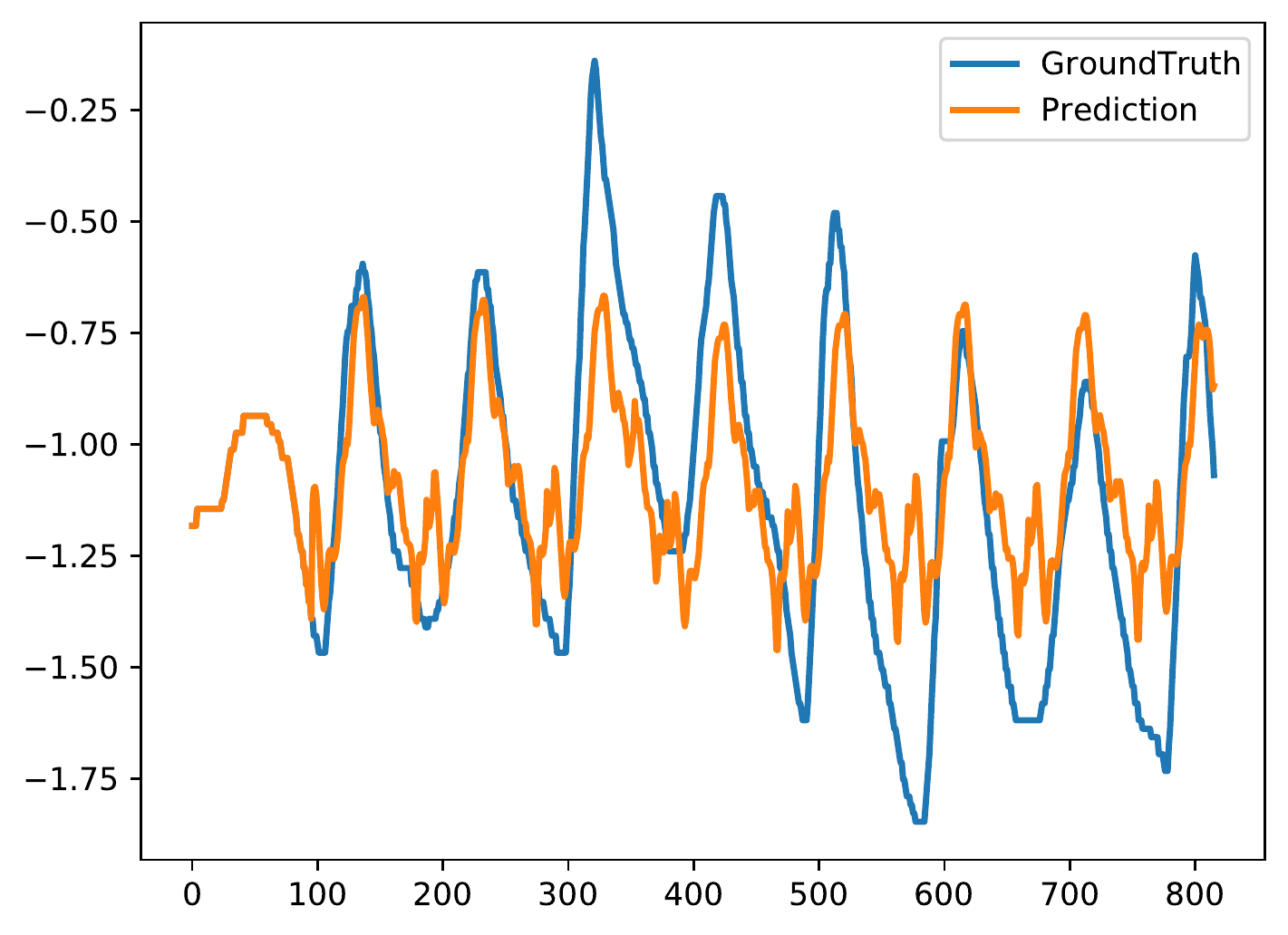}
    }
    \subfigure[Informer]{
    \includegraphics[width=0.22\columnwidth]{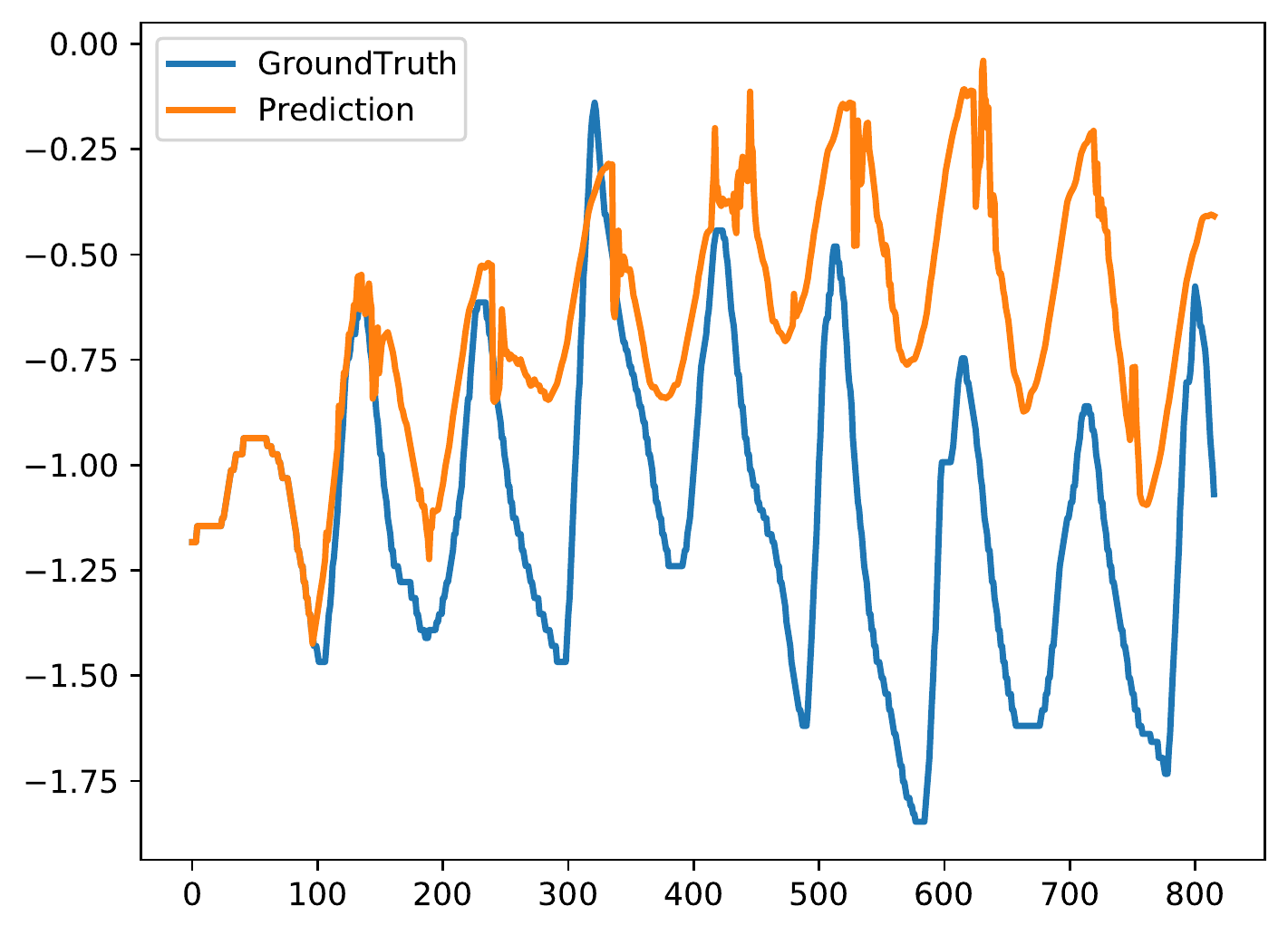}
    }
    \subfigure[LogTrans]{
    \includegraphics[width=0.22\columnwidth]{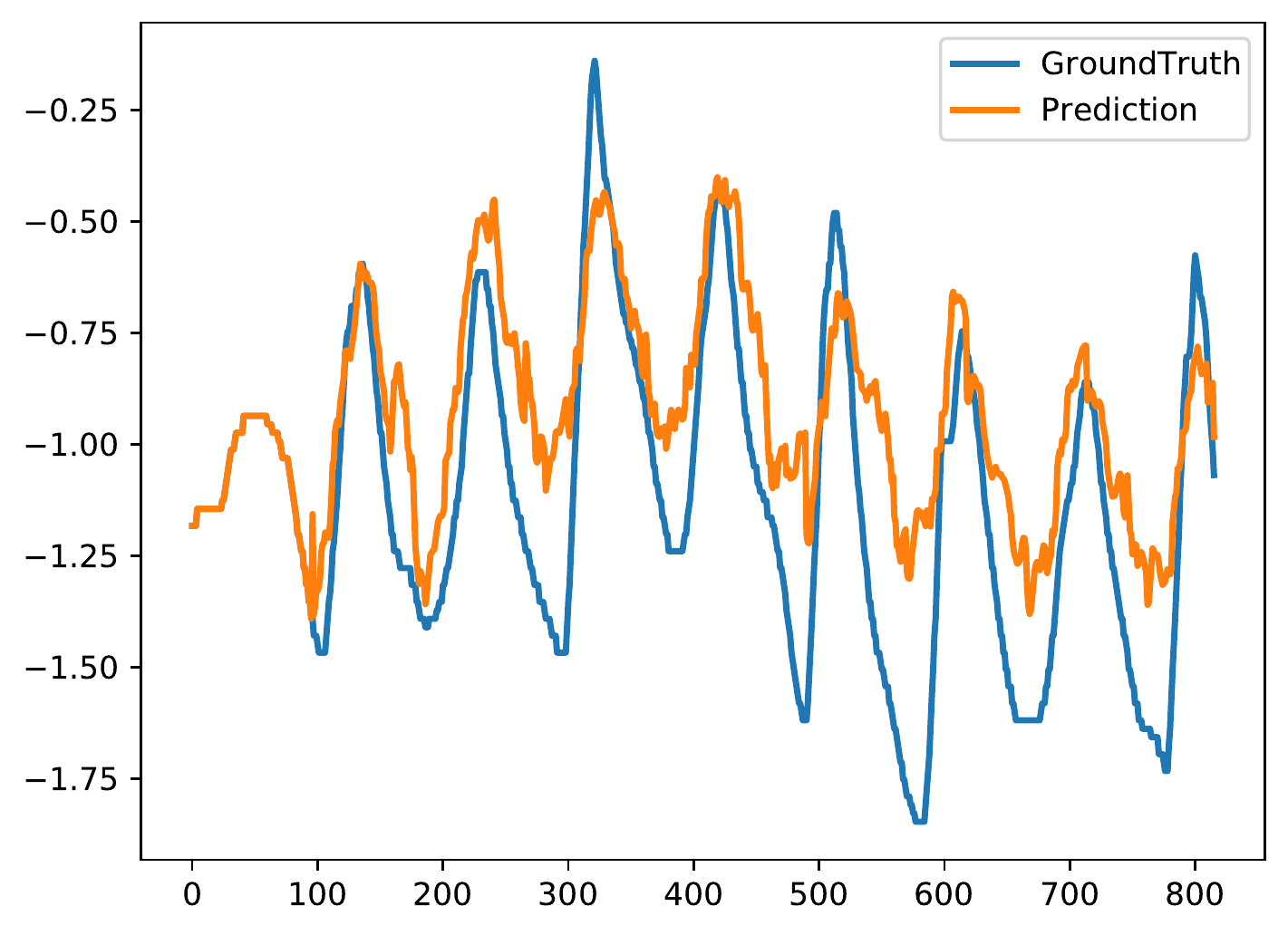}
    }
  \caption{The prediction results on the ETTm2 dataset under the input-96-predict-720 setting. }
    \label{fig:16}
\end{center}
\vskip -0.2in
\end{figure}

As shown in \cref{fig:13,fig:14,fig:15,fig:16}, we plot the univariate forecasting results of several Transformer-based models on the ETTm2 dataset. 

\end{document}